%% file: acm_main.tex
\documentclass[manuscript,screen,authorversion,nonacm]{acmart} 
\AtBeginDocument{%
  }
    
\setcopyright{acmcopyright}
\copyrightyear{2024}
\acmYear{2024}
\acmDOI{XXXXXXX.XXXXXXX}

\acmJournal{TELO}
\acmVolume{37}
\acmNumber{4}
\acmArticle{111}
\acmMonth{8}

\usepackage{soul,url,tikz}
\usepackage{graphicx}
\usepackage{booktabs,balance}
\usepackage{dsfont}
\usepackage{xcolor}
\usepackage{comment}
\usepackage{subfigure}
\usepackage{svg}

\newcommand{\iohx}[0]{IOHxplainer}

\begin{document}

\title[Explainable Benchmarking for Iterative Optimization Heuristics]{Explainable Benchmarking\\ for Iterative Optimization Heuristics}

\author{Niki van Stein}
\email{n.van.stein@liacs.leidenuniv.nl}
\orcid{0000-0002-0013-7969}
\affiliation{%
  \institution{LIACS, Leiden University}
  \streetaddress{Niels Bohrweg 1}
  \city{Leiden}
  \country{Netherlands}
  \postcode{NL-2333}
}

\author{Diederick Vermetten}
\email{d.vermetten@liacs.leidenuniv.nl}
\orcid{0000-0003-3040-7162}
\affiliation{%
  \institution{LIACS, Leiden University}
  \streetaddress{Niels Bohrweg 1}
  \city{Leiden}
  \country{Netherlands}
  \postcode{NL-2333}
}

\author{Anna V. Kononova}
\email{a.kononova@liacs.leidenuniv.nl}
\orcid{0000-0002-4138-7024}
\affiliation{%
  \institution{LIACS, Leiden University}
  \streetaddress{Niels Bohrweg 1}
  \city{Leiden}
  \country{Netherlands}
  \postcode{NL-2333}
}

\author{Thomas B{\"a}ck}
\email{t.h.w.baeck@liacs.leidenuniv.nl}
\orcid{0000-0001-6768-1478}
\affiliation{%
  \institution{LIACS, Leiden University}
  \streetaddress{Niels Bohrweg 1}
  \city{Leiden}
  \country{Netherlands}
  \postcode{NL-2333}
}

\renewcommand{\shortauthors}{van Stein, et al.}

\begin{abstract}
Benchmarking heuristic algorithms is vital to understand under which conditions and on what kind of problems certain algorithms perform well. In most current research into heuristic optimization algorithms, only a very limited number of scenarios, algorithm configurations and hyper-parameter settings are explored, leading to incomplete and often biased insights and results. This paper presents a novel approach we call explainable benchmarking. Introducing the \iohx{} software framework, for analyzing and understanding the performance of various optimization algorithms and the impact of their different components and hyperparameters. We showcase the framework  in the context of two modular optimization frameworks. Through this framework, we examine the impact of different algorithmic components and configurations, offering insights into their performance across diverse scenarios. We provide a systematic method for evaluating and interpreting the behaviour and efficiency of iterative optimization heuristics in a more transparent and comprehensible manner, allowing for better benchmarking and algorithm design.
\end{abstract}



\maketitle

\section{Introduction}
The rapid development of new algorithmic ideas and the modification of existing algorithms in optimization heuristics~\cite{baeck2023evolutionary} poses a challenge in understanding their overall impact. Traditional benchmarking methods are often used to evaluate algorithms in isolation, with a single algorithm configuration (hyper-parameter setting) or with a limited set of a few variations against a limited set of state-of-the-art algorithms, leading to limited insights into their comparative performance and practical applicability. This study addresses these challenges by employing modular optimization approaches and explainable AI techniques in order to derive insights into the algorithmic behaviour of a large set of algorithm components (modules) and their hyperparameters. Modular optimization  frameworks allow for the comparison of various modifications on a core algorithm, facilitating a deeper understanding of each component's influence on the algorithm's  performance in different scenarios. There is already a wide variety of modular algorithm frameworks available, but their application for explicit explainability of the various algorithmic components and settings has been relatively unexplored. This paper aims to bridge this gap by providing a comprehensive framework for explainable benchmarking in iterative optimization heuristics and by providing a software library (\iohx{}) to facilitate researchers to use the proposed framework.

\section{Background}\label{sect:related}

\subsection{Modular Optimization Heuristics}

While new algorithmic ideas or modifications of existing algorithms are proposed constantly, it is generally hard to understand the impact of these changes within a wider context. Algorithms are often benchmarked in isolation and compared to other implementations of the same algorithm~\cite{kononova2023tiobr}, which limits the interpretation of the results. Modular approaches to optimization algorithms resolve some of these concerns by enabling comparisons of a wide variety of modifications on the same core algorithm implementation. \textit{Modularization} is achieved in such algorithms via maintaining multiple options for each algorithm operator (module) and making option choices per module in a fully independent manner since options of all modules are ensured to be compatible with each other. While modular algorithms have been around for decades, e.g. in the paradisEO framework~\cite{paradiseo22} and many algorithm families have been modularized~\cite{psox, lopez2010automatic}, their use for the explicit purpose of explainability has been relatively limited. 

Often, modular algorithms are combined with algorithm configuration techniques to obtain specialized parameter settings for specific optimization problems~\cite{schede2022survey, lopez2012automatic, prager2020per, thomaser2023optimizing}. Similarly, large-scale benchmarking of many algorithms has been used to create powerful algorithm selectors~\cite{nevergrad_wizard}. However, further explanations of the interactions between modules or their impact on specific aspects of algorithm behaviour have been rather limited, focussing mostly on studying the link between low-level landscape features and the performance of different modules~\cite{KG_mod_pred, kostovska2022importance}. 







\subsection{Explainable AI for EC}

Explainable Artificial Intelligence (XAI) has emerged as a critical field in the landscape of AI and machine learning, addressing the growing need for transparency, trustworthiness and interpretability of AI models. As AI systems increasingly influence various aspects of daily life and critical decision-making processes, understanding how these systems arrive at their decisions has become paramount. This understanding is vital not only for the users and developers of AI systems but also for regulators and stakeholders who require assurance that AI systems are fair, safe and aligned with ethical standards.

In the context of optimization algorithms, particularly in evolutionary computation and metaheuristics, XAI plays a vital role in deciphering the mechanisms behind the algorithms' search processes and behaviours. These algorithms often involve complex interactions and adaptations, and XAI can aid in demystifying these processes. By applying XAI techniques, researchers and practitioners can gain insights into how different components of an algorithm contribute to its overall performance, how solutions evolve over time and what factors influence the convergence towards optimal solutions.

XAI can illuminate the internal workings of algorithm components, such as selection, crossover and mutation in genetic algorithms, or the update rules in swarm intelligence algorithms. Understanding these components through the lens of XAI can lead to the development of more efficient and effective algorithms. Furthermore, by analyzing the search behaviour of these algorithms, XAI can help identify strengths and weaknesses in different problem domains, guide the tuning of algorithm parameters and inspire new algorithmic innovations.

The field of XAI meets the field of EC in two directions, using XAI to enhance and better understand EC algorithms, and using EC to advance XAI \cite{Zhou2024}. For example, the approach GP-tSNE \cite{lensen2020genetic} adapts the classical t-SNE algorithm, which is used for visualizing data sets, with a Genetic Programming approach to provide an interpretable mapping from the original data to the embedded points. In \cite{evans2019s} the authors propose a method using multi-objective genetic programming to construct decision trees that represent any black-box classifier while being as much interpretable as possible. The proposed method aims to simultaneously
maximize the ability of the tree to reconstruct the predictions of a black-box model and also maximize interpretability by minimizing the complexity of trees. The authors use a modified version of NSGA-II \cite{deb2002fast}.
In the work conducted by Ferreira et al. \cite{ferreira2020applying}, a novel approach named Genetic Programming Explainer (GPX) is introduced. This method, rooted in Genetic Programming (GP), is designed to develop a localized explanatory model tailored for a specific input instance. Mirroring the strategy employed by LIME, GPX starts by selecting a target input that needs explanation. Around this chosen input, it generates a cluster of closely related data points. The key differentiation of GPX from LIME lies in its use of GP for the creation of symbolic expression trees. These trees are evolved through GP to effectively represent the functioning of the underlying complex, black-box model within the vicinity of the sampled data points. This approach allows GPX to produce more nuanced and expressive local explanations compared to the linear models typically used in LIME.

The above is just some of the works in this growing `bridging' field of research. While these works focus on using EC methods to enhance and develop XAI methods in order to understand black-box models, XAI can also be used in order to get an understanding of `black-box' EC algorithms.

To start with, Landscape analysis, as highlighted by Malan \cite{malan_survey_2021}, stands as a crucial intersection between XAI and EC. This field utilizes a range of tools aimed at comprehending and elucidating algorithm behaviour based on problem features. It also focuses on predicting algorithm performance and facilitating automatic algorithm configuration and selection. Recently, works like those by Trajanov et al. \cite{trajanov2022explainable,trajanov2021explainable,van2023doe2vec} have specifically targeted explainable landscape-aware prediction.
One way of capturing an algorithm's behaviour is through its search landscape trajectory. This trajectory is indicative of the algorithm's evolution, revealing moments of discovery, stagnation, or premature convergence. Ochoa et al. \cite{OCHOA2021SearchTrajectoryNetworks} introduced search trajectory networks as a method for visualizing these trajectories. 
Fyvie et al. \cite{Fyvie2021} proposed search trajectories as a pathway to XAI in EC. They employed Principal Component Analysis on solutions explored by an Evolutionary Algorithm to identify prevalent population features at each generation. This approach allows for a visual representation of the algorithm's progress.
Hyper-heuristics and parameter selection studies, like the work by Drake \cite{drake2020recent}, shows that certain parameter settings enable EC methods to exhibit generalist behaviours, performing well across diverse functions. 

While the above works focus on either using specific EC methods to enhance XAI, or use XAI to better understand particular EC methods, in this work XAI is used to better differentiate and understand the influence between many different EC methods and their many different configurations. Allowing researchers to benchmark large numbers of algorithms on a wide variety of benchmark functions and enabling the answer of various questions such as, e.g.,:
\begin{itemize}
    \item Which hyperparameters are important to tune?
    \item Which benchmark functions are most descriminative between configurations?
    \item Which algorithm configuration performs most stable (performs well on average)?
    \item How much can we gain from per-instance configuration?
\end{itemize}

\section{Explainable Benchmarking}\label{sect:XB}
The field of benchmarking for black-box optimization has matured a lot over the last decades, with well respected benchmark suites such as the many versions of BBOB in the COCO platform~\cite{hansen2021coco}, the CEC function suites~\cite{cec2017} and the wide variety of benchmarks within Nevergrad~\cite{nevergrad}, and tools and analysis platforms such as IOHanalyzer~\cite{wang2022iohanalyzer}. 
Nevertheless, most of the research papers published to date which introduce new optimization heuristics contain only a very limited exploration of the capabilities and limitations of the proposed approach. Often these experimental setups are also biased towards the strengths of the newly proposed algorithm (component) to increase the chance of paper acceptance.
Since we already know that one overall best heuristic does not exist \cite{NFLT}, it is imperative that we not only aim to develop and research optimizers that improve the state of the art (on a limited and often biased experimental setup) but also that we aim to understand the search behaviour~\cite{Mitran2023} and characteristics of algorithms, their inner components and their hyperparameters. Once we have a clear understanding of how these components and their parameters influence the search (or once an AI model can capture this information), we can develop tailor-made optimisation algorithms with good performance generalisable over specific domains or landscape features. 

When introducing a new algorithm or a new algorithm component, it is essential to know on what kind of function landscapes this new contribution has either a positive or negative influence on the performance, how much it interacts with other parameters of the algorithm configuration and how much the component's hyperparameters require tuning depending on the function landscapes. This and other behaviour analyses give new insights for the community to work with. Instead, most existing works only show that the novel proposed component improves the base-algorithm (with just one or a few specific configurations) on average. Even worse, this improvement is most of the time only marginal and it is often unclear what the underlying reasons for the improvement are, such as on what kind of function landscape, dimensionality or other landscape feature it works better or worse.

In this work, we propose a new framework for experimentation and analysis of black-box iterative optimization heuristics, which we call \emph{Explainable Benchmarking}, including a software framework implementation called \emph{\iohx{}}~\cite{code_xplainer_niki_van_stein_2024_10568760}. In this section, we will first discuss the past and current practices of benchmarking heuristics together with their pros and cons, we will then introduce the proposed framework in detail and show how to use the framework by applying it on two large groups of modular heuristics. 

Note that the proposed framework could in principal even be used in analysing and experimenting in other computer science fields such as machine learning and AI, but for clarity and concrete examples, here we limit its usage to iterative optimization heuristics.

\subsection{Existing methodologies}
In this subsection, we briefly survey the current and older ways of performing benchmarking in the field of EC. 

\subsubsection{Visual test}\label{sect:old_visual}

The oldest approach to benchmarking EC algorithms constitutes a simple demonstration of the applicability of these algorithms to a class of problems and consists of plotting fitness values over time, averaged across several independent runs, with no plots from competitor algorithms but rather `verdict-like' applicability conclusions, e.g., ~\cite{dejong1981using} on Boolean Satisfiability problems from as long ago as 1989.

Over time, this `record-average-plot-observe' methodology persisted and evolved into reporting average performance values in time for a selection of handpicked competitor algorithms, on functions from slowly emerging benchmarking suites which are subsequently visually analyzed and discussed in the body of the paper, e.g.,~\cite{kaveh2016}. A similar approach of simple comparisons persists also in other closely related parts of the optimization community such as papers on quality-diversity algorithms, e.g., \cite{fontaine2023}. A marginally deeper approach consists of showing a multitude of convergence plots on the entire benchmark suite and visually identifying differences in performance, as encouraged in, e.g., workshops for individual benchmarking suites like ~\cite{espinoza2023} from the recurring ACM GECCO Workshop on Black Box Optimization Benchmarking. Making meaningful conclusions from this multitude of plots requires attention and significant experience. Sometimes, such plots can be enriched with statistical testing on algorithm rankings (e.g., Friedman, Nemenyi tests~\cite{mengqi2022}). In other cases, the authors still opt for showing performance results in the table form with only mean, standard deviation, best and worst values, significantly limiting the interpretability of their results. 

\subsubsection{Benchmarking Function Groups}\label{sect:function_groups} 
The idea of identifying well-performing algorithms depending on the high-level properties of the objective functions such as multimodality, presence of global structure, separability, etc., has found its way to the setup of most benchmarking suites to test different behaviours, thus, prescribing a methodology of analyzing algorithm performance per function group, e.g., ~\cite{pierezan2018,Boks2021}. Such grouping of results can be further based on estimated problem properties like, e.g., Exploratory Landscape Analysis (ELA) features~\cite{mersmann2011exploratory,munoz2015alg}.
 
\subsubsection{Domain Specific Benchmarks}\label{sect:domain}
Optimization algorithms are typically evaluated using the now-de-facto standardised benchmark suites to discern variations in their performance. Nonetheless, the extent to which such benchmark functions accurately reflect the characteristics of the real-world problems remains uncertain leading to a potential overfitting in algorithmic choices~\cite{vanderblom2023}. Therefore, some researchers advocate in favour of designing specialized algorithms that work well on a specific problem domain, e.g, structural optimization of vehicles~\cite{kohira2018}, 
hydrothermal emission optimisation~\cite{glotic2015} and  
underwater glider path planning~\cite{zamuda2019success}. 
Such setup is also possible for multiobjective optimization~\cite{Tanabe2020}. 

\subsubsection{Modern benchmarking practices that lack immediate explainability}\label{sect:good_nonxai_bench}
Despite further advancements in benchmarking methodologies~\cite{bartzbeielstein2020benchmarking}, current practices still often overlook the necessity of ablation studies and fail to investigate the influence of parameters on algorithm performance thoroughly. This oversight can lead to an incomplete understanding of how and why certain algorithms perform better under specific conditions. 
A common approach in modern benchmarking involves aggregating data from multiple runs on multiple functions (but not over dimensionalities)~\cite{hansen2021coco}. This method assumes a uniform probability distribution over a set of problems, which simplifies the analysis but may not accurately reflect real-world scenarios where problem frequencies and distributions can significantly vary. 
Furthermore, the reliance on absolute runtime distributions and performance profiles as universally comparable metrics across publications is problematic~\cite{hansen2021coco}. While these metrics provide valuable insights, they often lack context regarding the scalability and adaptability of algorithms to different problem dimensions or target precisions. 
To address these shortcomings, it is crucial to include evaluations on the scalability of solvers across various target precisions and problem dimensions~\cite{hansen2021coco}. Such analyses can offer more nuanced insights into the algorithm's performance, highlighting its strengths and limitations in a more granular and explainable manner.
It is furthermore recommended to limit claims of algorithm or parameter suitability strictly to the tested problem instances and to exercise caution when generalizing performance across different problem instances or classes. Such generalizations could lead to inaccuracies and overestimations of algorithmic versatility~\cite{Brownlee2007}. 

A further step in the direction of extending the explainability of benchmarking results has been taken in~\cite{vermetten2024largescale}, where the authors outline the methodology of exploring complementarity in algorithm performances to aid automated data-driven algorithm selection and switching, assess algorithm parameterisations and implementations for completeness and tunability. 

\subsection{\iohx{}}\label{sec:toolbox}
\begin{figure*}[t]
\centering
	\includegraphics[width=0.9\textwidth,trim=0mm 5mm 0mm 10mm,clip]{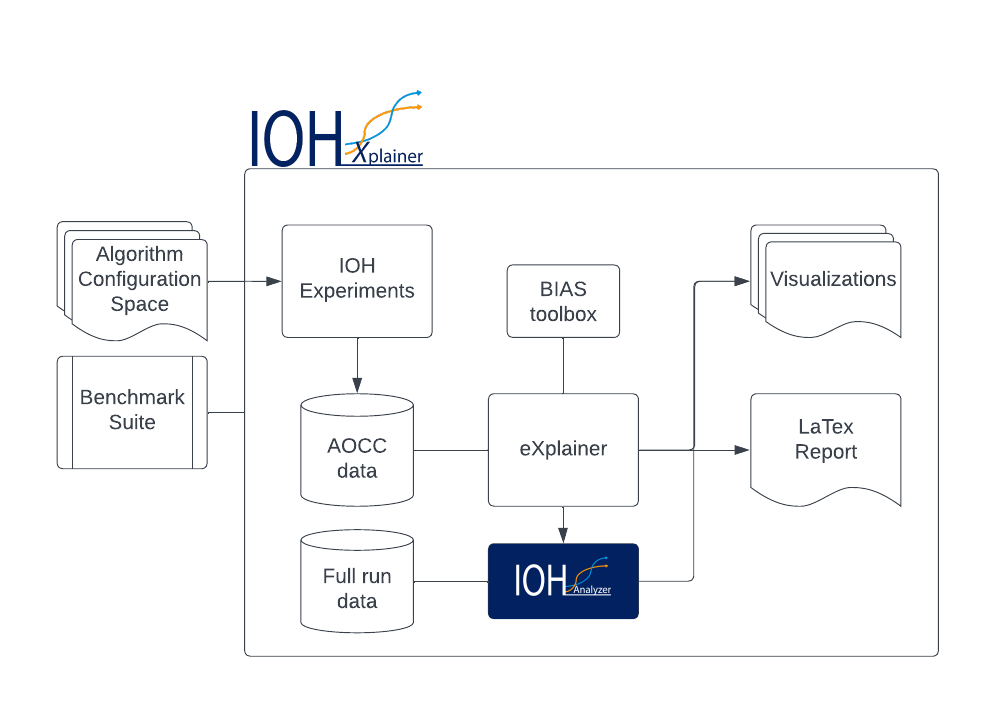}
\caption{The proposed \emph{\iohx{}} framework,
for the automatic and explainable analysis of algorithm components and \\ hyperparameters. \label{fig:framework} }
\end{figure*}

To overcome some of the limitations listed above and to better analyze, compare and explain behaviour in optimization heuristics, we propose \textbf{\iohx{}} (Figure \ref{fig:framework}), a software package following the proposed Explainable Benchmarking framework. The framework is utilizing XAI techniques in order to automatically extract meaningful visualisations and statistics from large empirical studies, allowing the benchmarking of thousands, even millions of algorithm configurations.

The proposed framework works with any (modular) algorithm and with (large) collections of algorithms. By using a flexible configuration space definition~\cite{configspace}, similar to the configuration spaces used by SMAC~\cite{lindauer2022smac3}, we support both continuous, integer and categorical parameters. Hyper-parameter 
dependencies can also be defined and taken into account. \iohx{} uses either a full exhaustive grid for running the experiments, or a random design with a user-defined number of algorithm configurations. Once the configuration space of algorithms and the budget to evaluate this space has been defined, each algorithm configuration (instance) is evaluated for a fixed evaluation budget and a set of dimensions on a benchmark suite. In this work, we used the well-known and popular black-box optimization benchmark (BBOB) with $24$ noiseless functions \cite{bbobfunctions} originally introduced as part of the COCO framework~\cite{hansen2021coco}. In general, the \iohx{} 
framework can be used with any benchmark.
From the (parallel) benchmark runs, we collect the performance measure, which can be fixed-budget, fixed target or any-time, as is the case here. 
Once each instance of the experiment is finished, the explainer part of \iohx{} uses several explainable AI techniques to visualize the contribution of each of the algorithm's hyperparameters and components. First, for each function in the benchmark suite, a deep Xg-boost regression model is trained on the algorithm configuration and performance data for that function. Once the regression model is fitted to the data, a Shapley approximation technique, TreeSHAP \cite{lundberg2020local2global} is utilized to extract the SHAP values for each hyper-parameter and algorithm component. These SHAP values represent the marginal contribution of the algorithm components and parameters towards the any-time performance measure of each run. 
When a full grid over the configuration space of algorithm instances is performed, it would even be possible to calculate the exact SHAP values from the data, however, in practice, this is infeasible due to the high computational cost.
All the approximated SHAP values are aggregated per function and visualized using a swarm plot, allowing algorithm designers and practitioners to inspect the influence of each of the hyperparameters per function. 


While the proposed framework uses SHAP \cite{NIPS2017_7062} as local explainer, in principle any model-agnostic feature attribution XAI method can be used, such as LIME~\cite{ribeiro2016should} or others.
In addition, global explainers can be used such as global sensitivity analysis methods \cite{GSApaper,GSAcode,sobol1993sensitivity,hutter2014efficient} (Morris, Sobol, functional ANOVA, etc.) to understand the global first and second order sensitivity scores of each module and hyper-parameter. See our Github repository~\cite{code_xplainer_niki_van_stein_2024_10568760} for the sensitivity analysis reports of our data.

The proposed explainable benchmarking pipeline is setup in a modular fashion and allows for easy expansion in both analysis techniques as well as new algorithms and algorithm components.
To use \iohx{}, first a data collection step such as a large grid or large random sample of algorithm configurations has to be performed. The \iohx{} toolbox runs these configurations on a benchmark suite such as BBOB for a pre-defined number of functions, instances and random seeds.
Optionally the full convergence data can be logged using IOHanalyzer, 
otherwise only the final performance measure 
from each run is being logged to save storage space.
Once the benchmark data has been collected, the data can be either combined with existing data collections from modCMA or modDE, or processed individually. \iohx{} automatically extracts the SHAP values and creates a LaTeX report with extensive plots and tables for an easy comparison of the outcome. The toolbox can also utilize the BIAS Toolbox to perform additional checks on the single and average best configurations.

\subsection{Robustness to Instance Variation and Random Seed Variation}
In addition to analyzing the impact of algorithmic components, the approach described in Section~\ref{sec:toolbox} can be used to study the robustness of an optimization algorithm. Since we collect performance data for multiple runs of each algorithm configuration, the seed used for these runs can be added in as a virtual module of \iohx{} for the analysis. If the algorithm's performance is independent of the used seed, as is generally expected from the state-of-the-art optimizers, the seed value should have little explanatory power and thus be assigned a low SHAP value. As such, the \textit{seed-based variance}~\footnote{In \iohx{}, such seed-based variance will be referred to as `stochastic variance' (see Section~\ref{sect:contrib_hyperp}).} is a useful sanity check for algorithm designers. 

In addition to the stochasticity inherent to the optimization algorithm, such algorithms are generally run on multiple instances of a problem. These instances can be created, e.g., by rotation and translation of the search space, resulting in slightly different problems, to which an algorithm might or might not be invariant~\cite{hansen2011impacts, caraffini2019study}. 
Thus, looking at the explanatory power of the instance ID (`instance variance') 
allows us to check these algorithmic invariances. It should be noted that this analysis is only applicable to benchmark sets which support this type of instance creation and can thus not directly be used for arbitrary optimization problems. 


\section{Explainable Benchmarking of 52\,128 Algorithm Configurations}\label{sect:real-world}


This section discusses benchmarking a large set of heuristic optimisation algorithms by applying the \iohx{} toolbox. 

\subsection{Used Algorithms}
\input{tables/modcma_modules}
\input{tables/modde_modules}
To illustrate the functionality of \iohx{}, we perform two large-scale experiments on modular optimization algorithms: modular CMA-ES (modCMA)~\cite{modcma} and modular DE (modDE)~\cite{modDE}. For both algorithms, we perform a full enumeration of a set of available modules and hyperparameters (categorical values) and we perform a grid of discritized values for the continuous parameters.

For clarity, we refer to every option we vary as a \emph{module} throughout the remainder of this paper, and a fully specified combination of these modules creates a \emph{configuration} of the algorithm. For modCMA, the modules listed in Table~\ref{tab:modcma_modules} lead to a total of $15\,840$ configurations, which is a significantly larger set of configurations than used in previous works~\cite{kostovska2022importance, kostovska_inprogress}. In the case of modDE, the modules we use are shown in Table~\ref{tab:modde_modules}, which leads to a total of $36\,288$ configurations. 

\subsection{Data collection setup}
For each configuration we consider, we collect performance data on all $24$ BBOB functions in both dimensionality $5$ and $30$, with a budget of $10\,000$ evaluations. We use the first $5$ instances of each function and perform $5$ independent runs per instance. We make use of an anytime performance measure: the normalized Area Over the Convergence Curve (normalized AOCC). This measure is defined as follows:
\begin{equation*}
    \textit{AOCC}(\Vec{y}) = \frac{1}{B} \sum_{i=1}^{B} \left( 1-\frac{ \min(\max((y_i), \textit{lb}), \textit{ub})  - \textit{lb}}{\textit{ub} - \textit{lb}} \right)
\end{equation*}
where $\Vec{y}$ is the sequence of best-so-far function values reached during the optimization run, $B=10\,000$ is the budget, $\textit{lb}$ and $\textit{ub}$ are the lower and upper bound of the function value range of interest. 
It should be noted that this definition of AOCC is equivalent to the area under the ECDF curve with infinite targets between the chosen bounds. To keep with the conventions of the BBOB suite~\cite{hansen2022anytime}, the function values are log-scaled before calculating the AOCC, and we use bounds $10^{-8}$ and $10^2$ for the $5D$ functions. Since $10^2$ is a hard target to reach within $10\,000$ evaluations on some $30D$ problems, we change the upper bound to $10^8$ for the $30D$ case. 

The proposed AOCC measure and experimental setup allows us to compare the any time performance and module contributions across functions, instances and dimensions.

\subsection{Discussion of results on ModCMA}

\subsubsection{Contribution of hyperparameters}\label{sect:contrib_hyperp}

\begin{figure*}[t]
\centering
	\includegraphics[height=0.15\textheight,trim=4mm 0mm 30mm 0mm,clip]{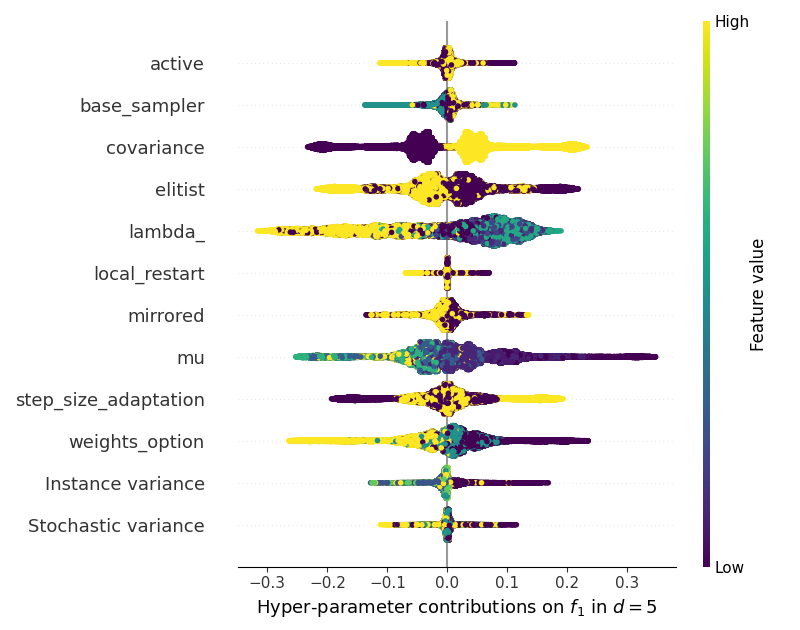}
	\includegraphics[height=0.15\textheight,trim=60mm 0mm 30mm 0mm,clip]{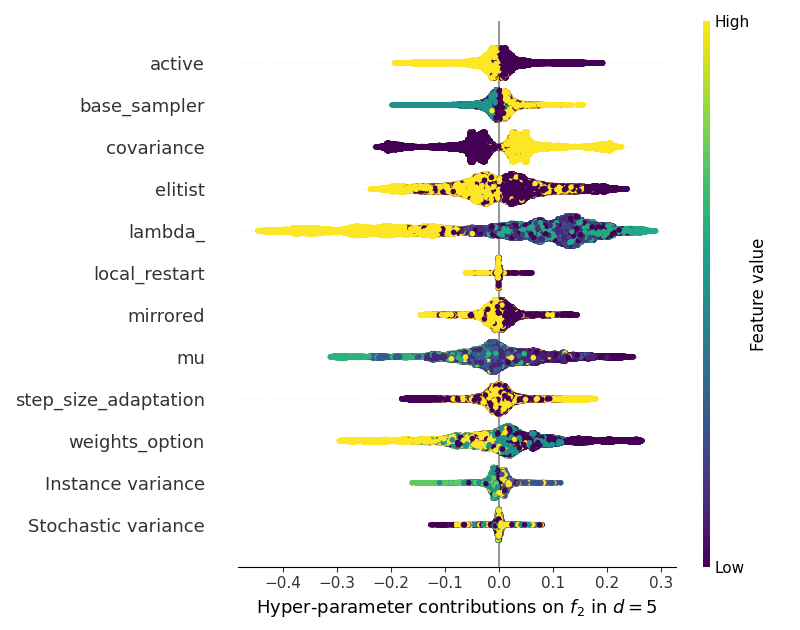}
	\includegraphics[height=0.15\textheight,trim=60mm 0mm 30mm 0mm,clip]{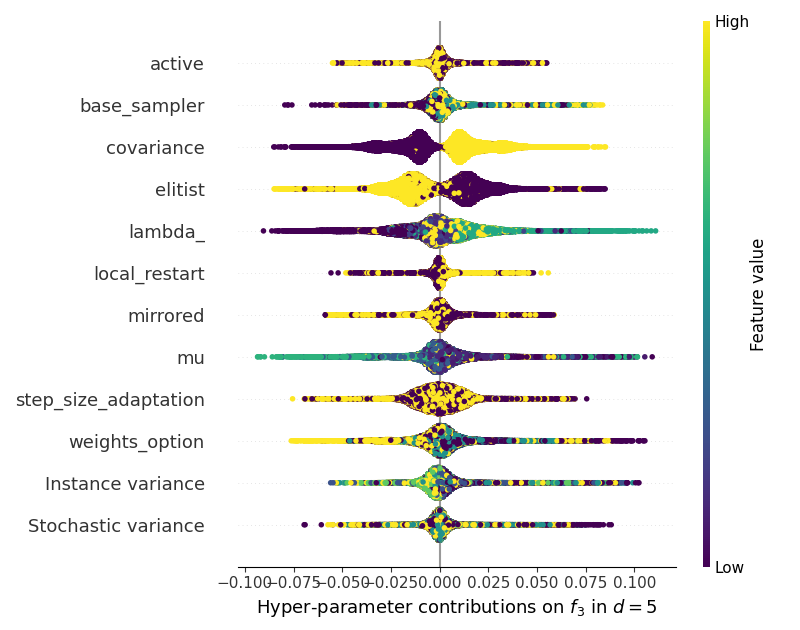}
	\includegraphics[height=0.15\textheight,trim=60mm 0mm 30mm 0mm,clip]{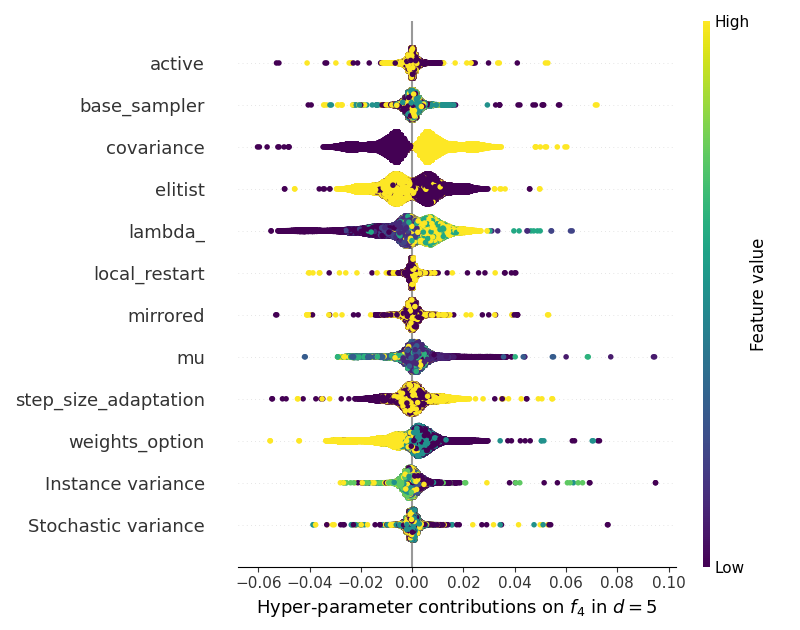}
	\includegraphics[height=0.15\textheight,trim=60mm 0mm 30mm 0mm,clip]{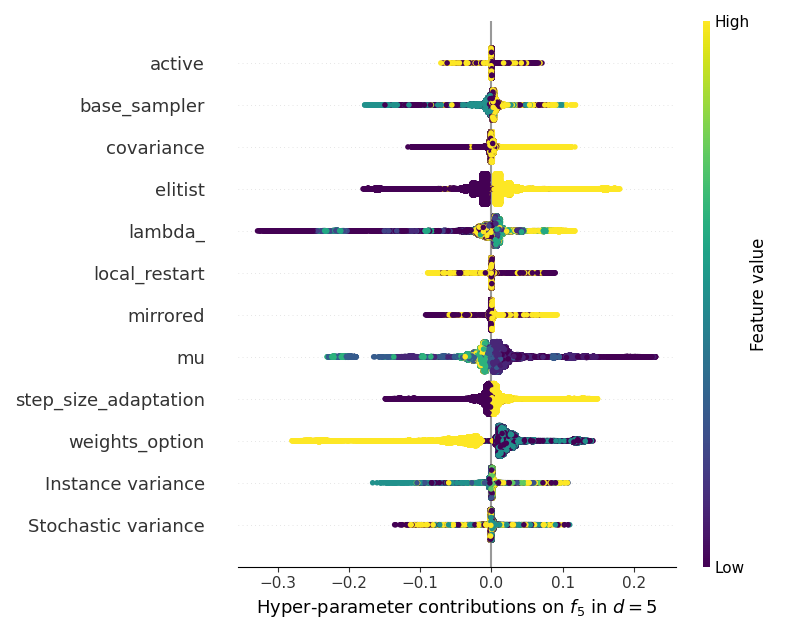}
	\includegraphics[height=0.15\textheight,trim=60mm 0mm 10mm 0mm,clip]{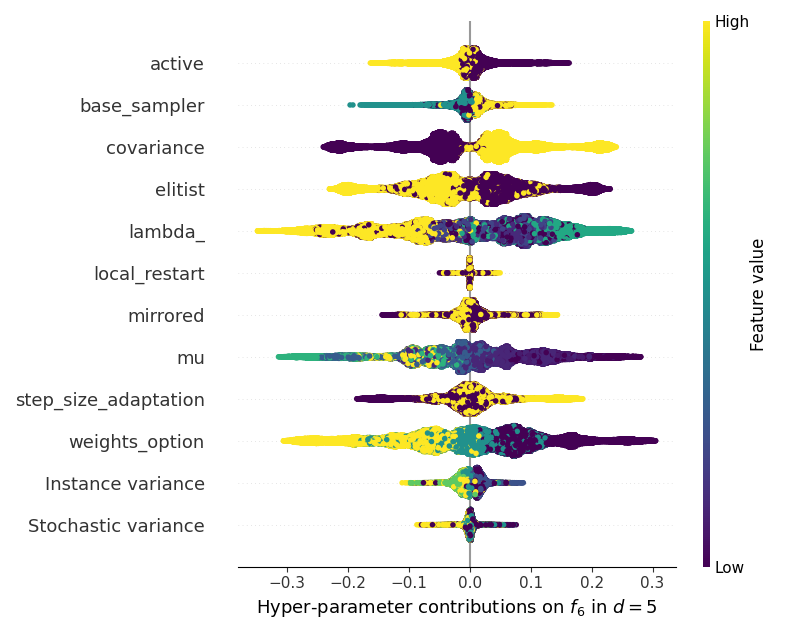}
 
	\includegraphics[height=0.15\textheight,trim=4mm 0mm 30mm 0mm,clip]{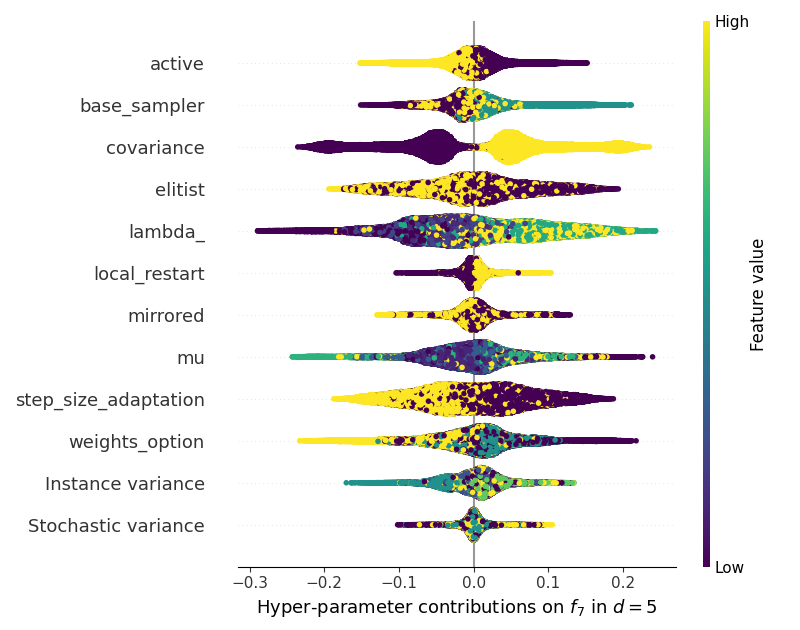}
	\includegraphics[height=0.15\textheight,trim=60mm 0mm 30mm 0mm,clip]{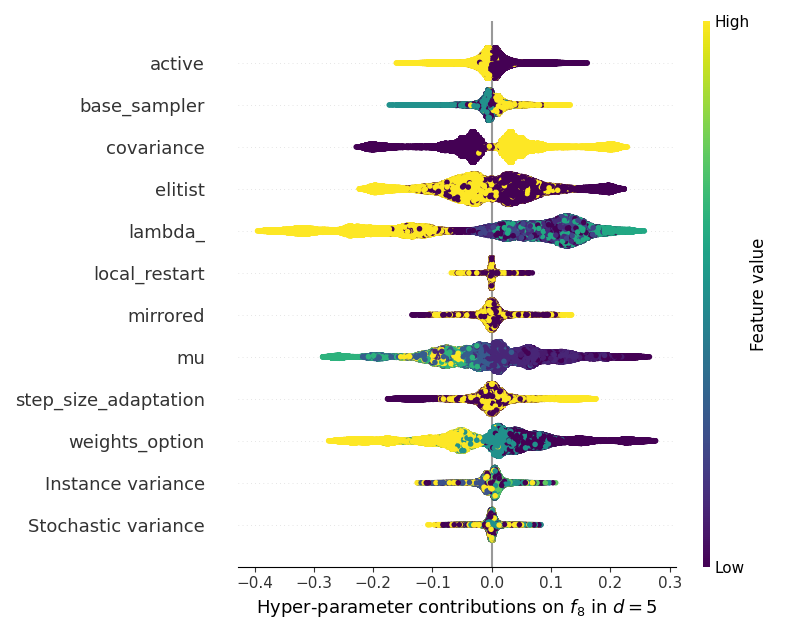}
	\includegraphics[height=0.15\textheight,trim=60mm 0mm 30mm 0mm,clip]{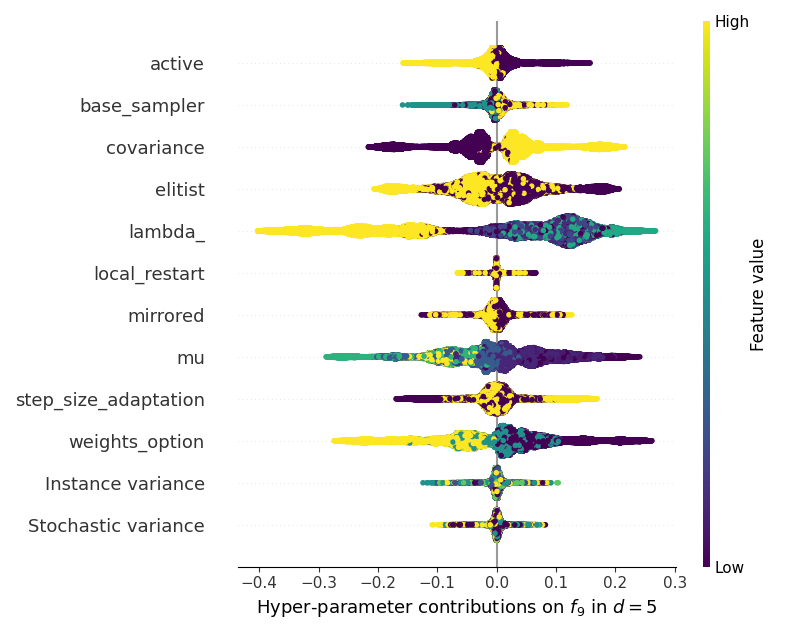}
	\includegraphics[height=0.15\textheight,trim=60mm 0mm 30mm 0mm,clip]{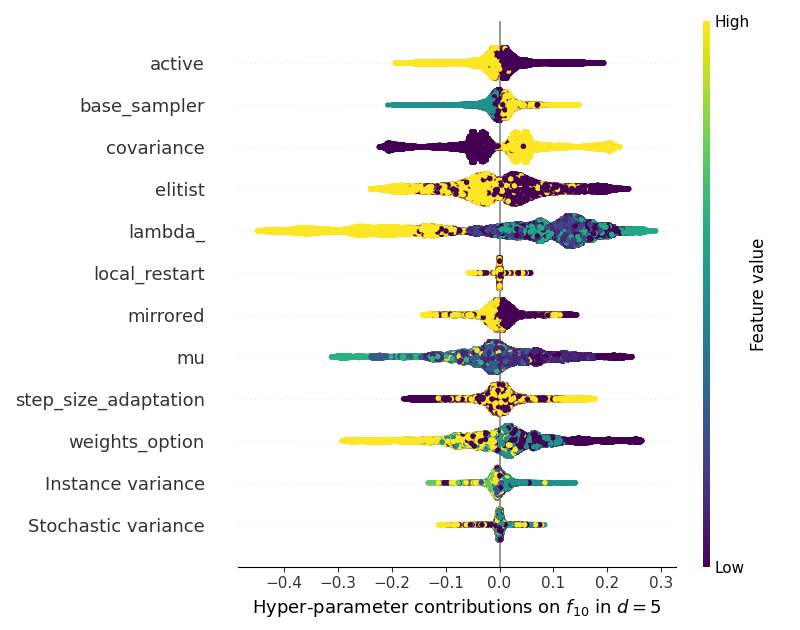}
	\includegraphics[height=0.15\textheight,trim=60mm 0mm 30mm 0mm,clip]{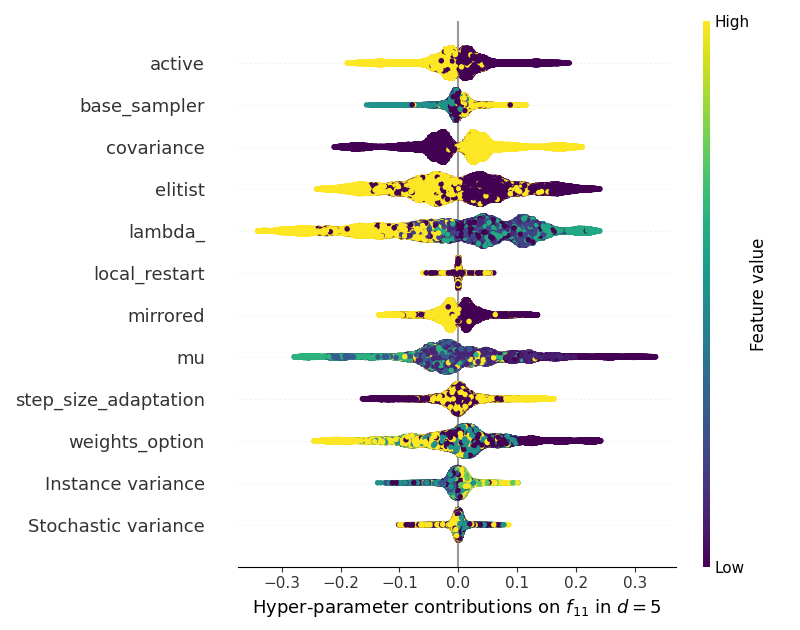}
	\includegraphics[height=0.15\textheight,trim=60mm 0mm 10mm 0mm,clip]{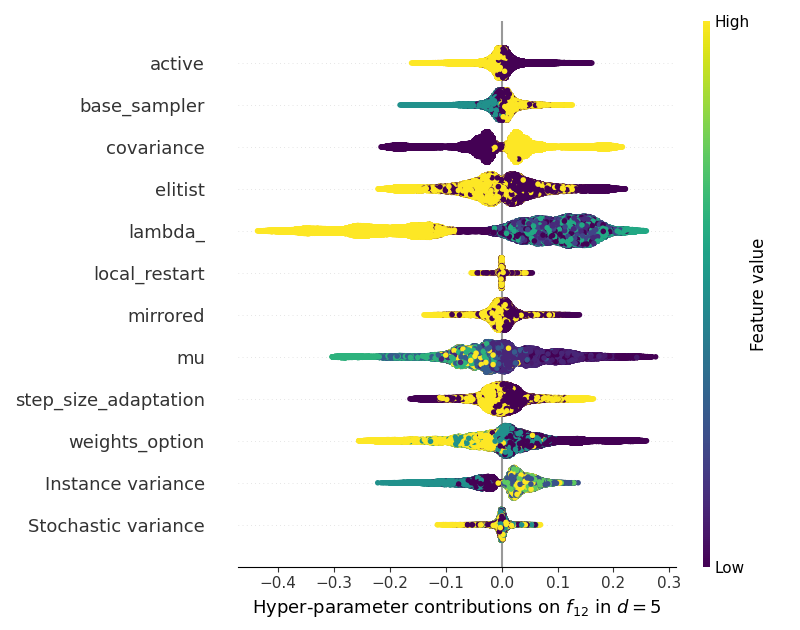}
 
	\includegraphics[height=0.15\textheight,trim=4mm 0mm 30mm 0mm,clip]{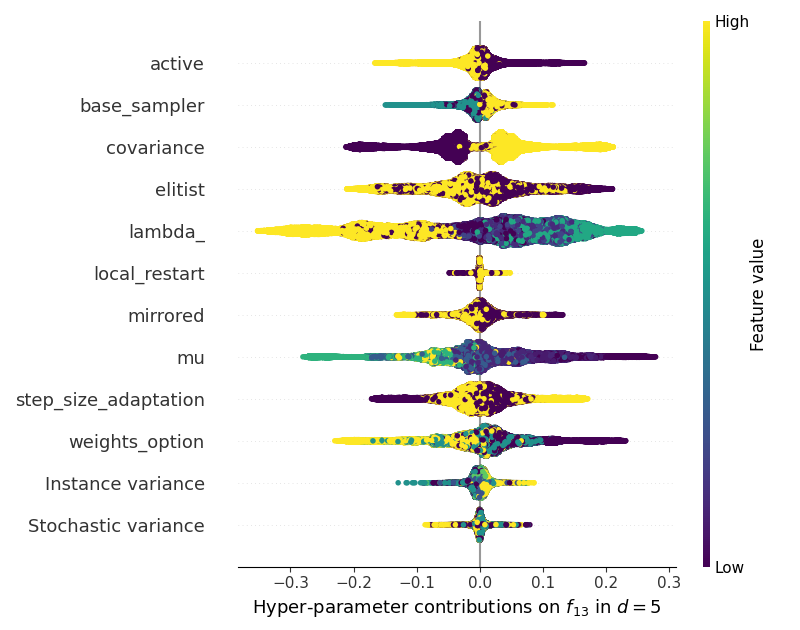}
	\includegraphics[height=0.15\textheight,trim=60mm 0mm 30mm 0mm,clip]{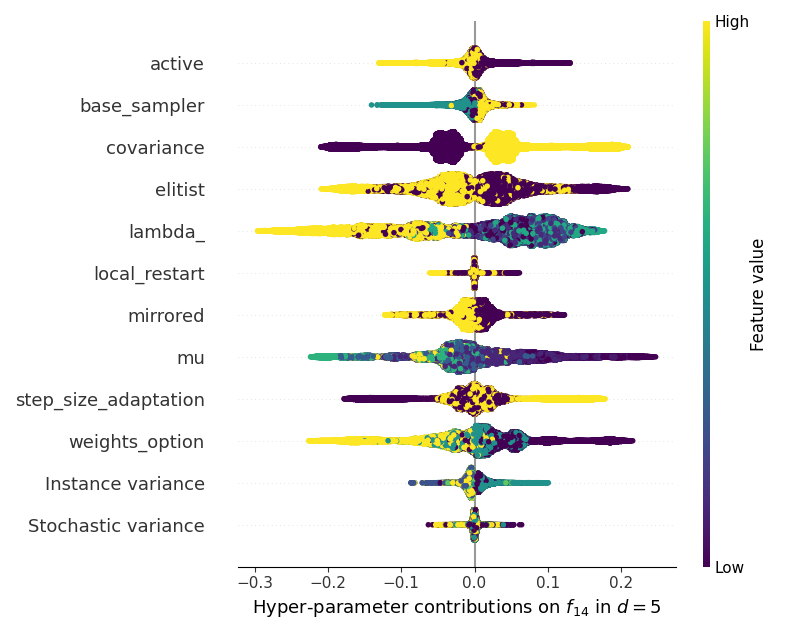}
	\includegraphics[height=0.15\textheight,trim=60mm 0mm 30mm 0mm,clip]{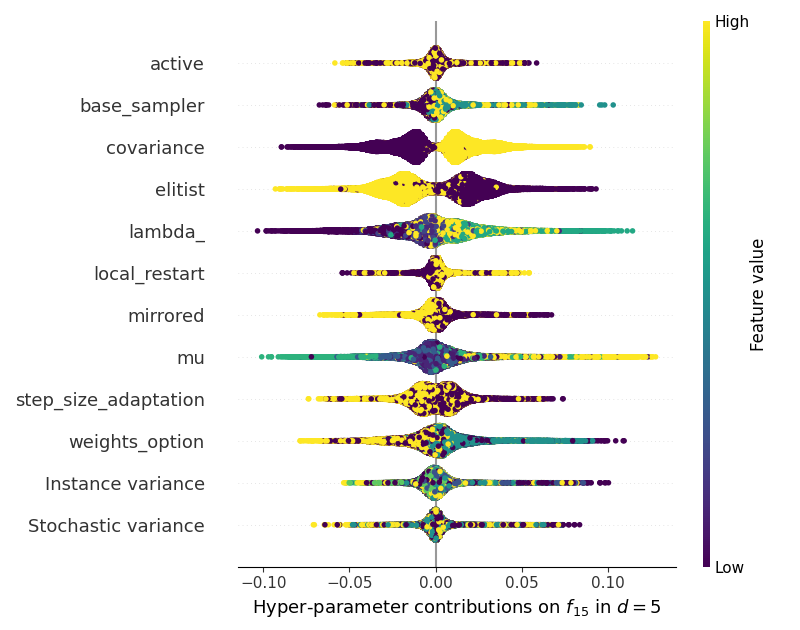}
	\includegraphics[height=0.15\textheight,trim=60mm 0mm 30mm 0mm,clip]{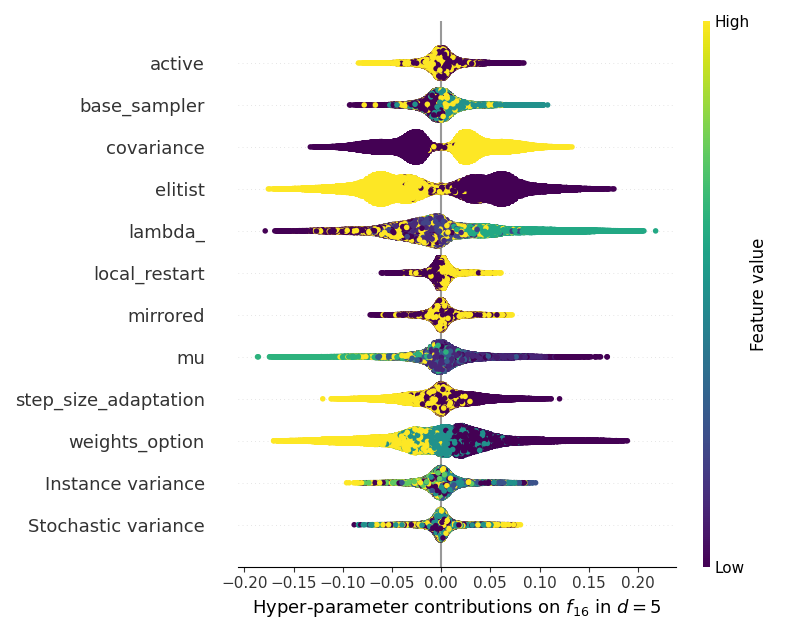}
	\includegraphics[height=0.15\textheight,trim=60mm 0mm 30mm 0mm,clip]{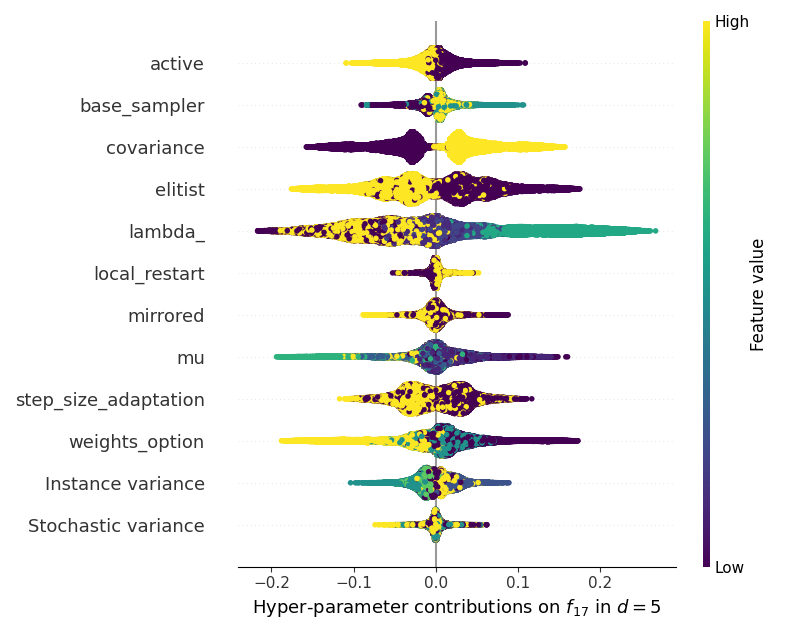}
	\includegraphics[height=0.15\textheight,trim=60mm 0mm 10mm 0mm,clip]{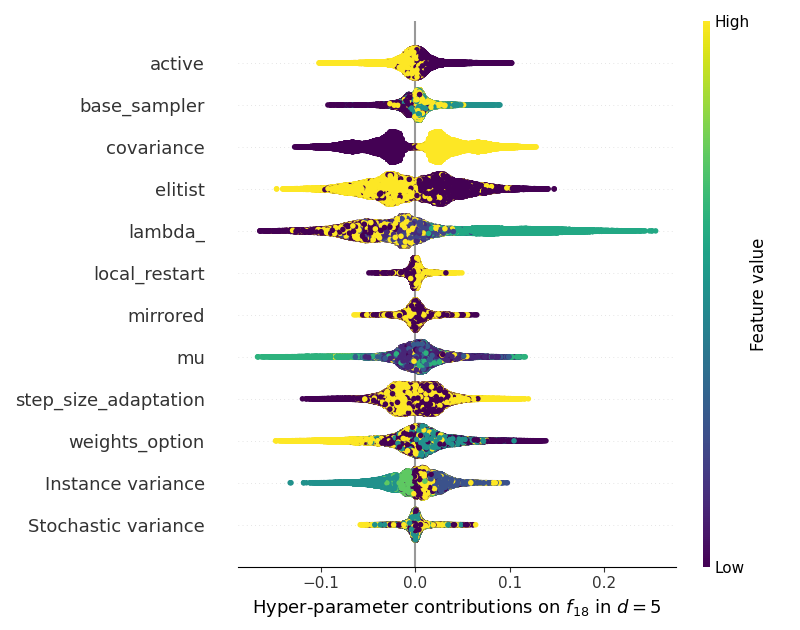}
 
	\includegraphics[height=0.15\textheight,trim=4mm 0mm 30mm 0mm,clip]{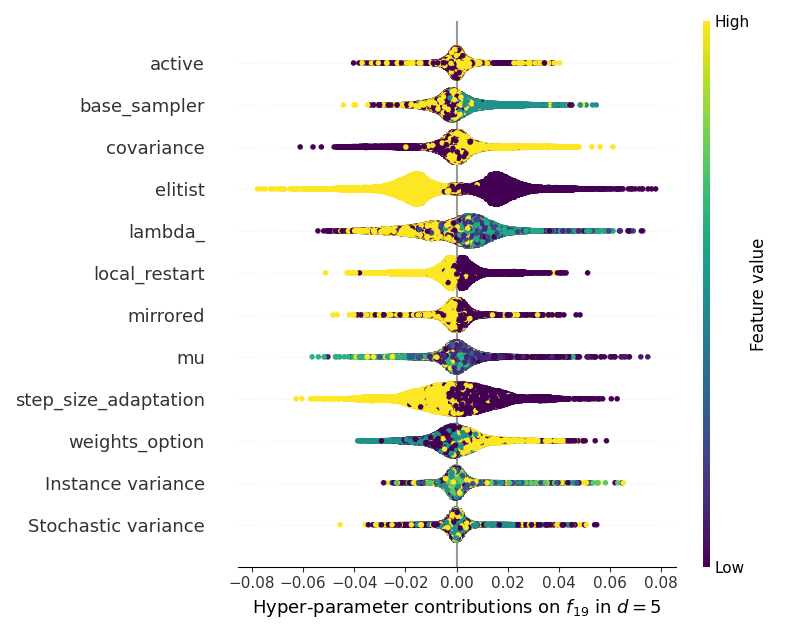}
	\includegraphics[height=0.15\textheight,trim=60mm 0mm 30mm 0mm,clip]{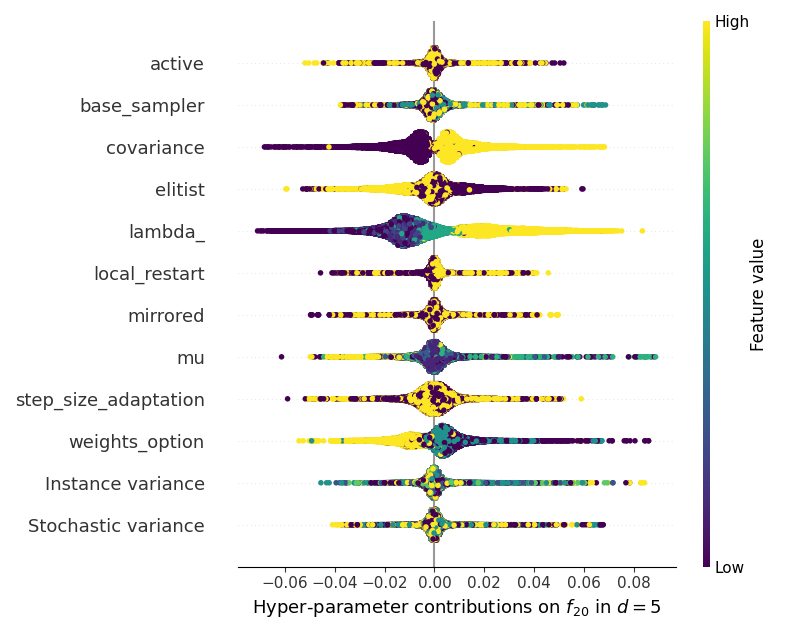}
	\includegraphics[height=0.15\textheight,trim=60mm 0mm 30mm 0mm,clip]{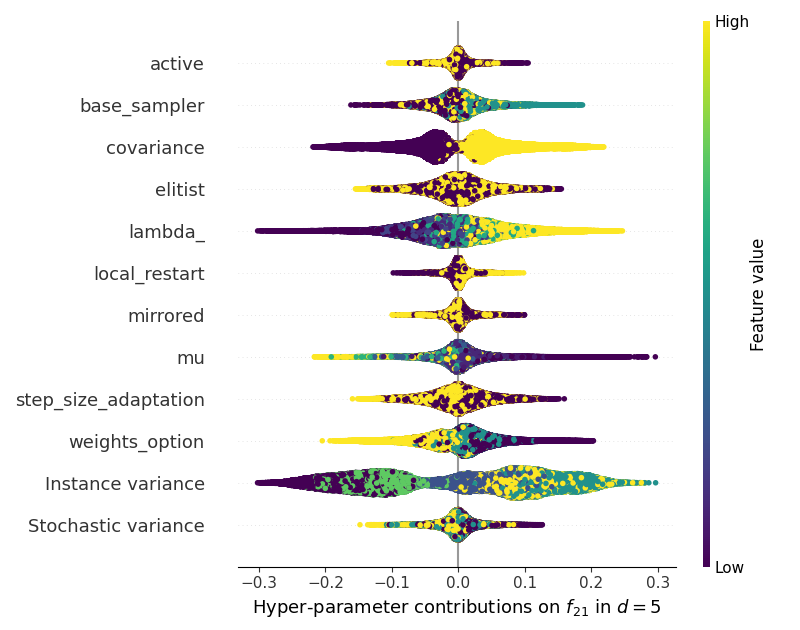}
	\includegraphics[height=0.15\textheight,trim=60mm 0mm 30mm 0mm,clip]{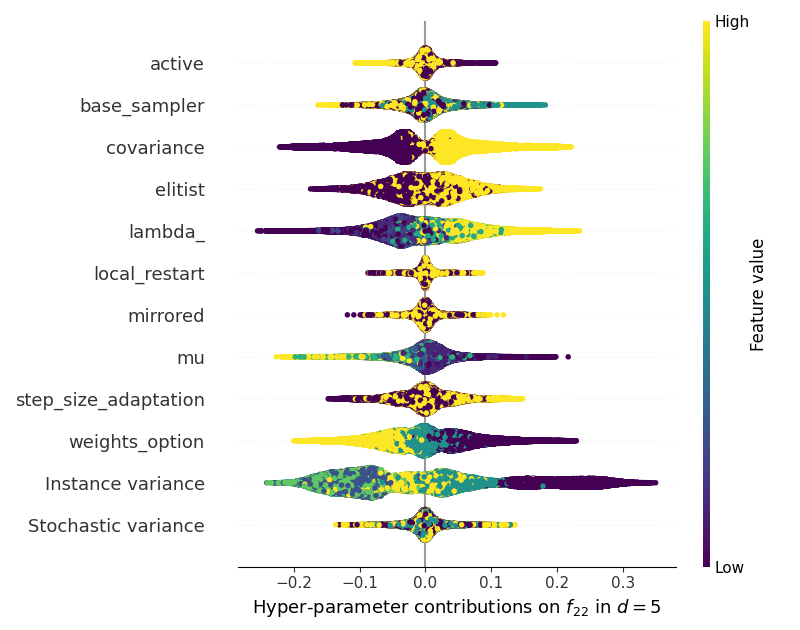}
	\includegraphics[height=0.15\textheight,trim=60mm 0mm 30mm 0mm,clip]{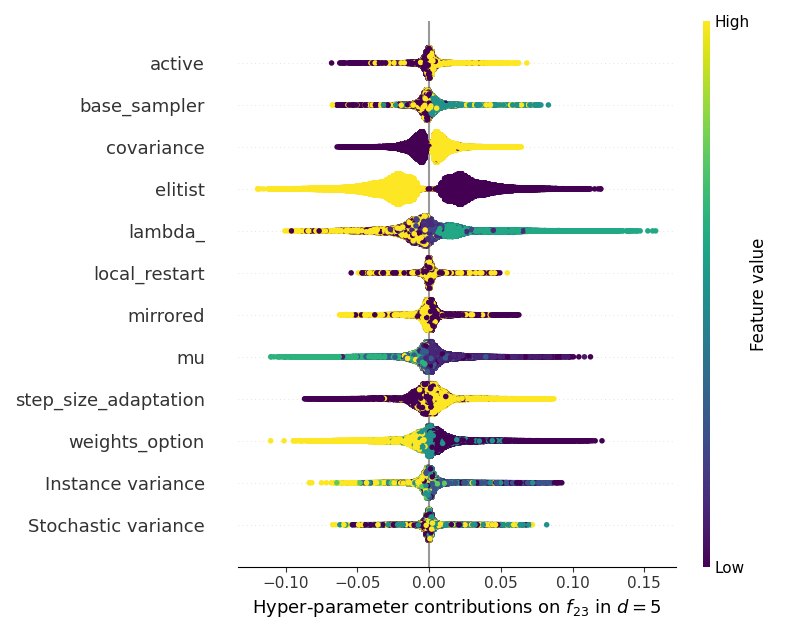}
	\includegraphics[height=0.15\textheight,trim=60mm 0mm 10mm 0mm,clip]{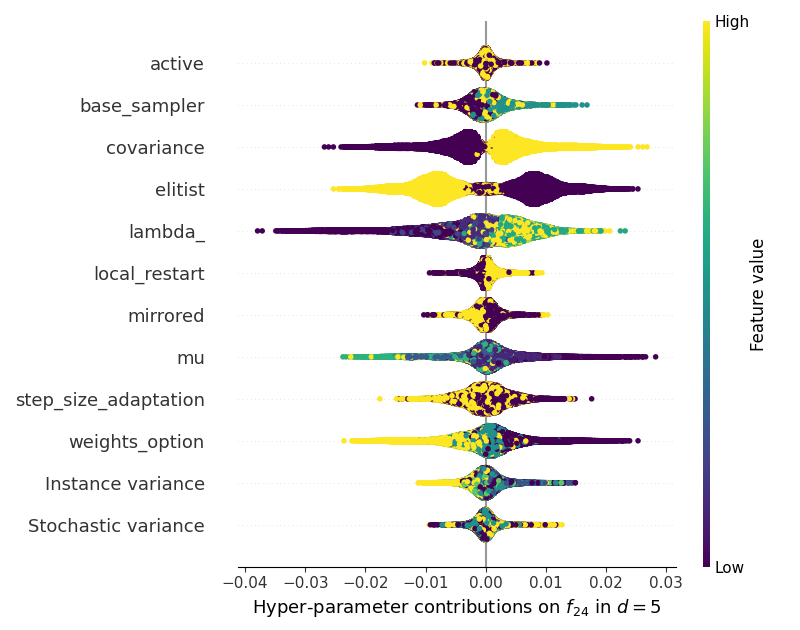}
\caption{
Hyper-parameter contributions per benchmark function for \textbf{d=5} for \textbf{modular CMA-ES}. Categorical hyperparameters are encoded to integer values in alphabetic order where NaN is encoded as $-1$. Refer to Table~\ref{tab:modde_modules} for the colour coding used.
\label{fig:shapxplaind5}}
\end{figure*}

\begin{figure*}[t]
\centering
	\includegraphics[height=0.15\textheight,trim=4mm 0mm 30mm 0mm,clip]{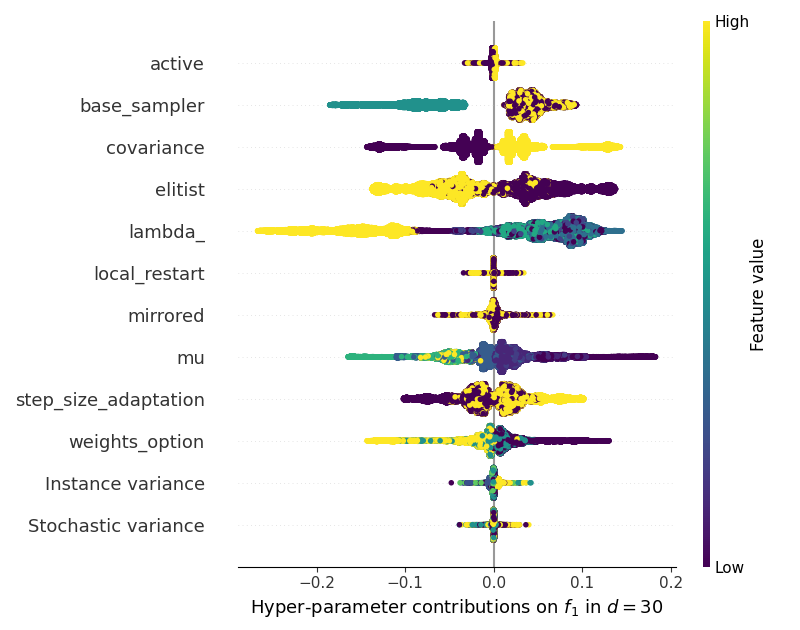}
	\includegraphics[height=0.15\textheight,trim=60mm 0mm 30mm 0mm,clip]{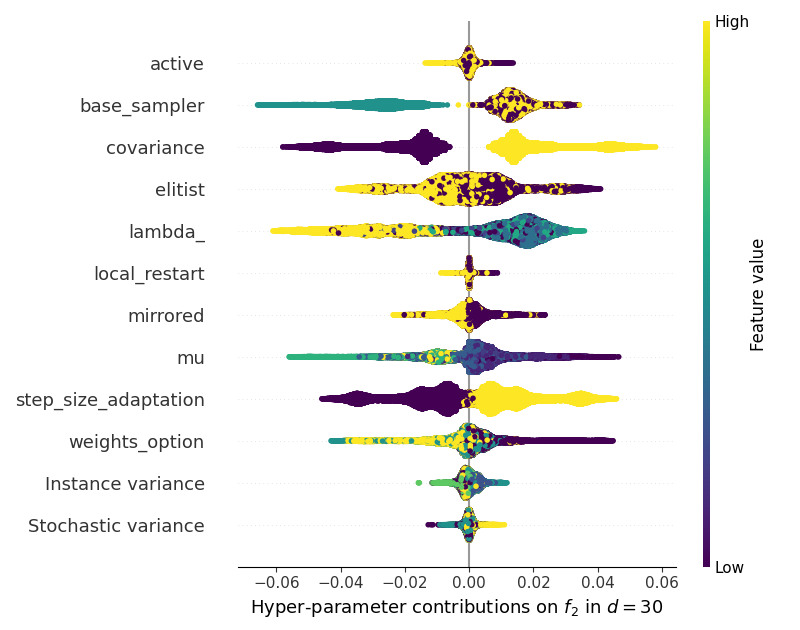}
	\includegraphics[height=0.15\textheight,trim=60mm 0mm 30mm 0mm,clip]{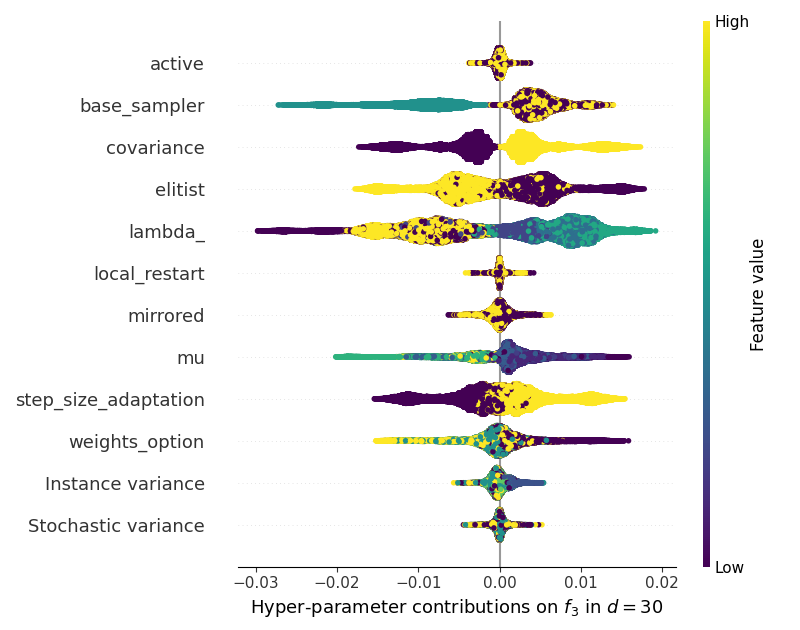}
	\includegraphics[height=0.15\textheight,trim=60mm 0mm 30mm 0mm,clip]{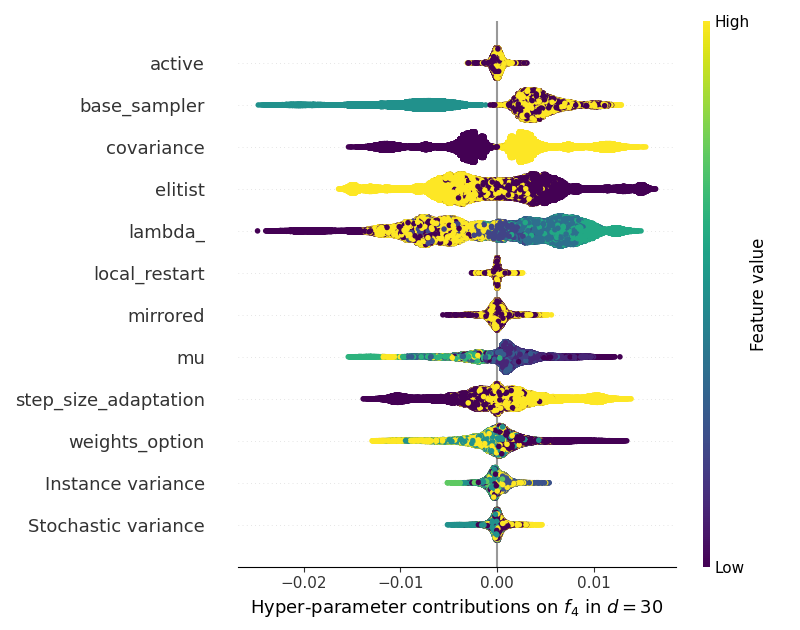}
	\includegraphics[height=0.15\textheight,trim=60mm 0mm 30mm 0mm,clip]{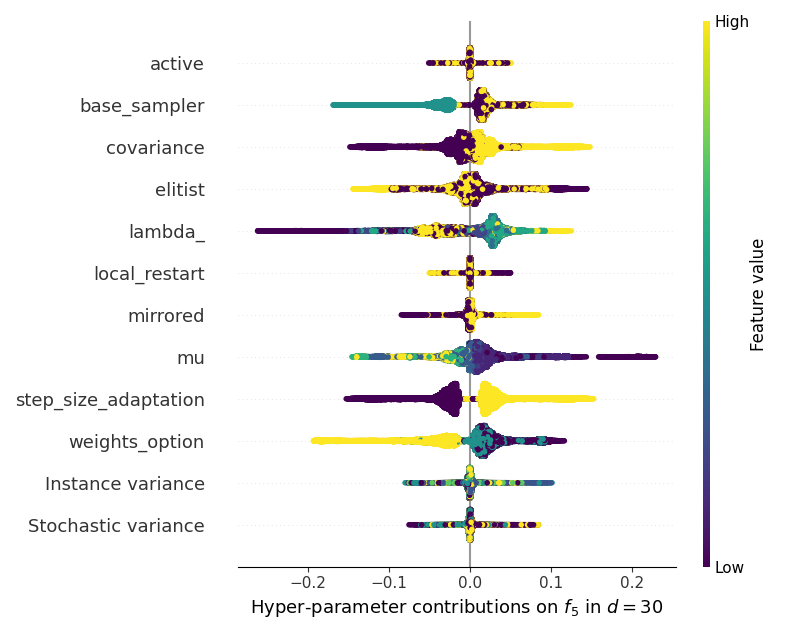}
	\includegraphics[height=0.15\textheight,trim=60mm 0mm 10mm 0mm,clip]{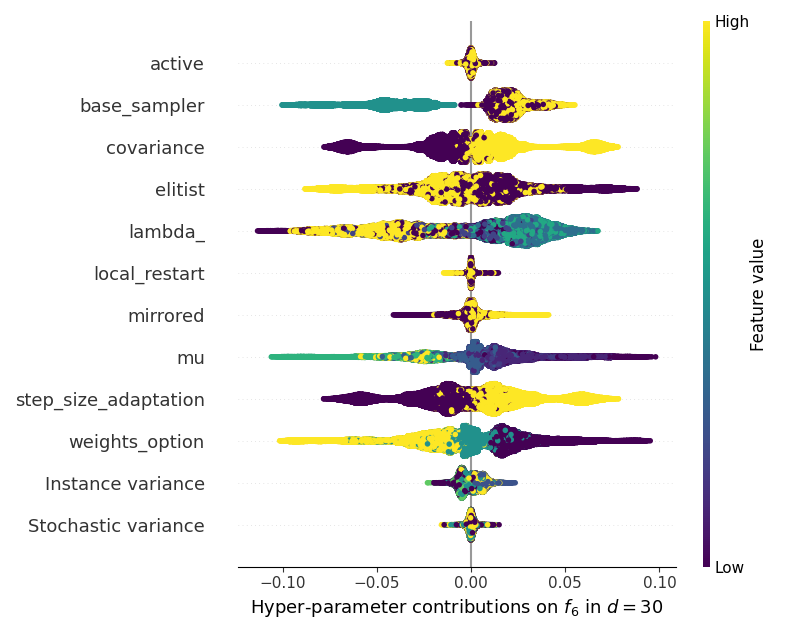}
 
	\includegraphics[height=0.15\textheight,trim=4mm 0mm 30mm 0mm,clip]{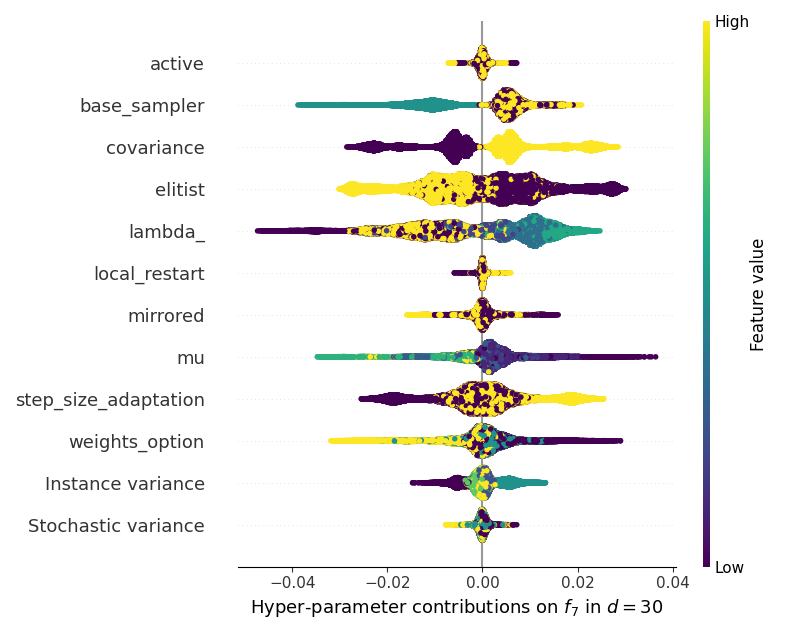}
	\includegraphics[height=0.15\textheight,trim=60mm 0mm 30mm 0mm,clip]{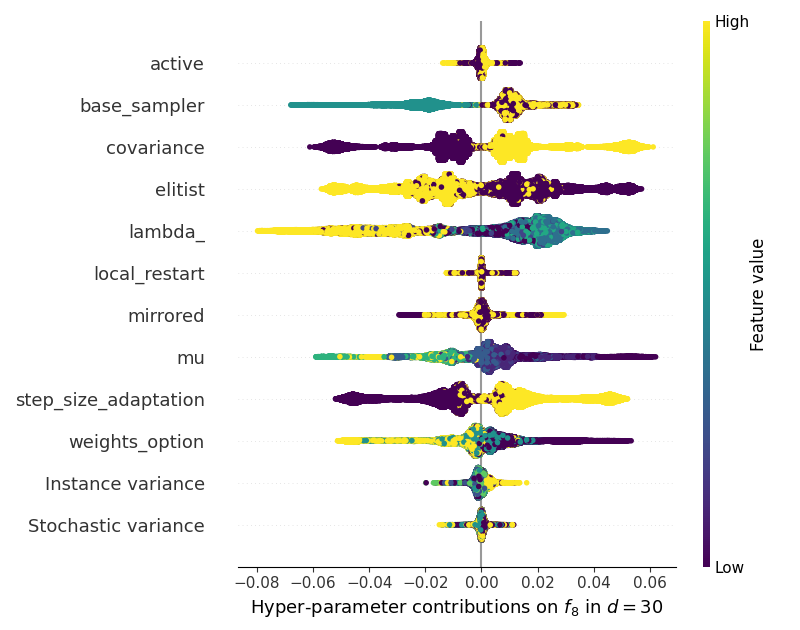}
	\includegraphics[height=0.15\textheight,trim=60mm 0mm 30mm 0mm,clip]{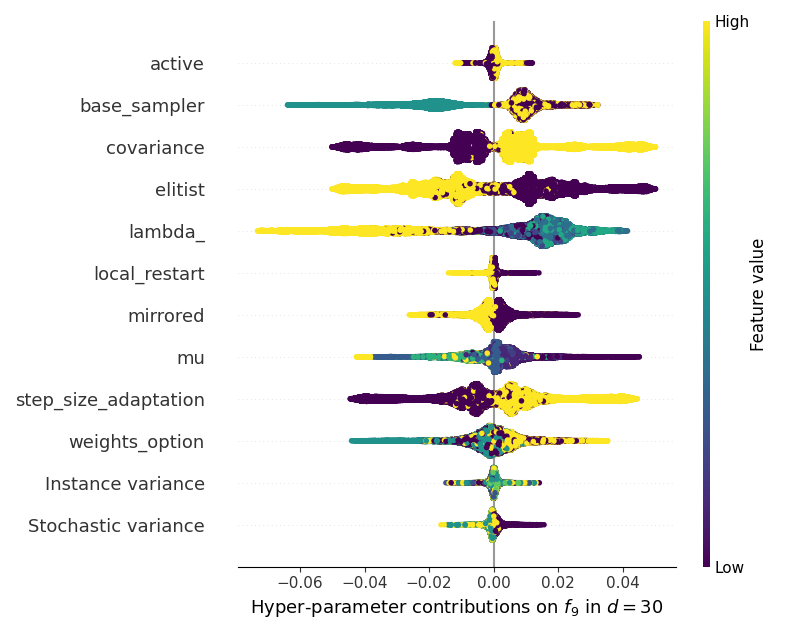}
	\includegraphics[height=0.15\textheight,trim=60mm 0mm 30mm 0mm,clip]{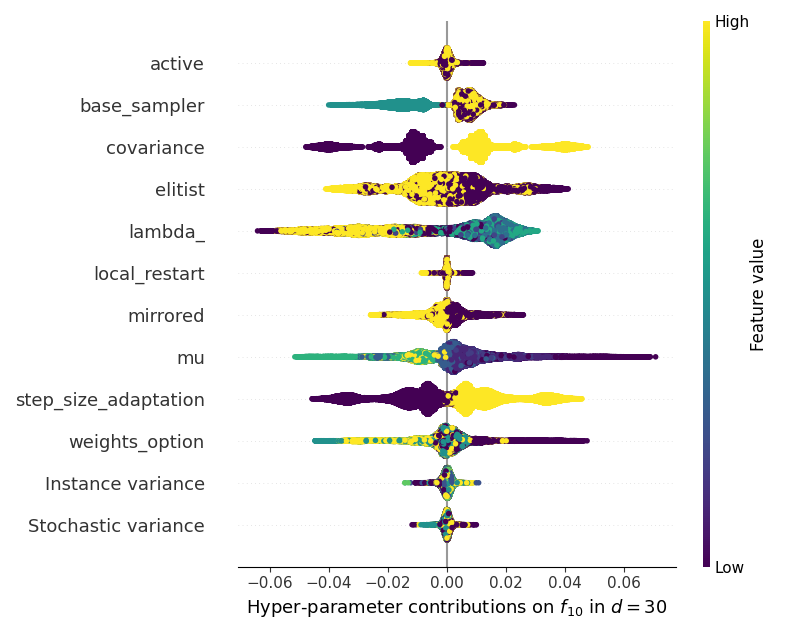}
	\includegraphics[height=0.15\textheight,trim=60mm 0mm 30mm 0mm,clip]{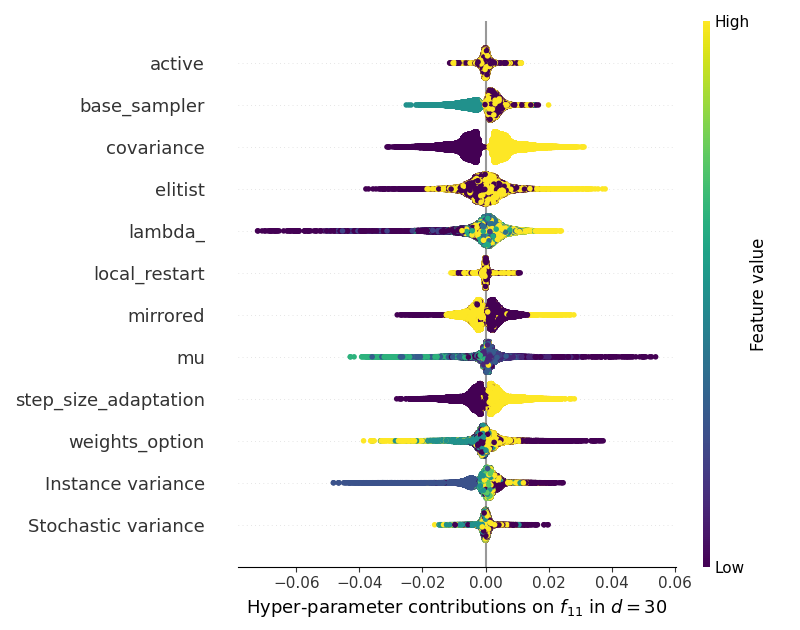}
	\includegraphics[height=0.15\textheight,trim=60mm 0mm 10mm 0mm,clip]{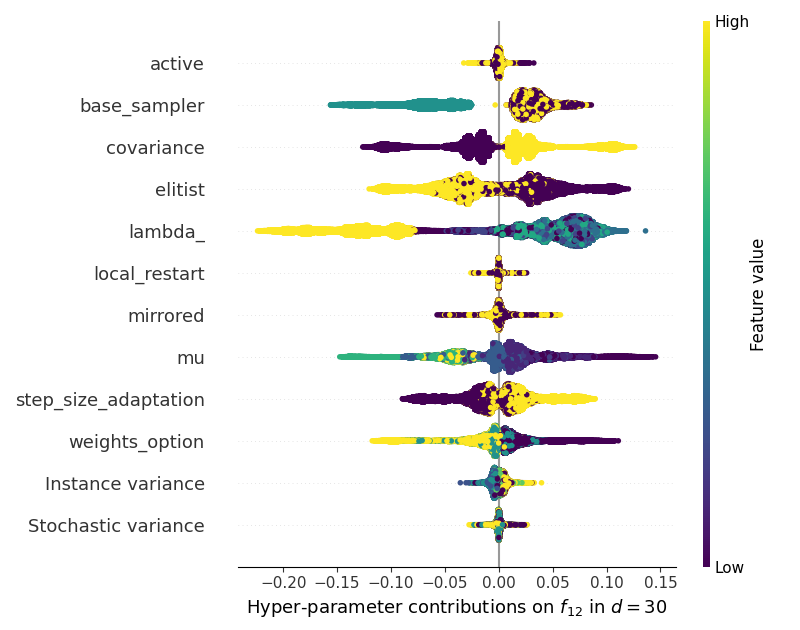}
 
	\includegraphics[height=0.15\textheight,trim=4mm 0mm 30mm 0mm,clip]{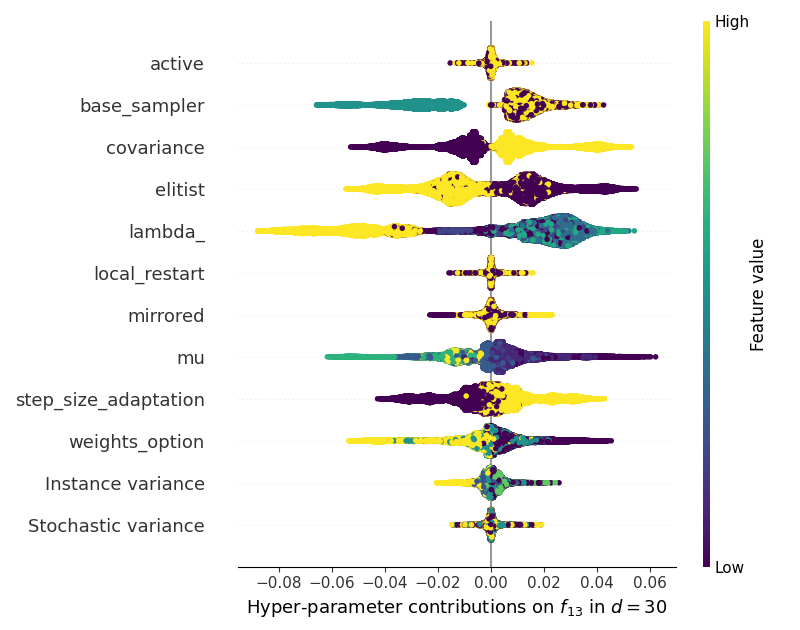}
	\includegraphics[height=0.15\textheight,trim=60mm 0mm 30mm 0mm,clip]{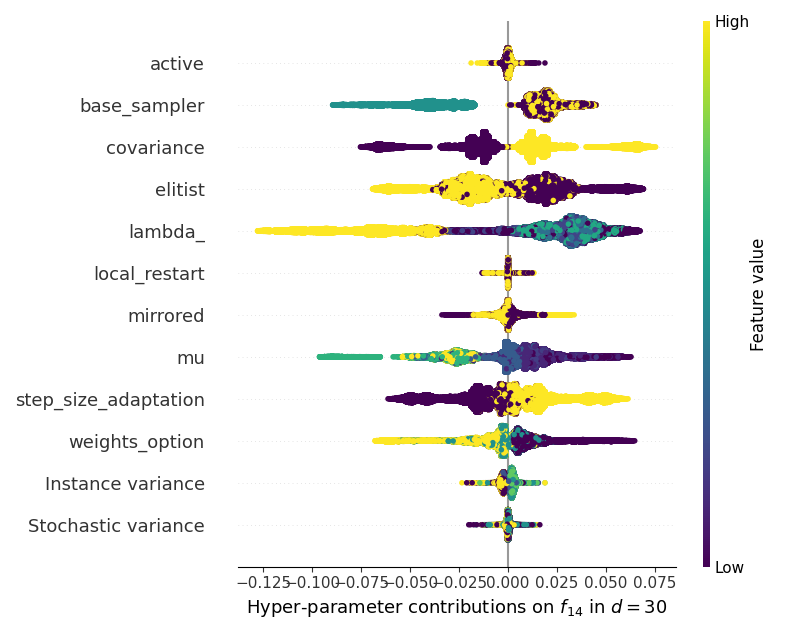}
	\includegraphics[height=0.15\textheight,trim=60mm 0mm 30mm 0mm,clip]{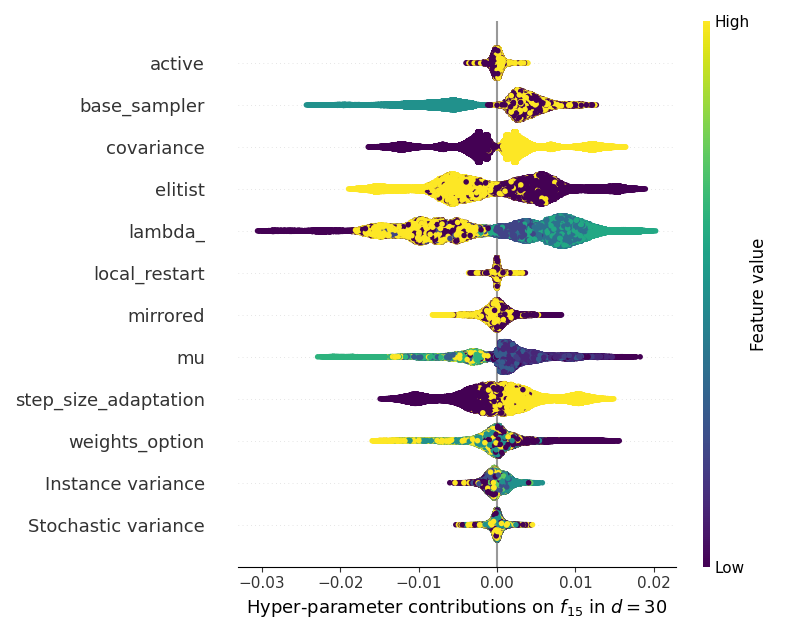}
	\includegraphics[height=0.15\textheight,trim=60mm 0mm 30mm 0mm,clip]{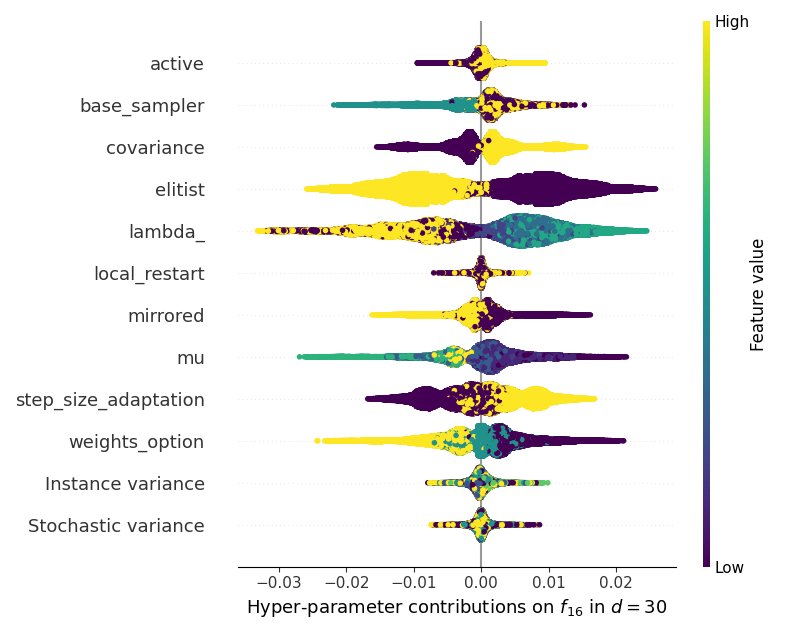}
	\includegraphics[height=0.15\textheight,trim=60mm 0mm 30mm 0mm,clip]{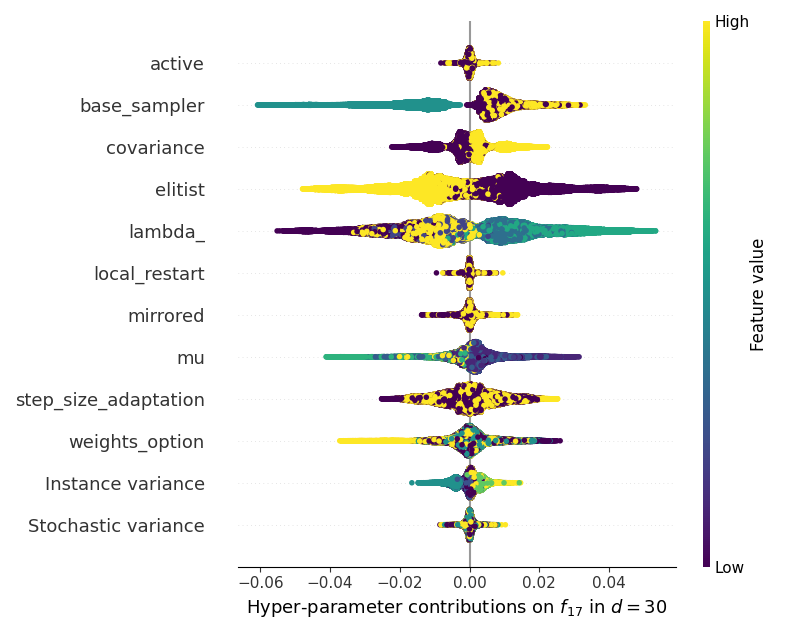}
	\includegraphics[height=0.15\textheight,trim=60mm 0mm 10mm 0mm,clip]{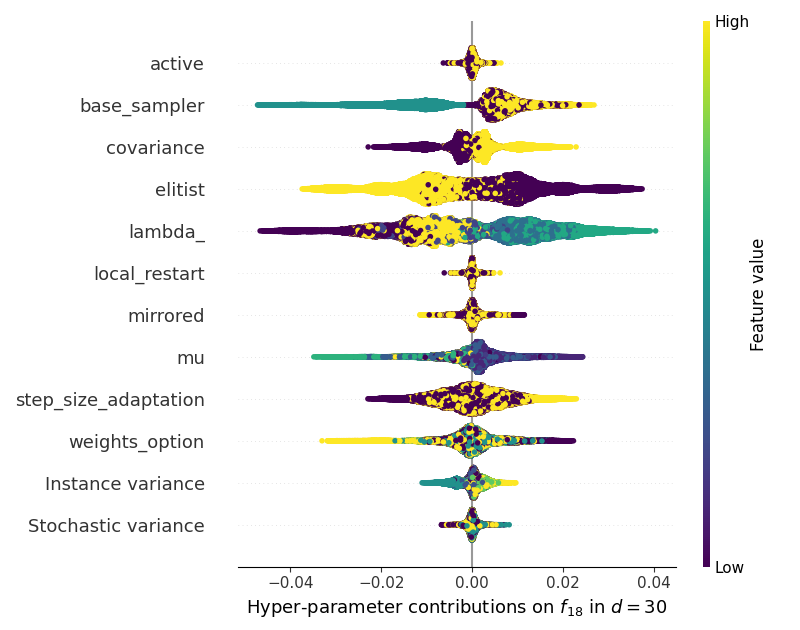}
 
	\includegraphics[height=0.15\textheight,trim=4mm 0mm 30mm 0mm,clip]{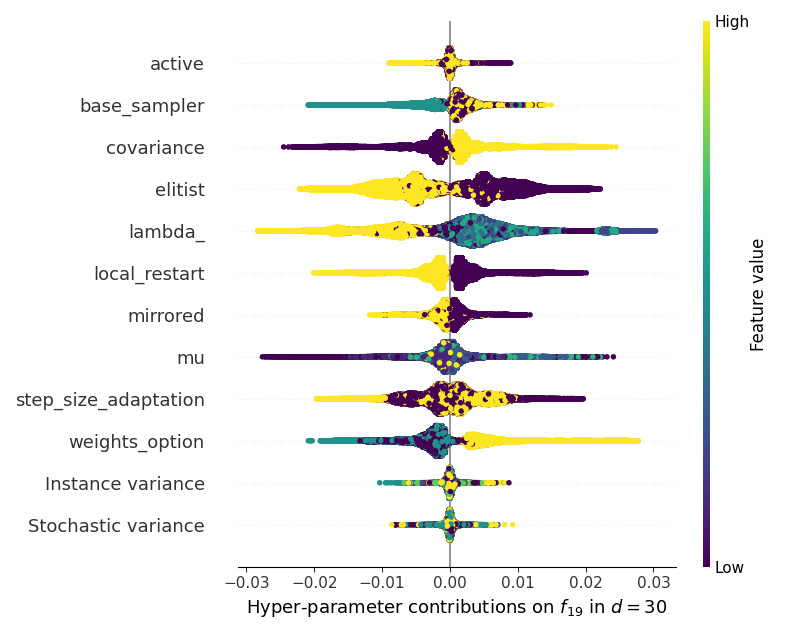}
	\includegraphics[height=0.15\textheight,trim=60mm 0mm 30mm 0mm,clip]{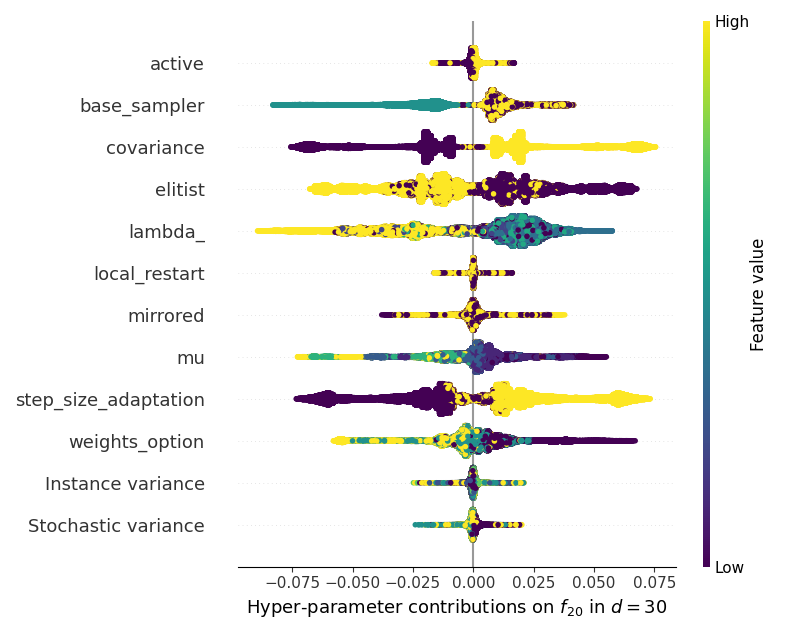}
	\includegraphics[height=0.15\textheight,trim=60mm 0mm 30mm 0mm,clip]{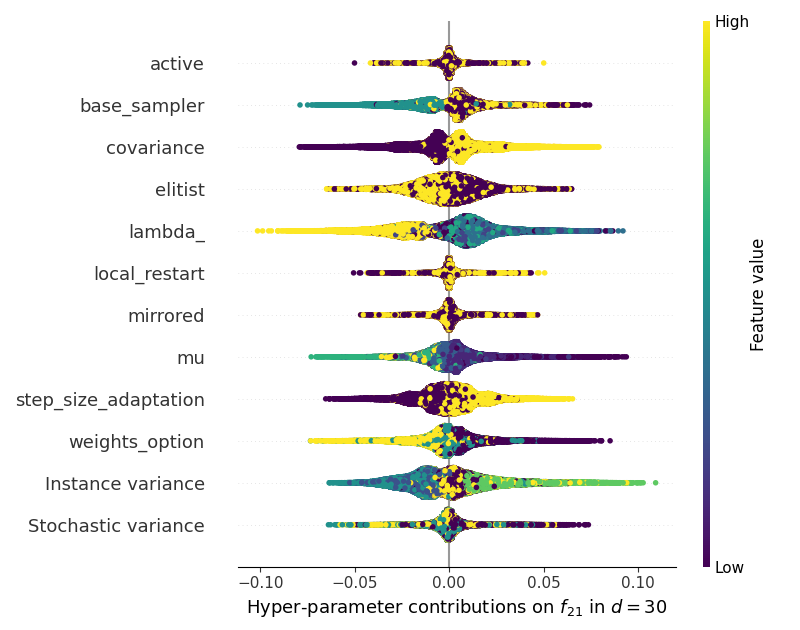}
	\includegraphics[height=0.15\textheight,trim=60mm 0mm 30mm 0mm,clip]{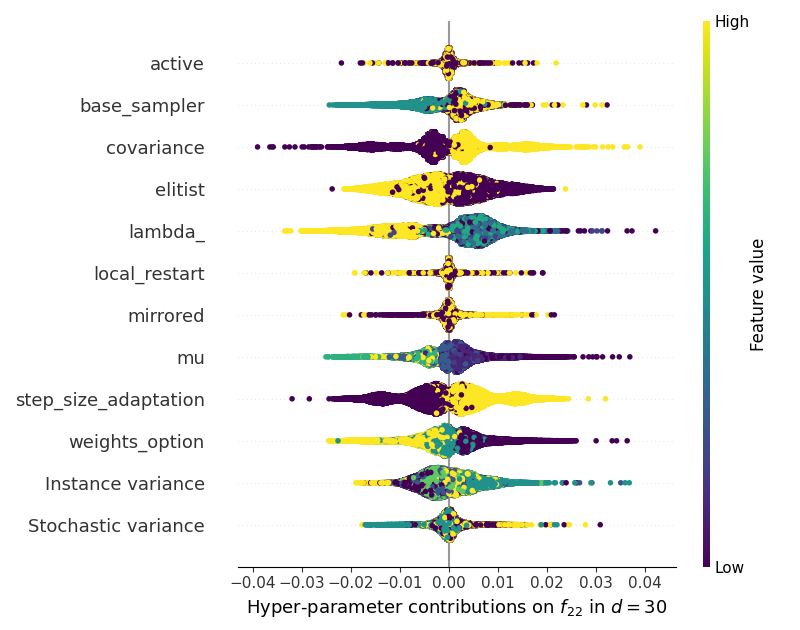}
	\includegraphics[height=0.15\textheight,trim=60mm 0mm 30mm 0mm,clip]{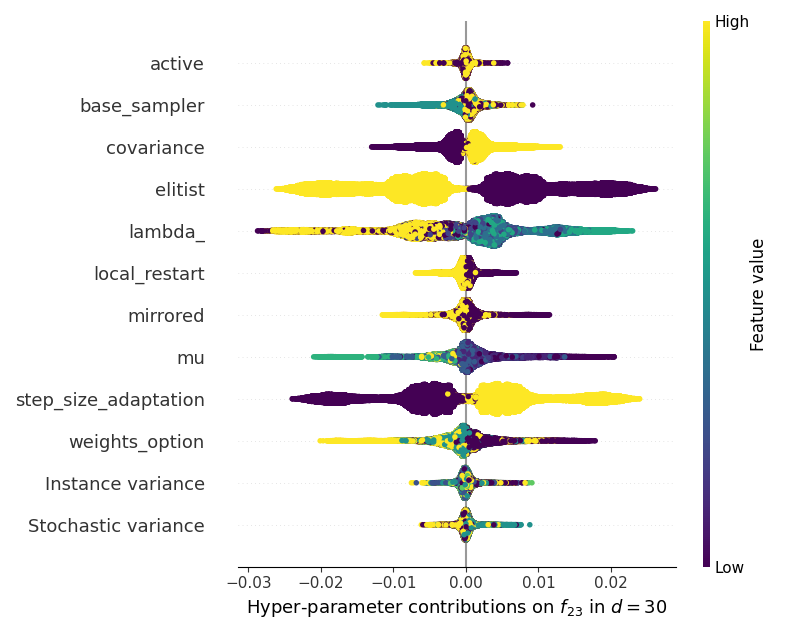}
	\includegraphics[height=0.15\textheight,trim=60mm 0mm 10mm 0mm,clip]{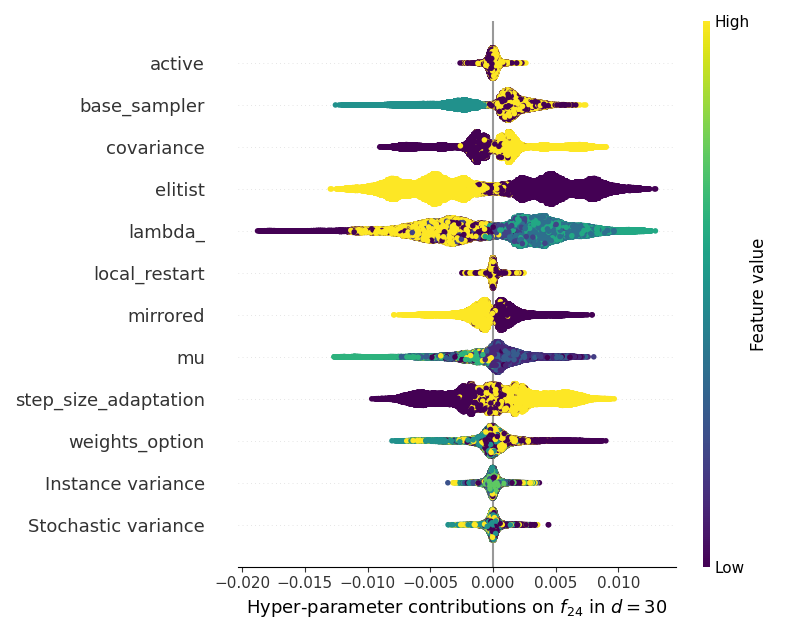}
\caption{Hyper-parameter contributions per benchmark function for \textbf{d=30} for \textbf{modular CMA-ES}. Categorical hyperparameters are encoded to integer values in alphabetic order where NaN is encoded as $-1$. Refer to Table~\ref{tab:modde_modules} for the colour coding used.
\label{fig:shapxplaind30}}
\end{figure*}

We start our analysis with the $15\,840$ configurations of the modular CMA-ES, where we first look at the contribution of each module to the performance on each of the 24 BBOB functions. Figures \ref{fig:shapxplaind5} and \ref{fig:shapxplaind30} show the resulting SHAP-based plots for dimensionalities 5 and 30, respectively. In these figures, we see a cell for each BBOB problem, within which the contribution to the predictive model (SHAP-value) of each module is indicated. Since the model is intentionally overfitted ($R^2$ score of $0.99$ ), the x-value of each dot can be interpreted as how much the performance (AOCC) is impacted by setting the specified module to the option indicated by the dot's color. 

For several modules, we see a consistent separation between the two colors, such as the `covariance' option, where violet dots consistently have negative SHAP values and yellow dots are positive. This indicates that turning off the covariance adaptation mechanism, thus transforming the algorithm into a version of MA-ES~\cite{beyer2017simplify}, leads to worse anytime performance reflected by AOCC. Similarly, many functions have a clear separation between the two options of the `elitist' module. This separation is not always in the same direction, so even though the contribution of this module is consistent on a per-function level, it is not immediately obvious whether elitism is useful for CMA-ES in general.

For non-binary modules, we also see some interesting patterns in contribution. For instance, the population size $\lambda$ shows a gradient going from high to low on most unimodal functions such as F1, while this trend is reversed on some of the more multimodal problems, e.g. F21. This pattern also occurs for the `local\_restart' parameter, which has little impact on the functions where low population sizes are optimal, but growing impact as population sizes become more important to achieve good performance~\cite{hansen2009benchmarking, auger2005restart}. 

In both Figure~\ref{fig:shapxplaind5} and \ref{fig:shapxplaind30}, it is important to point out the differences in the scale of the x-axis. For the `easier' functions, e.g. F2, the normalized AOCC can vary by as much as $0.4$, while for other functions, e.g. F24, the overall differences in performance between configurations are much smaller. While it might be natural to assume that for these harder functions, noise effects matter more than module choices, the relative contribution from the stochastic variance (i.e. the seed-based variance) is still quite small and trends in, e.g., `elitist' and `weight' options are rather consistent. We can thus conclude that even for cases where the total performance variation is low, our methodology can still clearly identify parameter contributions.

Finally, we want to highlight some differences between the two used dimensionalities. In particular, we observe a different pattern for the `base\_sampler' between these settings. This might be caused by an instability in the used Halton sampler for high dimensions, which leads to some bias in the direction of sampled points, giving worse optimization results. While these results provide some interesting insights into the impact these modules have on the CMA-ES performance, the precise contributions should be considered within the context of our experiments. Specifically, we should note that the used budget of $10\,000$ evaluations is relatively small for 30-dimensional functions, which specifically impacts modules such as the local restart. With this budget, the restart is unlikely to be performed very often, which might undervalue its importance when larger evaluation budgets are available.   



\subsubsection{Algorithm/configuration selection}
\begin{table*}
\centering
\caption{Performance (AOCC) of single-best, average-best and overall average algorithm performance for \textbf{modular CMA-ES} over all configurations per function and dimension. \textbf{Boldface} for the single-best configuration indicates a significant improvement over the average best configuration (for that dimension), \textbf{Boldface} for the average best configuration indicates a significant improvement ($p$ value $< 0.05$) over the average AOCC of all configurations. The standard deviation is denoted between brackets. \label{tab:cma_configs_perf}}
\resizebox{\textwidth}{!}{\begin{tabular}{llllllll}
\toprule
& \multicolumn{3}{c}{$d=5$} & \multicolumn{3}{c}{$d=30$} \\
Function & \multicolumn{1}{c}{single-best} & \multicolumn{1}{c}{avg-best} & \multicolumn{1}{c}{all} & \multicolumn{1}{c}{single-best} & \multicolumn{1}{c}{avg-best} & \multicolumn{1}{c}{all} \\
\midrule
f1 Sphere & \textbf{0.98 (0.00)} & \textbf{0.97 (0.00)} & 0.69 (0.30) & \textbf{0.95 (0.00)} & \textbf{0.90 (0.00)} & 0.62 (0.21) \\
f2 Ellipsoid & \textbf{0.91 (0.01)} & \textbf{0.90 (0.01)} & 0.39 (0.36) & \textbf{0.29 (0.01)} & \textbf{0.29 (0.01)} & 0.18 (0.06) \\
f3 Rastrigin & 0.38 (0.29) & \textbf{0.30 (0.23)} & 0.12 (0.08) & \textbf{0.40 (0.00)} & \textbf{0.39 (0.01)} & 0.35 (0.02) \\
f4 BuecheRastrigin & 0.17 (0.16) & \textbf{0.13 (0.02)} & 0.09 (0.04) & \textbf{0.38 (0.01)} & \textbf{0.38 (0.01)} & 0.34 (0.02) \\
f5 LinearSlope & \textbf{1.00 (0.00)} & 0.99 (0.00) & 0.95 (0.16) & \textbf{0.99 (0.00)} & \textbf{0.98 (0.00)} & 0.89 (0.17) \\
f6 AttractiveSector & \textbf{0.95 (0.01)} & \textbf{0.90 (0.01)} & 0.48 (0.34) & 0.58 (0.01) & \textbf{0.58 (0.03)} & 0.35 (0.11) \\
f7 StepEllipsoid & 0.95 (0.00) & \textbf{0.94 (0.03)} & 0.50 (0.31) & 0.45 (0.01) & \textbf{0.45 (0.01)} & 0.38 (0.04) \\
f8 Rosenbrock & \textbf{0.91 (0.02)} & \textbf{0.85 (0.02)} & 0.42 (0.34) & 0.41 (0.01) & \textbf{0.40 (0.01)} & 0.32 (0.08) \\
f9 RosenbrockRotated & \textbf{0.92 (0.01)} & \textbf{0.86 (0.01)} & 0.45 (0.33) & \textbf{0.41 (0.00)} & \textbf{0.41 (0.00)} & 0.34 (0.06) \\
f10 EllipsoidRotated & \textbf{0.90 (0.01)} & \textbf{0.89 (0.01)} & 0.39 (0.36) & 0.29 (0.01) & \textbf{0.28 (0.01)} & 0.19 (0.06) \\
f11 Discus & \textbf{0.92 (0.01)} & \textbf{0.91 (0.01)} & 0.37 (0.32) & 0.41 (0.01) & \textbf{0.40 (0.01)} & 0.35 (0.02) \\
f12 BentCigar & \textbf{0.86 (0.07)} & \textbf{0.73 (0.23)} & 0.33 (0.33) & 0.51 (0.09) & \textbf{0.45 (0.06)} & 0.21 (0.17) \\
f13 SharpRidge & \textbf{0.90 (0.01)} & \textbf{0.87 (0.01)} & 0.37 (0.31) & 0.51 (0.05) & \textbf{0.49 (0.05)} & 0.38 (0.07) \\
f14 DifferentPowers & \textbf{0.95 (0.01)} & \textbf{0.95 (0.00)} & 0.62 (0.27) & \textbf{0.70 (0.01)} & \textbf{0.69 (0.00)} & 0.54 (0.10) \\
f15 RastriginRotated & 0.48 (0.02) & \textbf{0.43 (0.19)} & 0.13 (0.10) & \textbf{0.40 (0.01)} & \textbf{0.39 (0.01)} & 0.35 (0.02) \\
f16 Weierstrass & 0.81 (0.16) & \textbf{0.69 (0.19)} & 0.26 (0.19) & 0.50 (0.01) & \textbf{0.50 (0.01)} & 0.43 (0.03) \\
f17 Schaffers10 & \textbf{0.90 (0.02)} & \textbf{0.87 (0.05)} & 0.41 (0.21) & \textbf{0.64 (0.02)} & \textbf{0.60 (0.03)} & 0.48 (0.04) \\
f18 Schaffers1000 & \textbf{0.82 (0.12)} & \textbf{0.68 (0.16)} & 0.30 (0.16) & \textbf{0.55 (0.01)} & \textbf{0.52 (0.02)} & 0.44 (0.04) \\
f19 GriewankRosenbrock & 0.40 (0.23) & \textbf{0.31 (0.02)} & 0.24 (0.06) & \textbf{0.55 (0.00)} & \textbf{0.48 (0.01)} & 0.46 (0.02) \\
f20 Schwefel & \textbf{0.36 (0.11)} & \textbf{0.21 (0.07)} & 0.19 (0.05) & \textbf{0.48 (0.00)} & \textbf{0.47 (0.00)} & 0.40 (0.09) \\
f21 Gallagher101 & \textbf{0.84 (0.23)} & 0.53 (0.32) & 0.40 (0.31) & 0.66 (0.22) & \textbf{0.56 (0.20)} & 0.45 (0.10) \\
f22 Gallagher21 & \textbf{0.79 (0.26)} & 0.46 (0.33) & 0.36 (0.30) & 0.48 (0.11) & 0.44 (0.01) & 0.42 (0.03) \\
f23 Katsuura & \textbf{0.66 (0.22)} & \textbf{0.49 (0.20)} & 0.21 (0.10) & \textbf{0.55 (0.01)} & \textbf{0.49 (0.01)} & 0.47 (0.03) \\
f24 LunacekBiRastrigin & 0.14 (0.03) & \textbf{0.13 (0.02)} & 0.09 (0.02) & \textbf{0.38 (0.00)} & \textbf{0.38 (0.00)} & 0.35 (0.01) \\
\bottomrule
\end{tabular}}
\end{table*}

\begin{table*}[!b]
\centering
\caption{Performance summary of \textbf{modular CMA-ES}. The gain avg-best is the average improvement of the average-best configuration over all configurations. The gain single-best is the average improvement of the single-best configurations (per function) over all configurations. \textbf{Boldface} in this table indicates a significant improvement (avg-best versus all and single-best versus avg-best).}\label{tab:cma_perf_summary}
\begin{tabular}{lrr}
\toprule
Measure & $d=5$ & $d=30$ \\
\midrule
Average performance & 0.37  & 0.40 \\
Gain avg-best & \textbf{0.30} & \textbf{0.09} \\
Gain single-best & \textbf{0.38} & 0.12 \\
\bottomrule
\end{tabular}
\end{table*}

As observed from the different patterns in Figures~\ref{fig:shapxplaind5} and \ref{fig:shapxplaind30}, modules show different contribution on different functions. While this means that there is no perfect setting of the modular CMA-ES, it also indicates that an algorithm selector has a lot of potential. To investigate just how large this potential is, we identify the best configuration for each BBOB function (`single-best') and the best across all functions (`avg-best'). The performance for both of these, as well as the average performance over all configurations (`all') is shown in Table~\ref{tab:cma_configs_perf}. In Table~\ref{tab:cma_perf_summary}, these performances are aggregated to show the improvement gained by selecting the overall best configuration over the whole set, as well as the improvement which could be gained from a perfect algorithm selector. This table indicates that using the avg-best in modCMA rather than an arbitrary combination would lead to an AOCC improvement of between $0.09$ and $0.30$ for 30 and 5 dimensional BBOB, respectively. With a perfect algorithm selector (operating on settings from Table~\ref{tab:modcma_modules} ), this performance would again be improved, with $0.09$ for 5d and $0.03$ for the 30d setting.

\include{hallOfFame/cma_es-hall-of-fame}

The exact settings for each single-best, as well as the avg-best configuration, are shown in Table~\ref{tab:fame-cma}. This table highlights that, while the contribution trend of a module might be negative in Figure~\ref{fig:shapxplaind5} or \ref{fig:shapxplaind30}, it might still be beneficial in combination with other modules, or its impact might be sufficiently low when performance is high that changing its setting does not change the performance significantly. This can be observed for example with the `covariance' parameter, which as mentioned before seems to have a positive contribution to performance when turned on. However, for 21 out of 50  settings, the best-performing configuration is an MA-ES instead of CMA-ES variant. Since Table~\ref{tab:fame-cma} also shows the impact of each module choice relative to the average of the other choices, we can isolate the performance gained from enabling or disabling the covariance adaptation in more detail. We observe that in the cases where this adaptation is turned off, the normalized AOCC would be lowered by between $0.00$ and $0.03$ in 16 out of 21 cases, while the reverse change would lead to much larger deterioration in performance. 

Other patterns, such as the impact of mirrored sampling, are more consistent between the two types of analysis. Table~\ref{tab:fame-cma} shows that all of the best configurations enable this module, with a slight preference for `mirrored' over `mirrored pairwise'. The changes in performance from the single-module changes also highlight the low impact of restart strategies on unimodal problems. From looking only at the selected modules, we might infer that `BIPOP' is needed to achieve optimal performance on F1, but since restarts are not triggered in this scenario, the performance of our selected configuration would not change if restart is changed to `IPOP', or even turned off completely. 
The choice of elitist and local restart modules seem more stable regardless of dimensionality (the choice winning on a function for 5d is often also winning for that function in 30d). To use the covariance module and different population sizes however are not stable between the two dimensionalities.

\subsection{Discussion of results on ModDE}\label{sec:modde_discussion}

\subsubsection{Contribution of hyperparameters}
\begin{figure*}[t]
\centering
	\includegraphics[height=0.15\textheight,trim=4mm 0mm 30mm 0mm,clip]{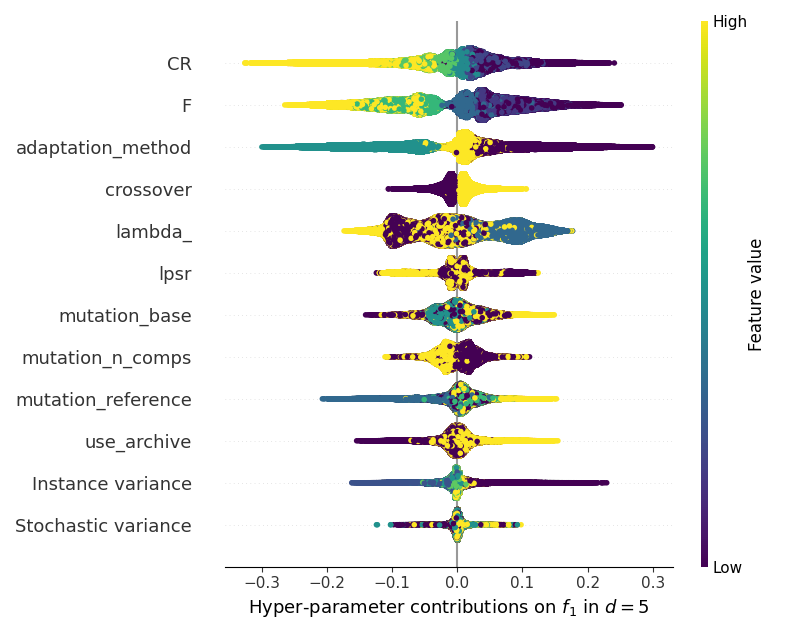} 
	\includegraphics[height=0.15\textheight,trim=60mm 0mm 30mm 0mm,clip]{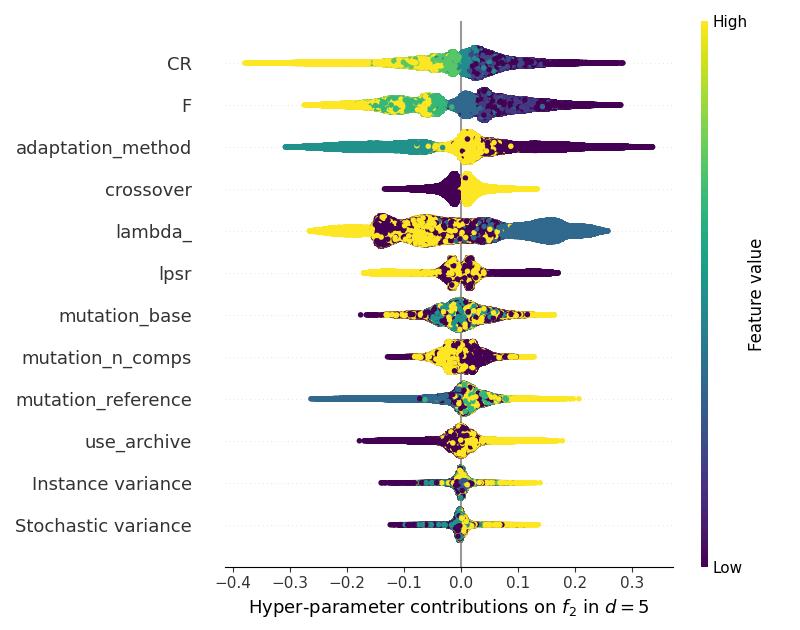}
	\includegraphics[height=0.15\textheight,trim=60mm 0mm 30mm 0mm,clip]{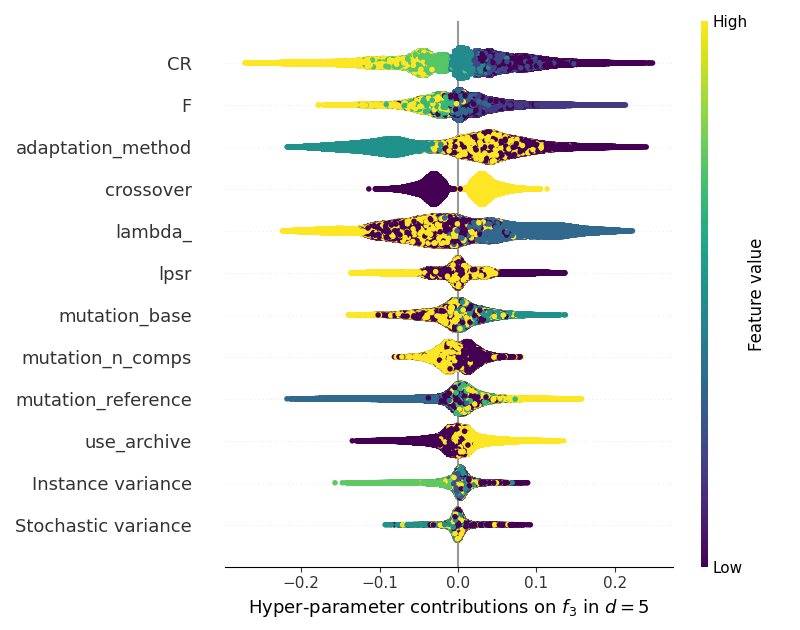}
	\includegraphics[height=0.15\textheight,trim=60mm 0mm 30mm 0mm,clip]{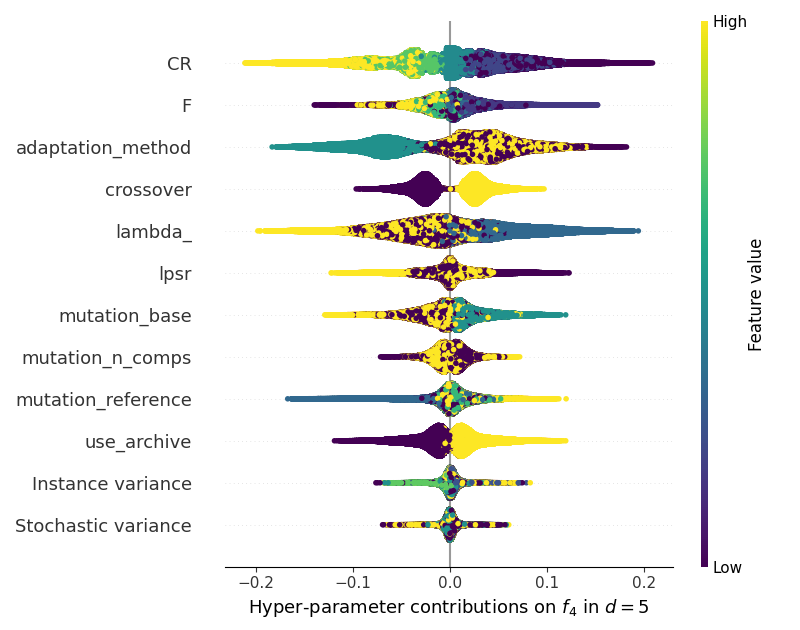}
	\includegraphics[height=0.15\textheight,trim=60mm 0mm 30mm 0mm,clip]{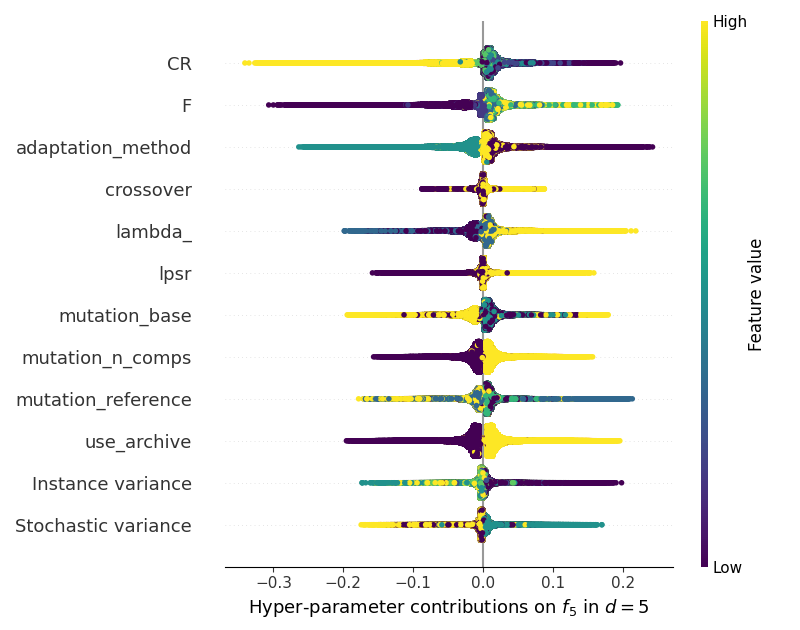}
	\includegraphics[height=0.15\textheight,trim=60mm 0mm 10mm 0mm,clip]{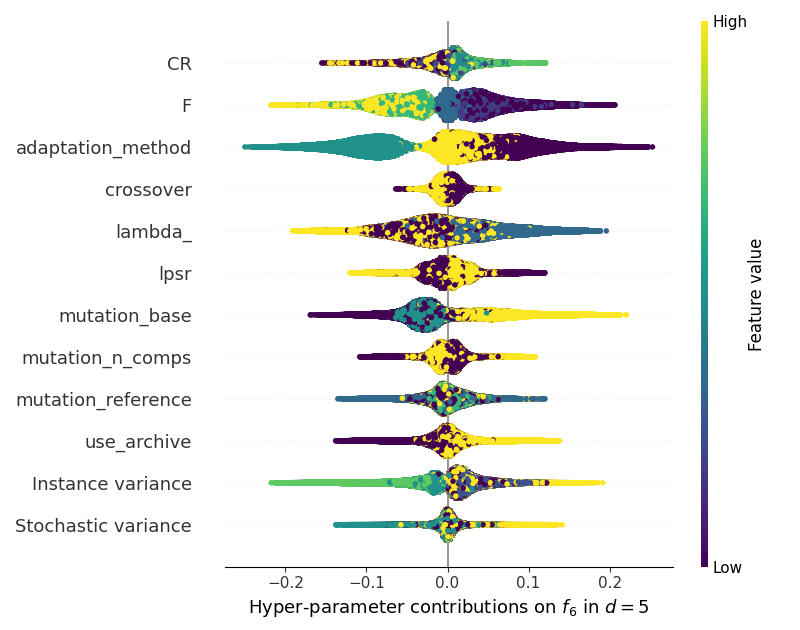}
 
	\includegraphics[height=0.15\textheight,trim=4mm 0mm 30mm 0mm,clip]{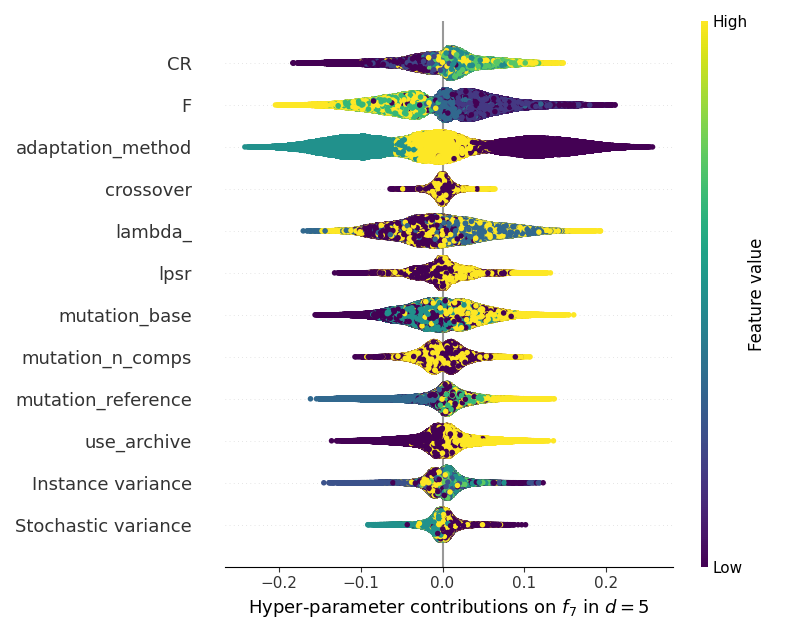}
	\includegraphics[height=0.15\textheight,trim=60mm 0mm 30mm 0mm,clip]{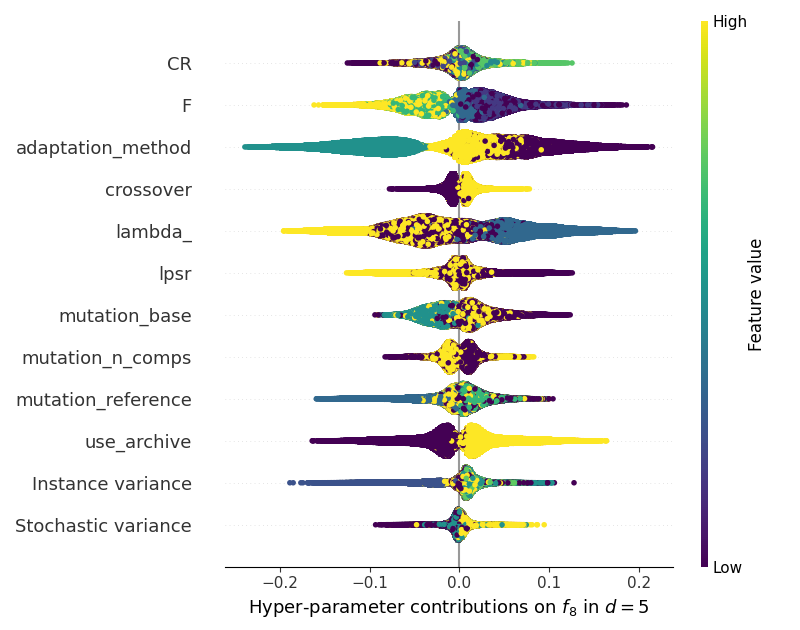}
	\includegraphics[height=0.15\textheight,trim=60mm 0mm 30mm 0mm,clip]{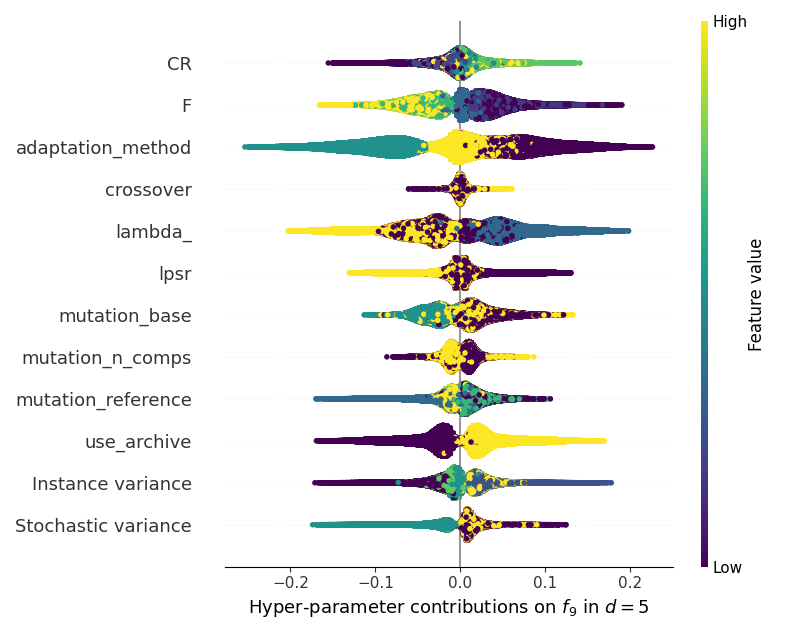}
	\includegraphics[height=0.15\textheight,trim=60mm 0mm 30mm 0mm,clip]{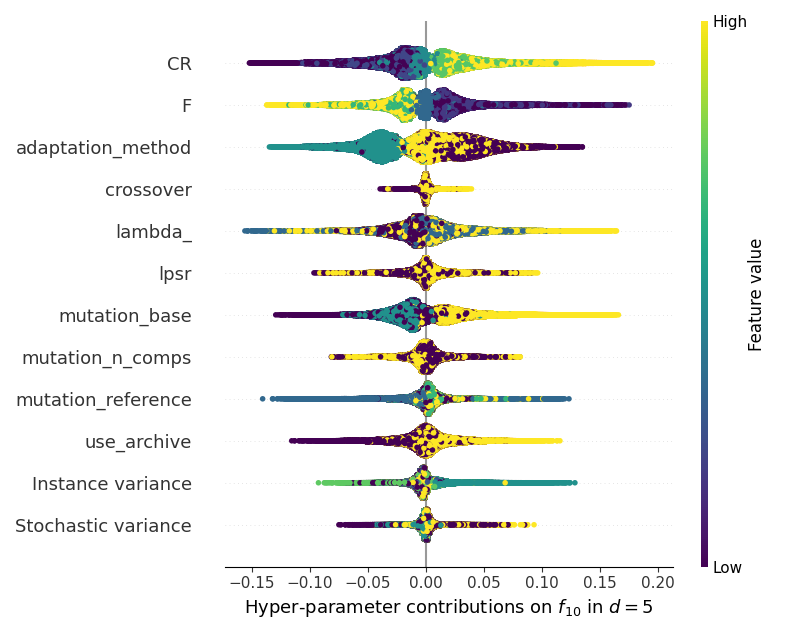}
	\includegraphics[height=0.15\textheight,trim=60mm 0mm 30mm 0mm,clip]{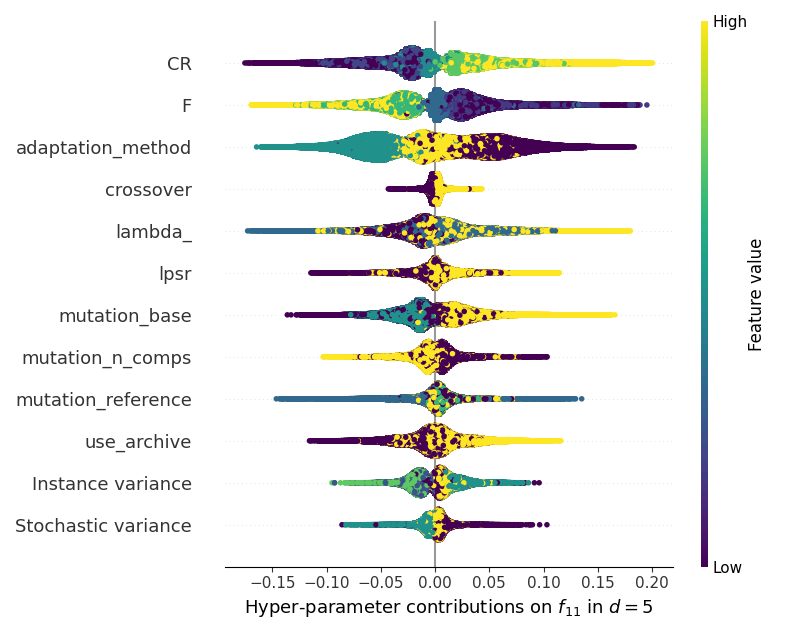}
	\includegraphics[height=0.15\textheight,trim=60mm 0mm 10mm 0mm,clip]{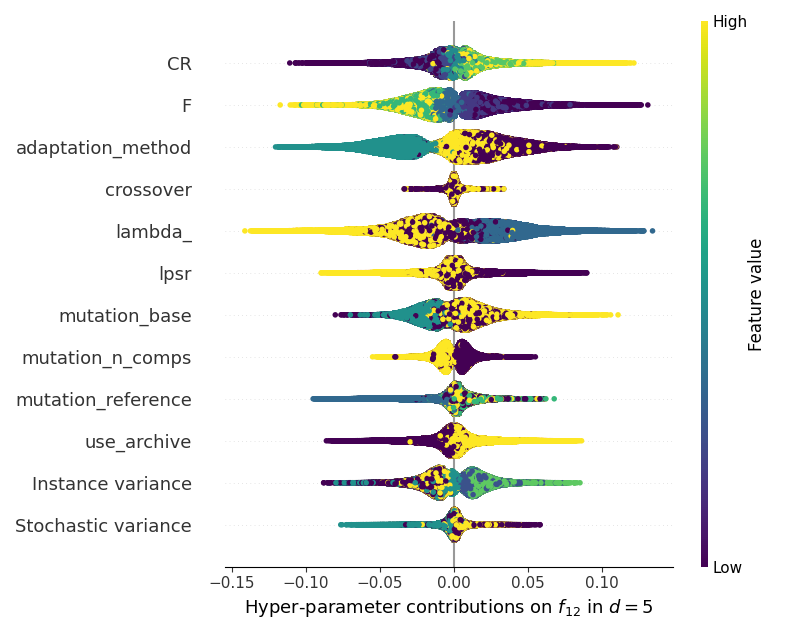}
 
	\includegraphics[height=0.15\textheight,trim=4mm 0mm 30mm 0mm,clip]{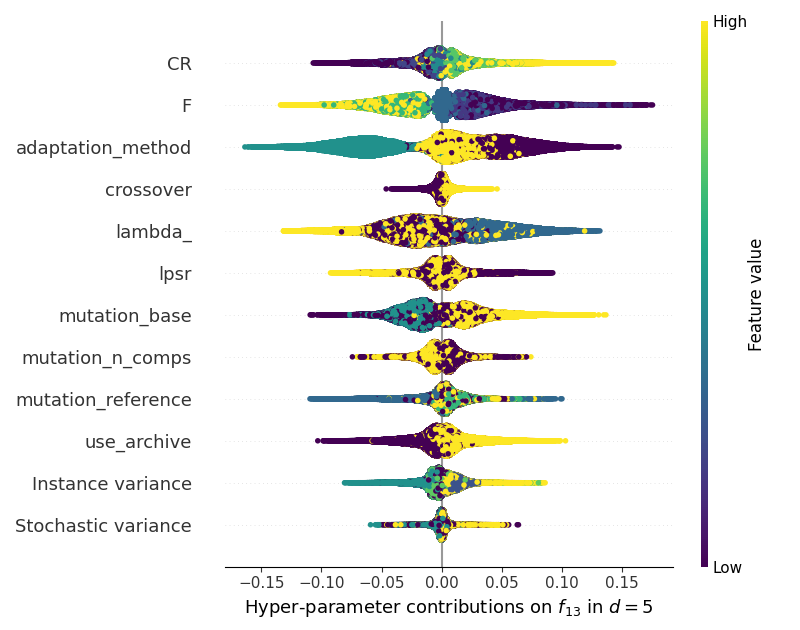}
	\includegraphics[height=0.15\textheight,trim=60mm 0mm 30mm 0mm,clip]{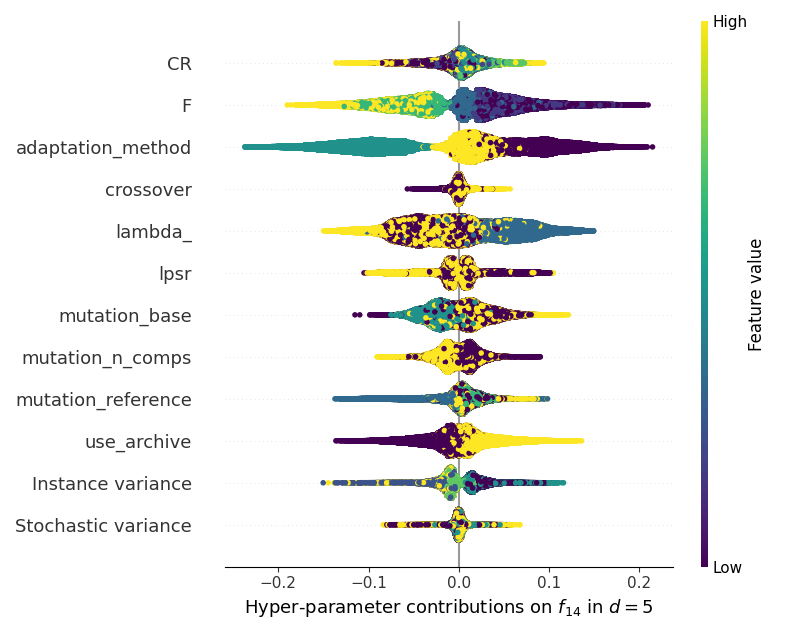}
	\includegraphics[height=0.15\textheight,trim=60mm 0mm 30mm 0mm,clip]{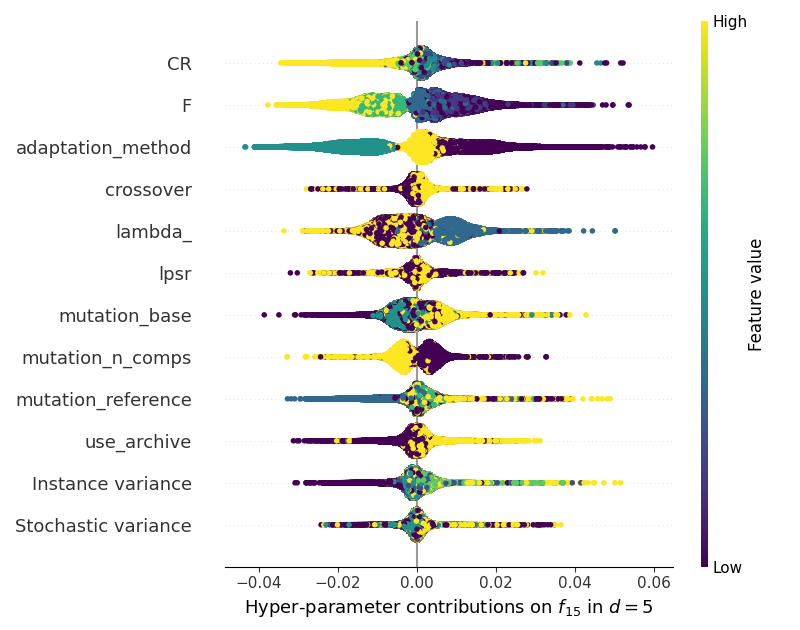}
	\includegraphics[height=0.15\textheight,trim=60mm 0mm 30mm 0mm,clip]{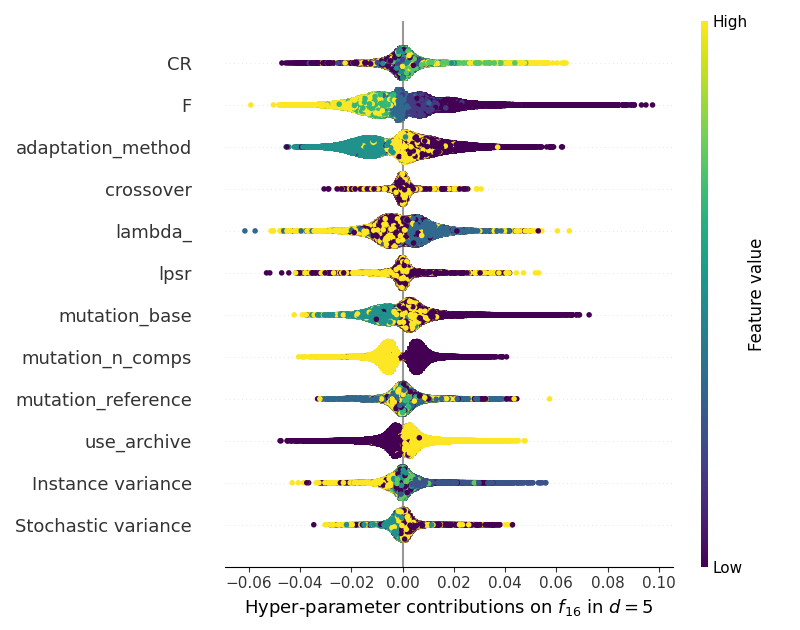}
	\includegraphics[height=0.15\textheight,trim=60mm 0mm 30mm 0mm,clip]{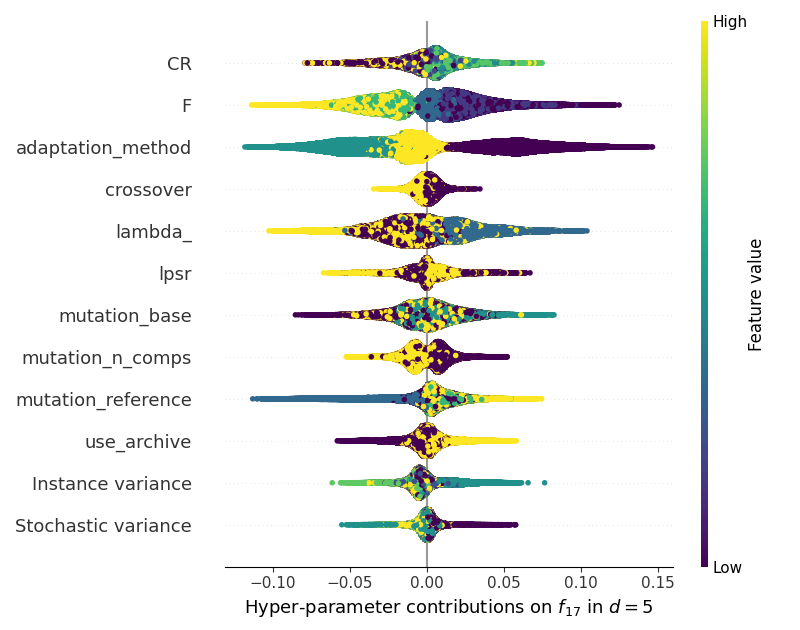}
	\includegraphics[height=0.15\textheight,trim=60mm 0mm 10mm 0mm,clip]{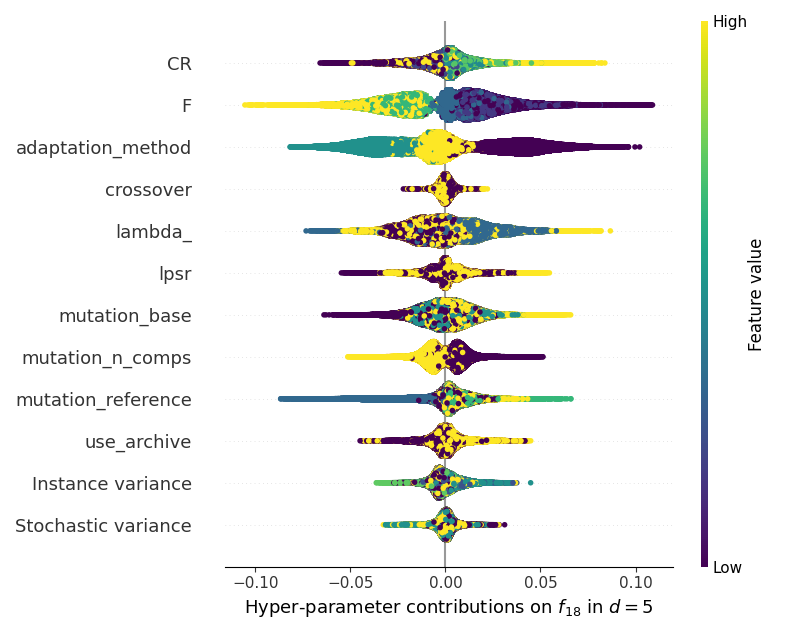}
 
	\includegraphics[height=0.15\textheight,trim=4mm 0mm 30mm 0mm,clip]{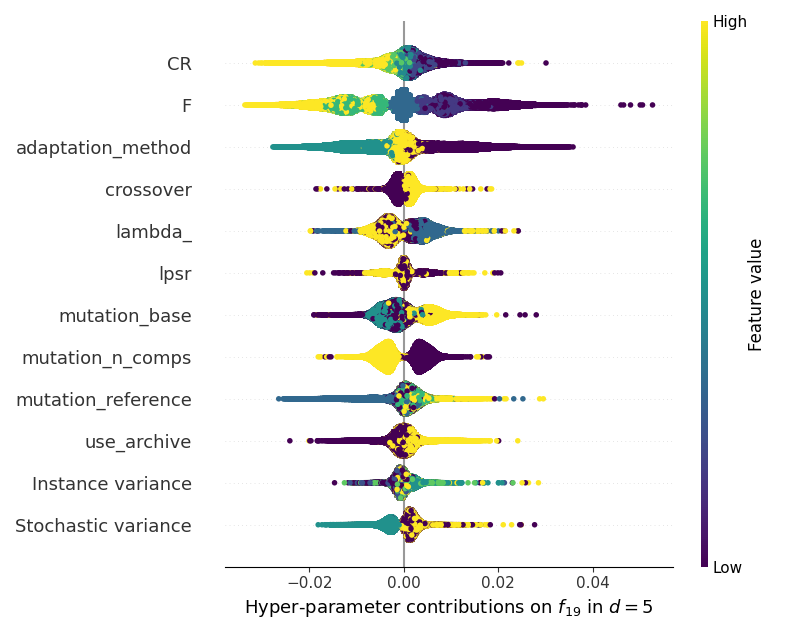}
	\includegraphics[height=0.15\textheight,trim=60mm 0mm 30mm 0mm,clip]{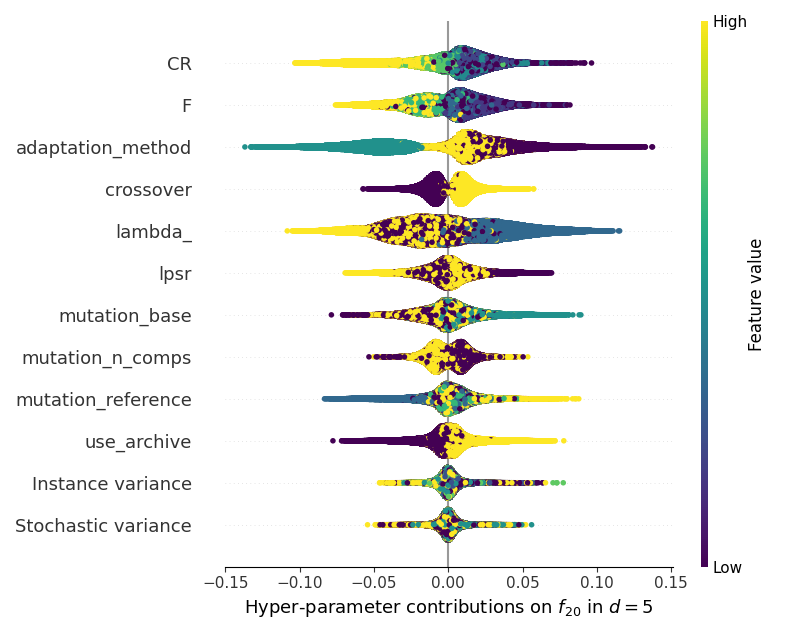}
	\includegraphics[height=0.15\textheight,trim=60mm 0mm 30mm 0mm,clip]{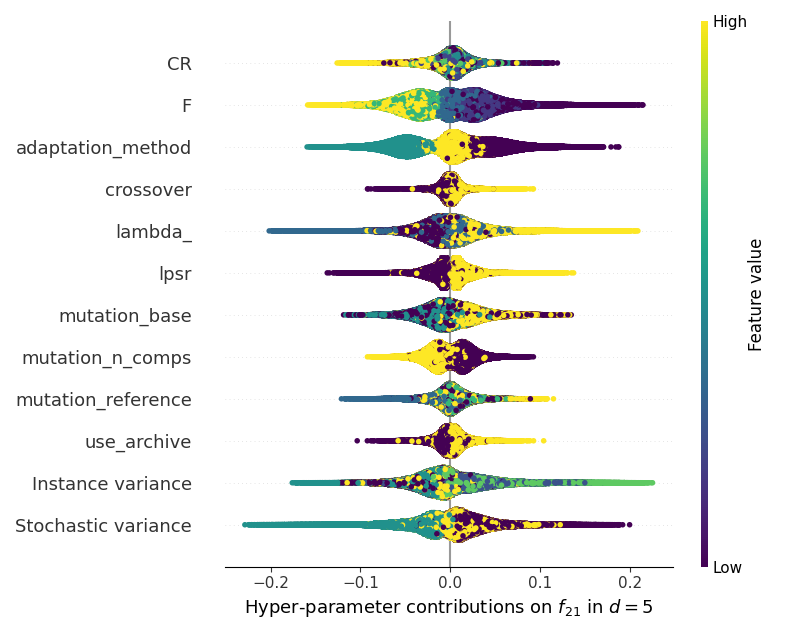}
	\includegraphics[height=0.15\textheight,trim=60mm 0mm 30mm 0mm,clip]{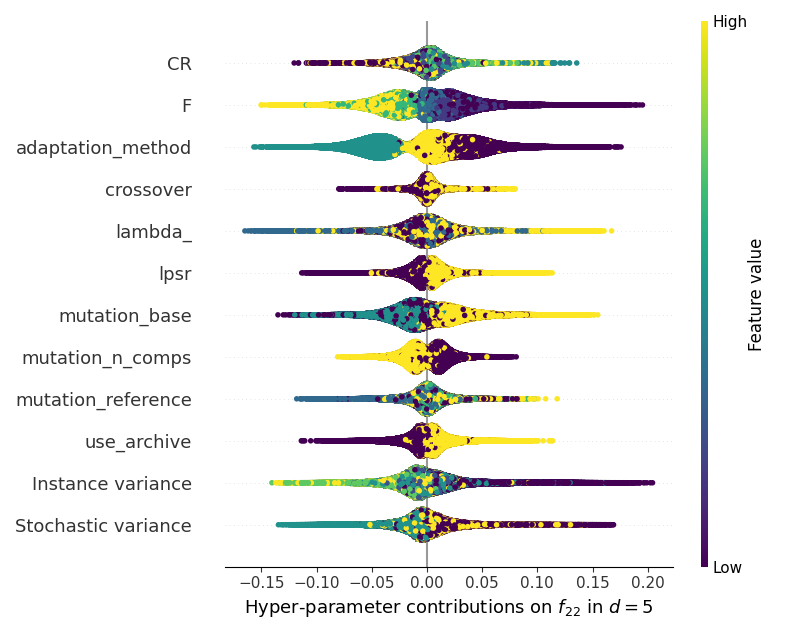}
	\includegraphics[height=0.15\textheight,trim=60mm 0mm 30mm 0mm,clip]{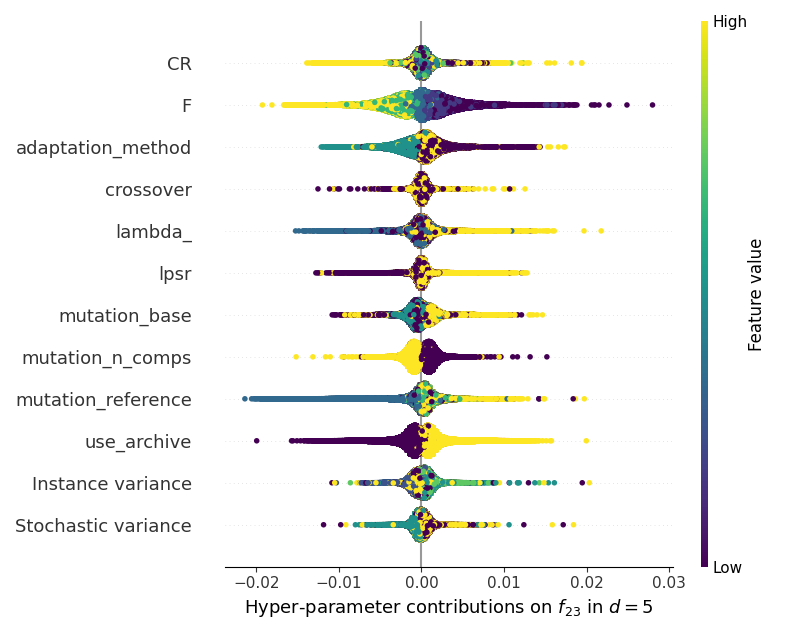}
	\includegraphics[height=0.15\textheight,trim=60mm 0mm 10mm 0mm,clip]{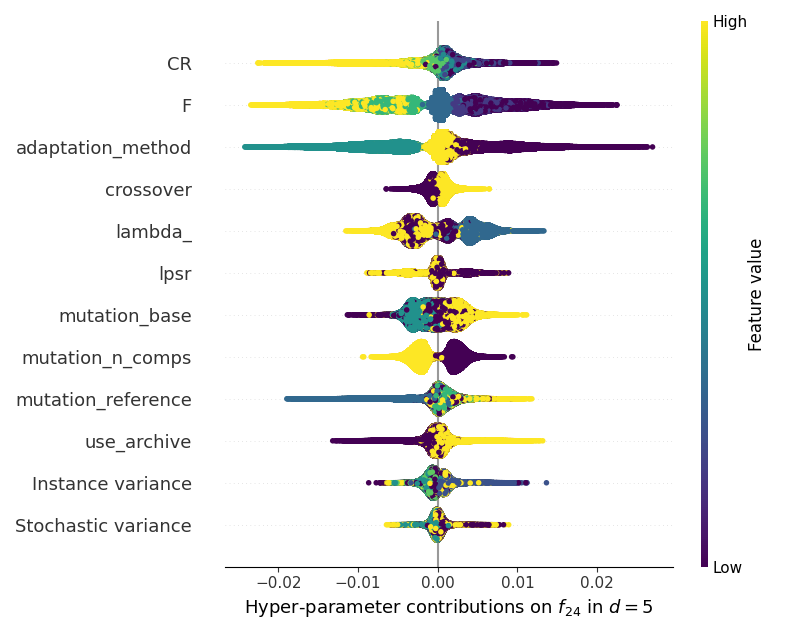}
\caption{Hyper-parameter contributions to the AOCC (shap values) per benchmark function for \textbf{d=5} for \textbf{modular DE}. Options per module are sorted alphabetically (for categorical parameters) or numerically, refer to Table~\ref{tab:modde_modules} for colour coding. 
\label{fig:de_shapxplaind5}} 
\end{figure*}

\begin{figure*}[t]
\centering
	\includegraphics[height=0.15\textheight,trim=4mm 0mm 30mm 0mm,clip]{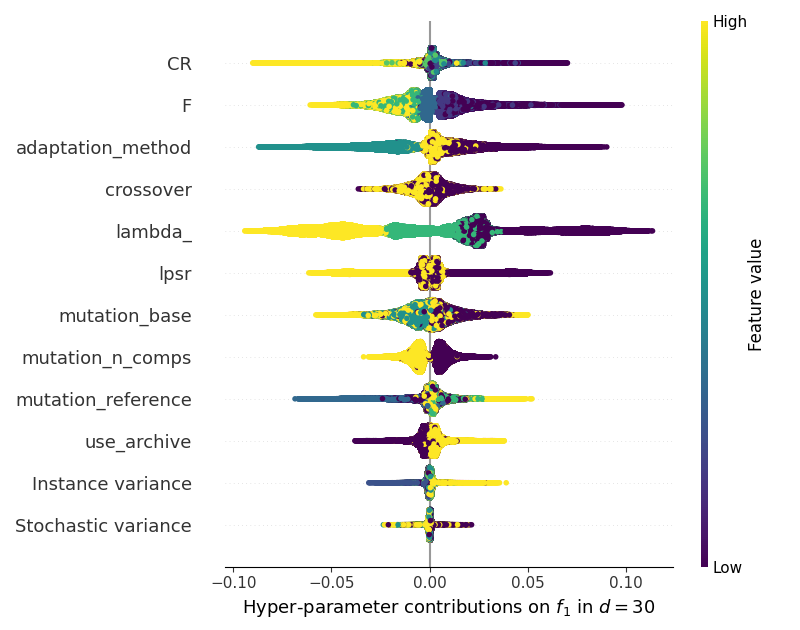}
	\includegraphics[height=0.15\textheight,trim=60mm 0mm 30mm 0mm,clip]{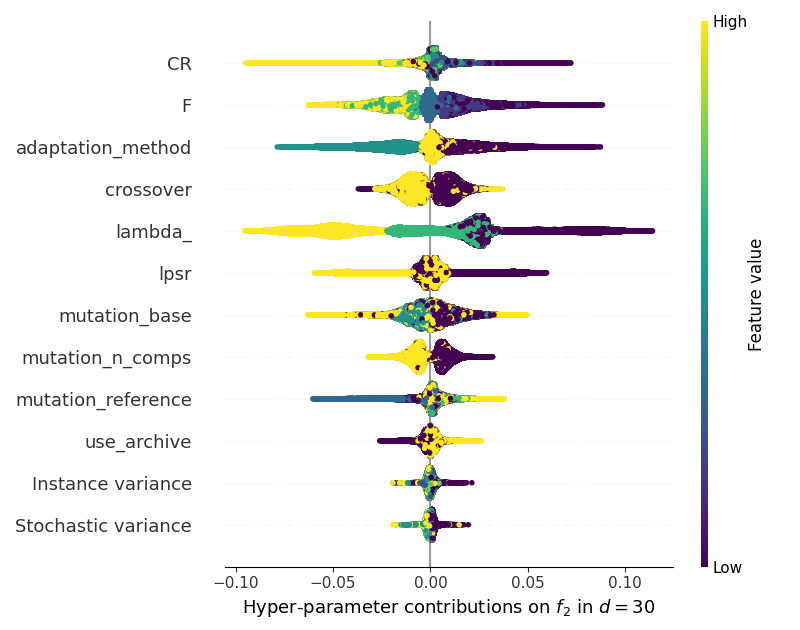}
	\includegraphics[height=0.15\textheight,trim=60mm 0mm 30mm 0mm,clip]{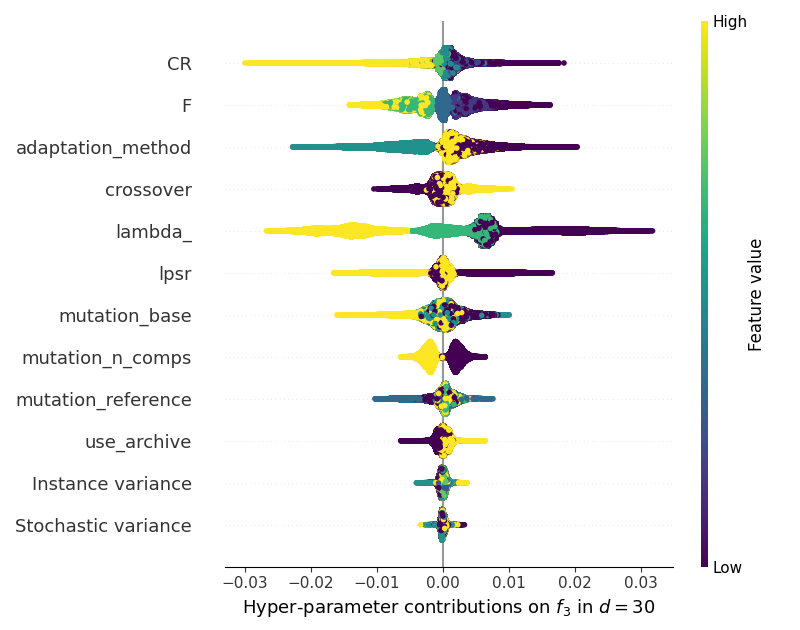}
	\includegraphics[height=0.15\textheight,trim=60mm 0mm 30mm 0mm,clip]{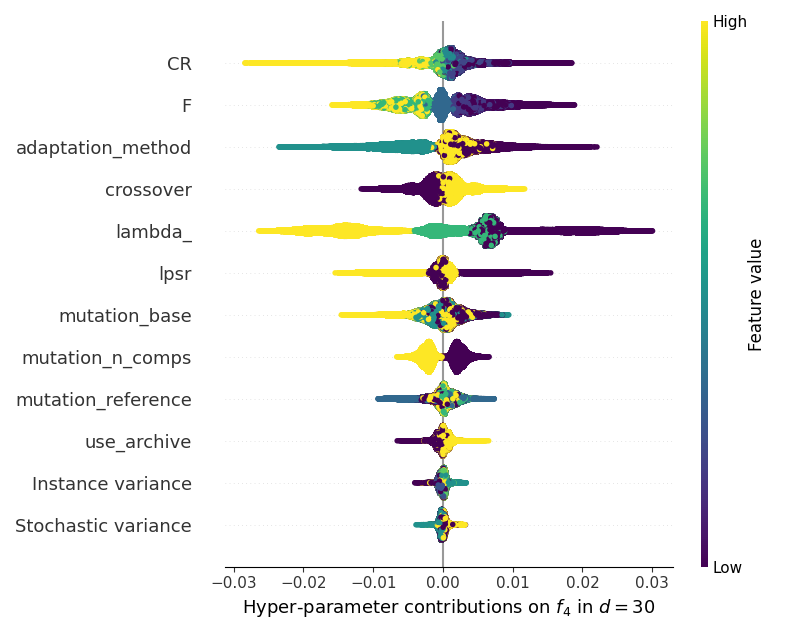}
	\includegraphics[height=0.15\textheight,trim=60mm 0mm 30mm 0mm,clip]{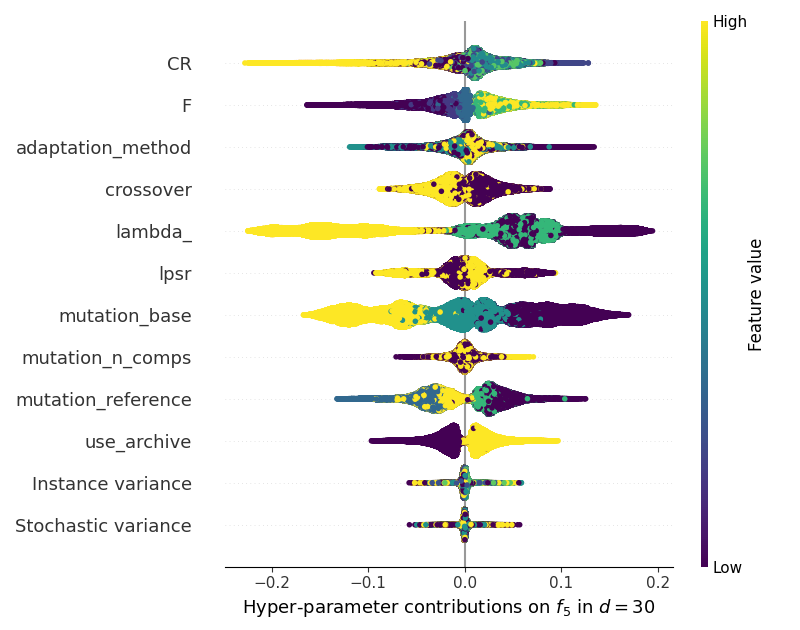}
	\includegraphics[height=0.15\textheight,trim=60mm 0mm 10mm 0mm,clip]{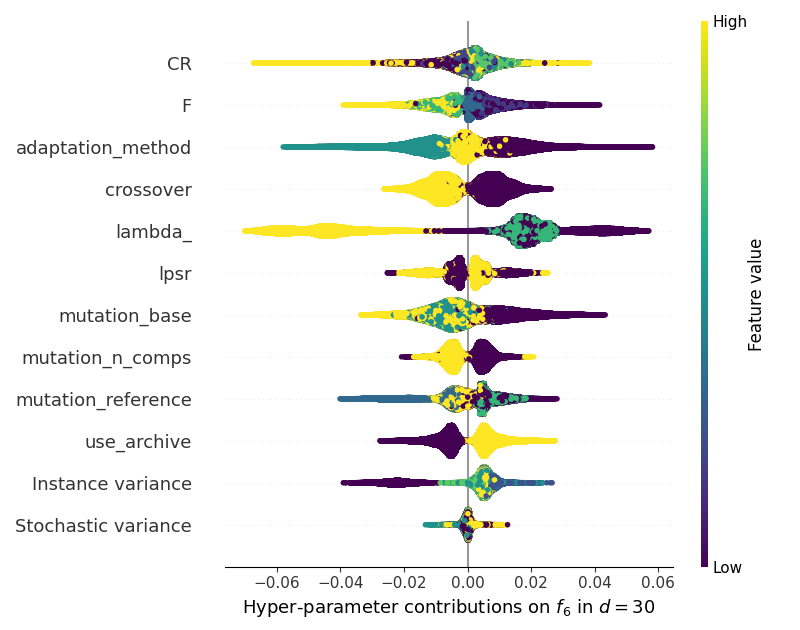}
 
	\includegraphics[height=0.15\textheight,trim=4mm 0mm 30mm 0mm,clip]{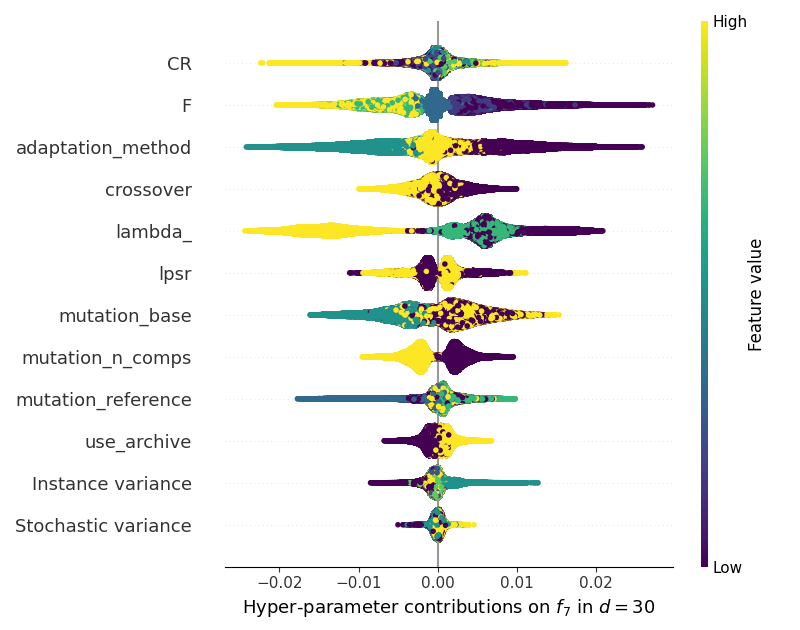}
	\includegraphics[height=0.15\textheight,trim=60mm 0mm 30mm 0mm,clip]{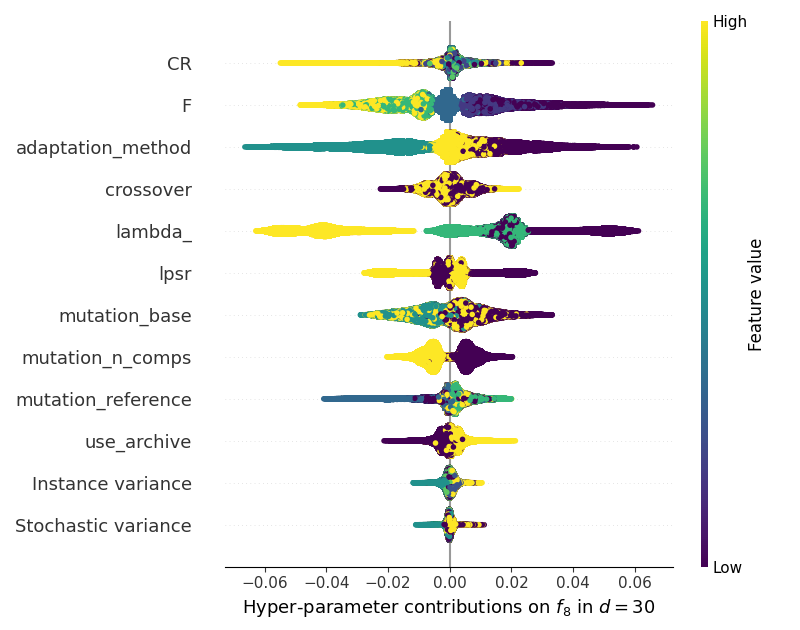}
	\includegraphics[height=0.15\textheight,trim=60mm 0mm 30mm 0mm,clip]{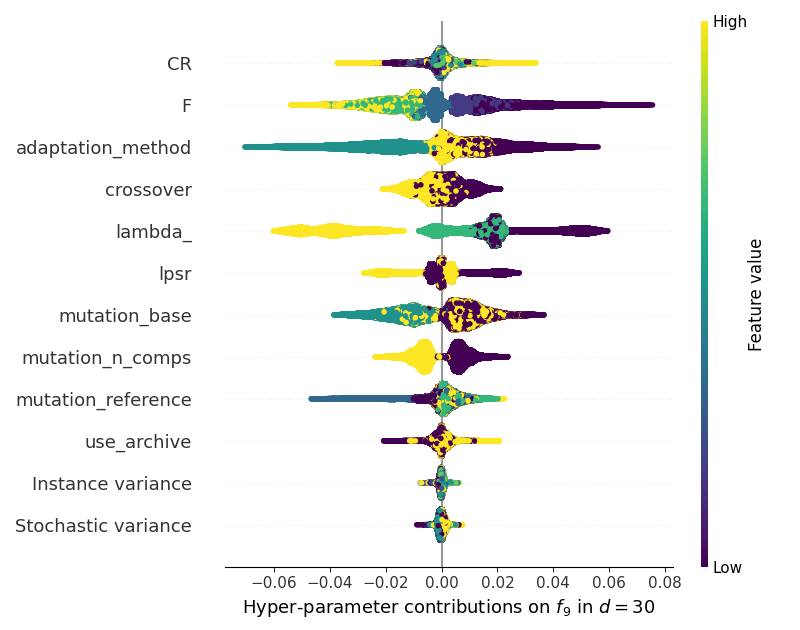}
	\includegraphics[height=0.15\textheight,trim=60mm 0mm 30mm 0mm,clip]{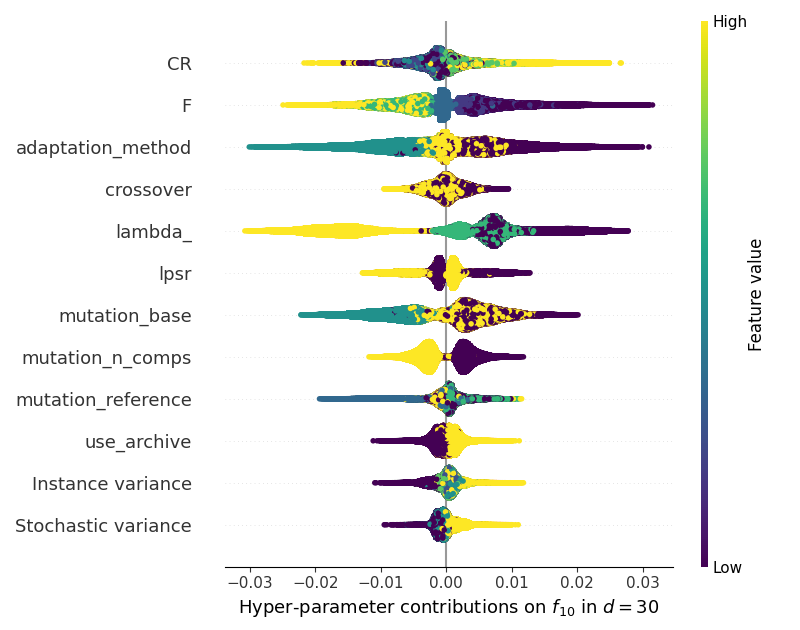}
	\includegraphics[height=0.15\textheight,trim=60mm 0mm 30mm 0mm,clip]{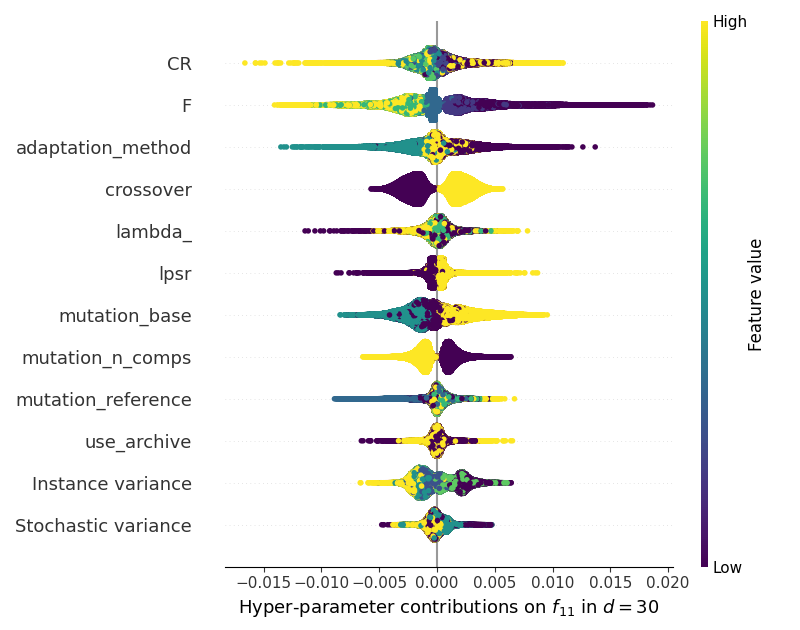}
	\includegraphics[height=0.15\textheight,trim=60mm 0mm 10mm 0mm,clip]{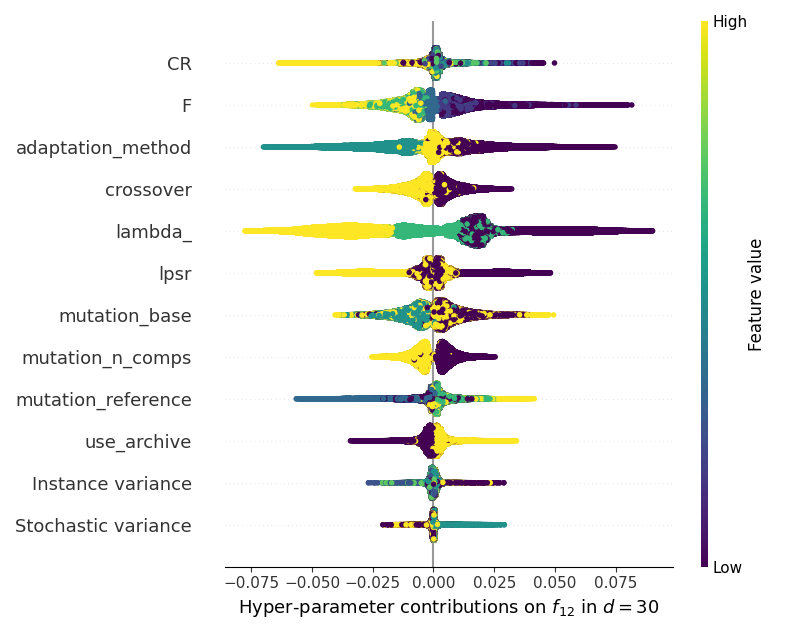}
 
	\includegraphics[height=0.15\textheight,trim=4mm 0mm 30mm 0mm,clip]{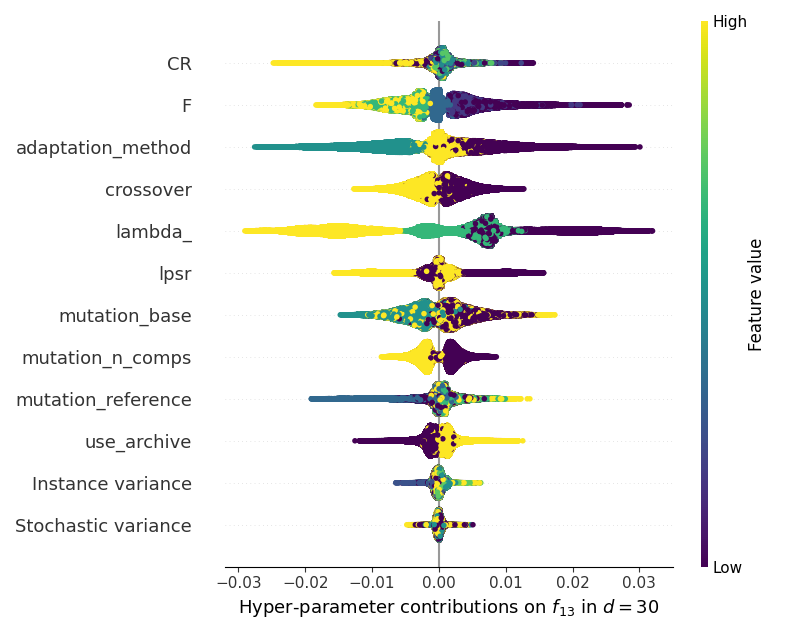}
	\includegraphics[height=0.15\textheight,trim=60mm 0mm 30mm 0mm,clip]{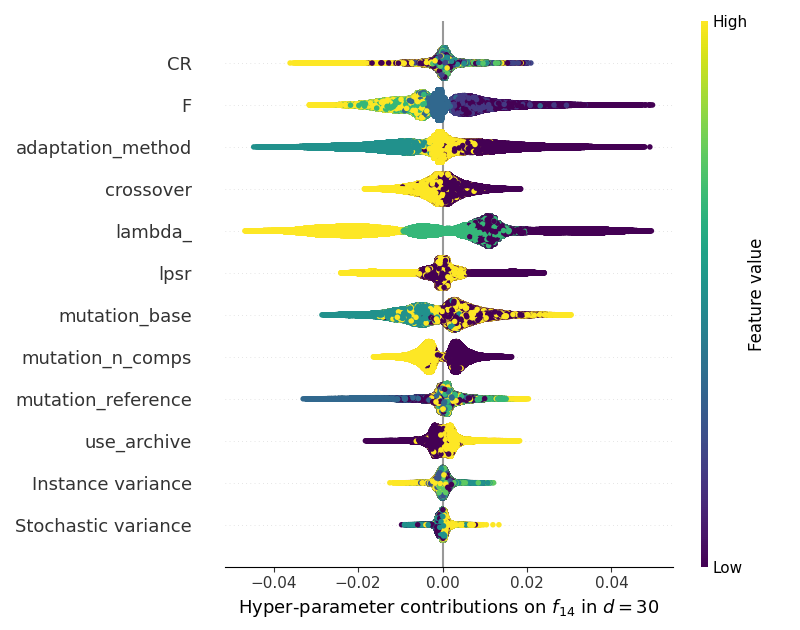}
	\includegraphics[height=0.15\textheight,trim=60mm 0mm 30mm 0mm,clip]{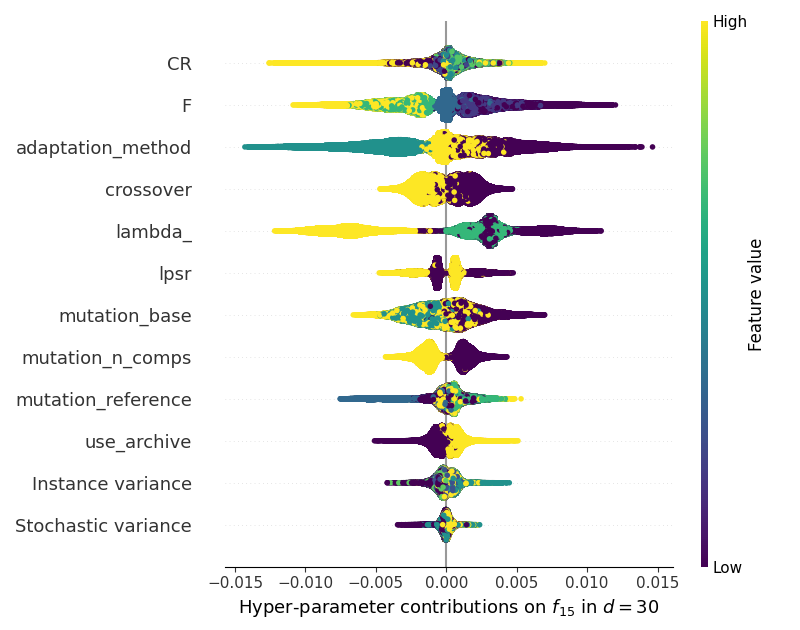}
	\includegraphics[height=0.15\textheight,trim=60mm 0mm 30mm 0mm,clip]{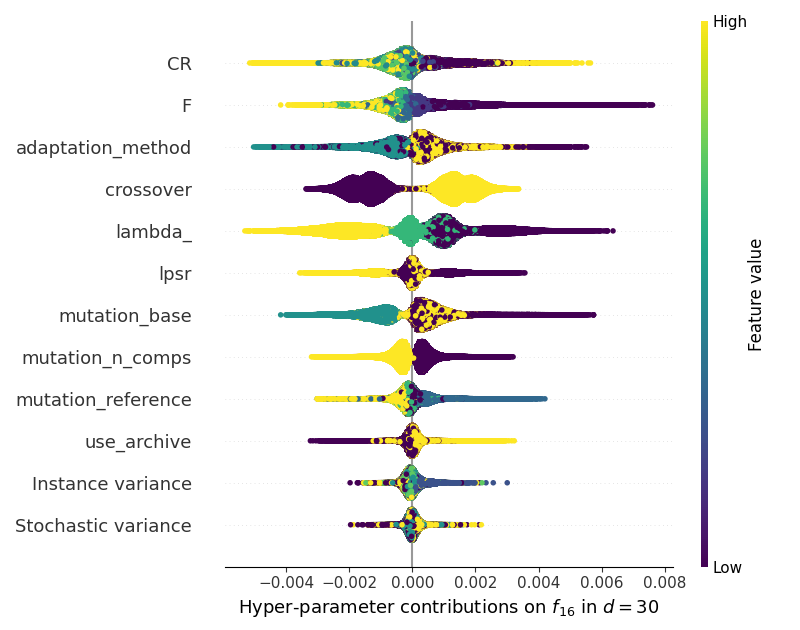}
	\includegraphics[height=0.15\textheight,trim=60mm 0mm 30mm 0mm,clip]{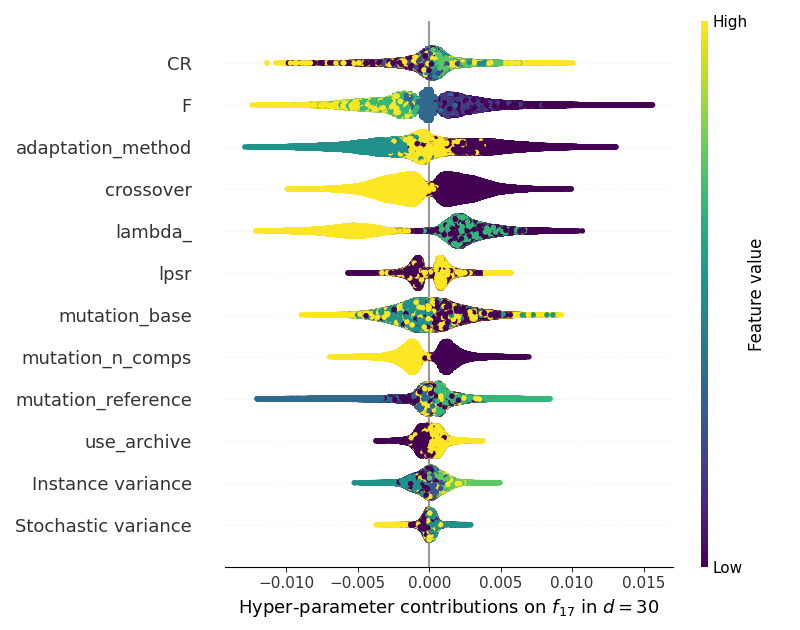}
	\includegraphics[height=0.15\textheight,trim=60mm 0mm 10mm 0mm,clip]{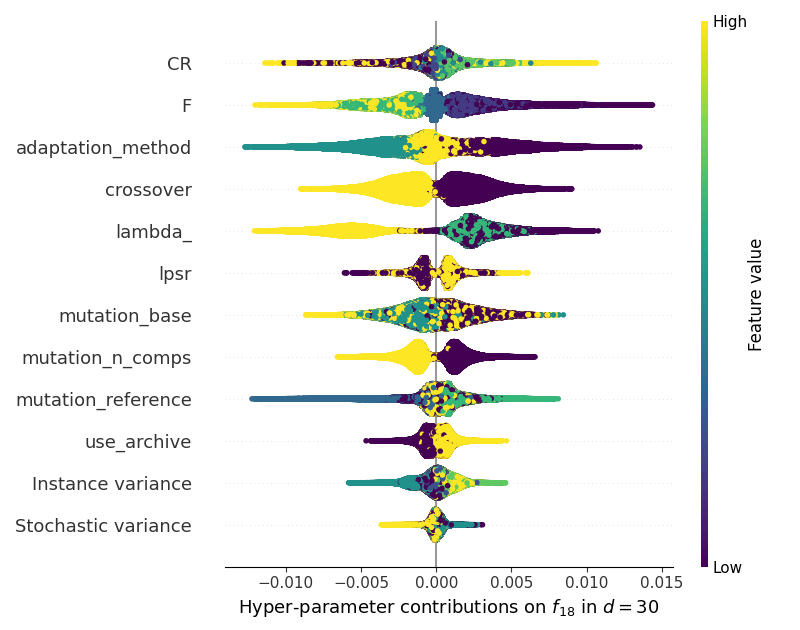}
 
	\includegraphics[height=0.15\textheight,trim=4mm 0mm 30mm 0mm,clip]{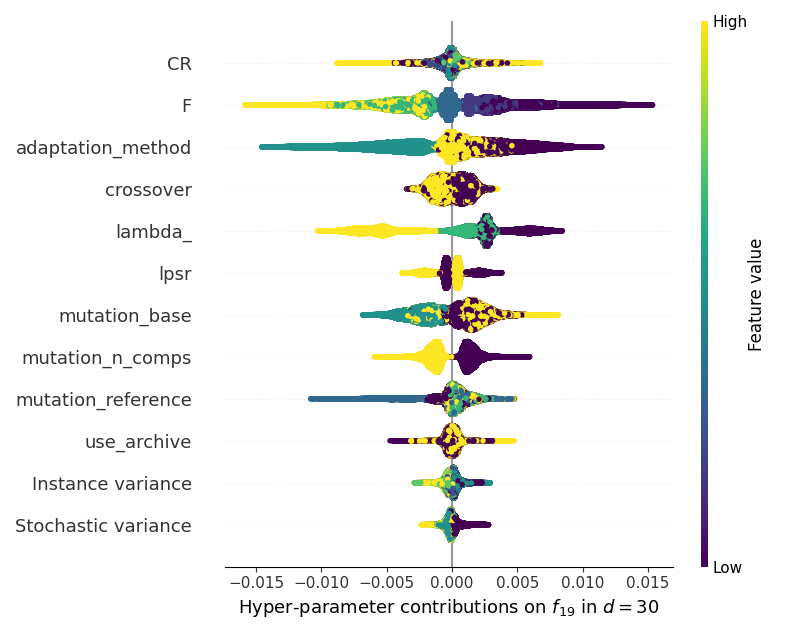}
	\includegraphics[height=0.15\textheight,trim=60mm 0mm 30mm 0mm,clip]{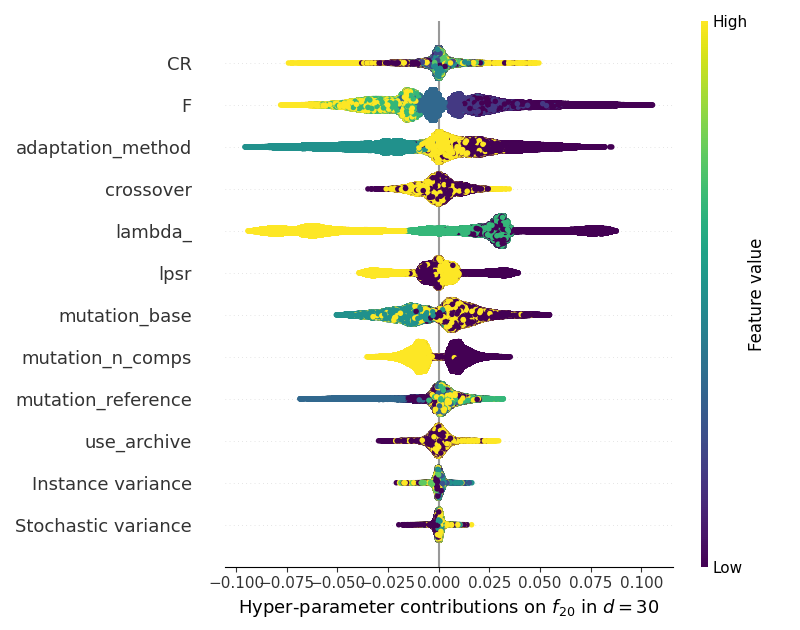}
	\includegraphics[height=0.15\textheight,trim=60mm 0mm 30mm 0mm,clip]{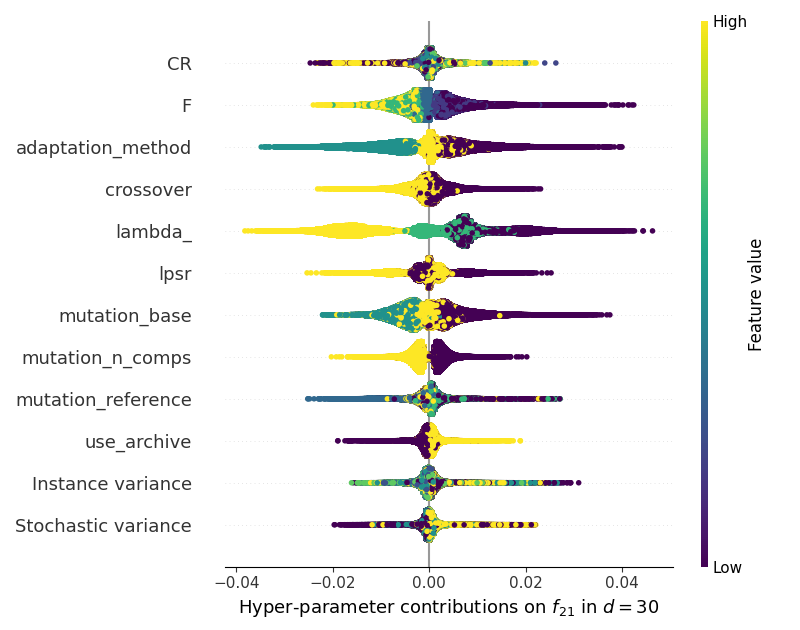}
	\includegraphics[height=0.15\textheight,trim=60mm 0mm 30mm 0mm,clip]{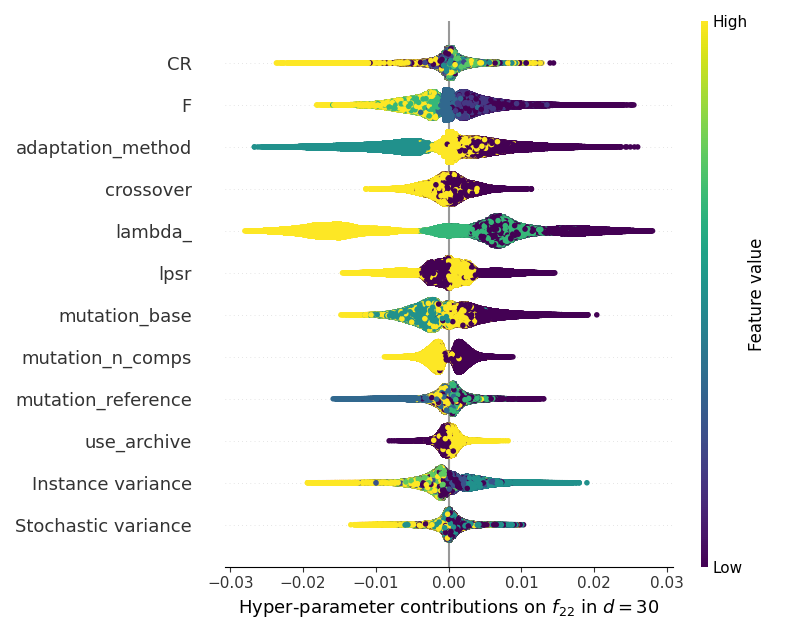}
	\includegraphics[height=0.15\textheight,trim=60mm 0mm 30mm 0mm,clip]{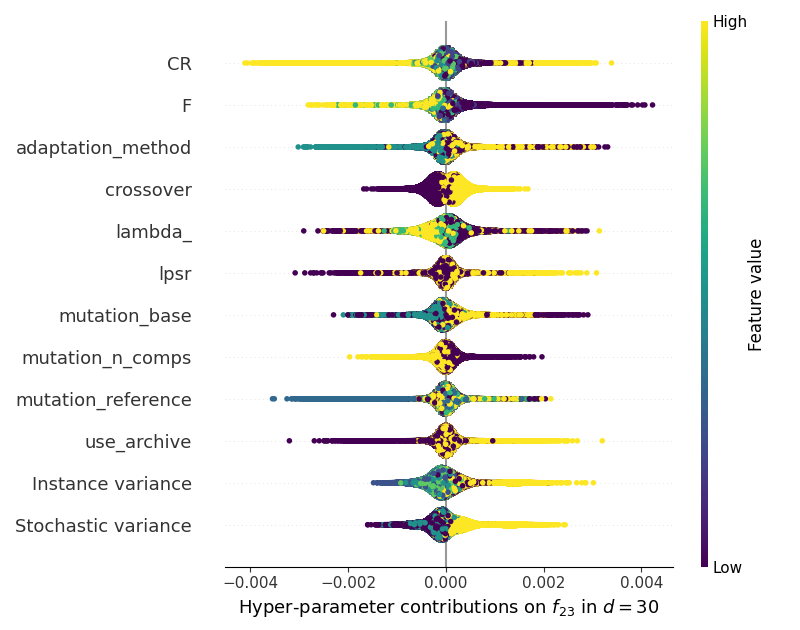}
	\includegraphics[height=0.15\textheight,trim=60mm 0mm 10mm 0mm,clip]{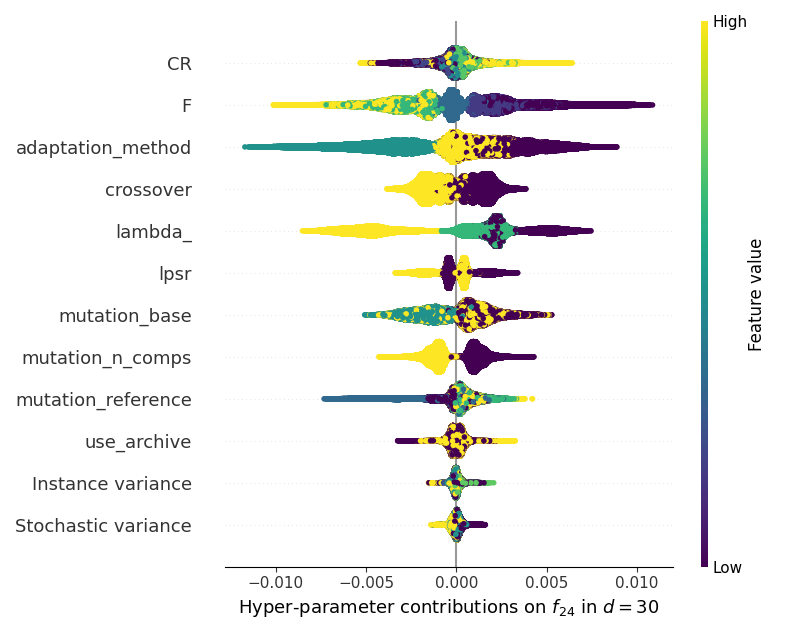}
\caption{Hyper-parameter contributions per benchmark function for \textbf{d=30} for \textbf{modular DE}. Options per module are sorted alphabetically (for categorical parameters) or numerically, refer to Table~\ref{tab:modde_modules} for colour coding. 
\label{fig:de_shapxplaind30}}
\end{figure*}

For the modular Differential Evolution, we have a total of $36\,288$ configurations. We should note that, similar to the CMA-ES, the models from which we calculate the SHAP-values are intentionally overfit, resulting in $R^2$ scores between $0.9$ and $0.96$. Figures~\ref{fig:de_shapxplaind5} and \ref{fig:de_shapxplaind30} show the resulting SHAP values for all functions, in dimensionality 5 and 30 respectively. For dimensionality 5, we observe strong effects from the population size parameter  $\lambda$, with the default (smallest) value often contributing negatively to the anytime performance. For the unimodal problems, the population size $2\cdot d$ seems to be preferred, while the multimodal functions benefit from an increase to $10\cdot d$.  For the adaptation method, we surprisingly find that turning this module off generally leads to good performance, while performance deteriorates when jDE is enabled. 

When looking at instance and stochastic variance, we see that on several problems, the impact of these two components is relatively large. However, the total impact of these two areas of variance seems roughly equal in size, which was not the case for the CMA-ES. This is particularly noticeable for functions 21 and 22 (5-dimensional) in Figure~\ref{fig:shapxplaind5}, where the instance variance dominates the CMA-ES module choices, while for DE this effect is much less severe. This suggests that the transformations used to create the different instances can make the problem relatively more easy/hard from the CMA-ES perspective, without impacting the performance of DE in a similar manner. 

When looking at the discretized parameters F and CR, we observe some interesting effects. While for most functions, the impact changes are seemingly in trend with the size of the parameter, e.g., F becoming more positive the lower the value becomes, some 30-dimensional functions show completely different types of impact for CR. As an example, F7 shows that the highest CR value has both a very positive and a very negative impact on the anytime performance. This highlights an inherent limitation with this analysis method, in that we only focus on the single-module impacts. In this version of DE, is seems intuitive that several modules might interact with each other, such as the choice of crossover method and CR, which might explain the unusual SHAP plot for F7. Within \iohx{} we can zoom in on these kinds of higher-order interactions by utilizing different explainability methods, we consider this outside the scope of this current paper.



\subsubsection{Algorithm/configuration selection}

While the SHAP-based analysis gives an overview of the global impact of module settings on different functions, we are also interested in considering the best combination of modules for each setting. To illustrate the potential gain in performance which could be achieved by selecting parameters for each individual function instead of the configuration which performs best on average, we contrast these two settings in Table~\ref{tab:de_configs_perf}. The results from this table are aggregated in Table~\ref{tab:de_perf_summary}. This table shows that the performance gain of selecting an average-best instead of a random configuration is on average $0.29$ for dimensionality 5 and $0.10$ for dimensionality 30.  On top of this, the performance of a perfect per-function algorithm selector would gain $0.12$ and $0.03$ AOCC respectively.

The parameter settings which constitute both the average-best and each per-function best are shown in Table~\ref{tab:fame-de}. Additionally, the bracketed number next to each module setting indicate the average loss in performance if it is switched to any of the other options displayed in Table~\ref{tab:modde_modules}. When looking at the overall best configuration in 5d, we see that a change to the base-vector would have the largest impact on AOCC, with similarly large performance changes on the individual functions. If we compare this to the results in Figure~\ref{fig:de_shapxplaind5}, we notice that across all configurations, the impact of the module is relatively minor. This suggests that interactions with this parameter might be relatively strong, and that the other modules are set in a way which is only well-performing when combined with this specific parameter setting. 

Another interesting observation from Table~\ref{tab:fame-de} is that the highest population sizes are never chosen in the 30-dimensional setting. This is likely influenced by the combination of chosen budget ($10\,000$) and performance measure, where very high population sizes could lead to slower, but eventually more precise, convergence. This highlights that, as with any benchmarking setup, the design choices made should be carefully considered when drawing conclusions from \iohx{}. Another interesting observation regarding population sizes is that the winning settings seem to be dimensionality-specific, as the winning configurations in 5d do not correspond to winning configurations in 30d (for the same function). In 5d smaller population sizes with no population size reduction (lpsr) is preferred on simpler functions while the opposite is true on more complex functions. In 30d, population size reduction seems less beneficial on most of the functions. The base-vector on the other hand is typically very stable over the different dimensionalities and should often be set to `target'.

\begin{table*}
\caption{Performance of single-best, average-best and overall average algorithm performance over all configurations per function and dimension for \textbf{modular DE}. \textbf{Boldface} for the single-best configuration indicates a significant improvement over the average best configuration (for that dimension), \textbf{Boldface} for the average best configuration indicates a significant improvement ($p$ value $< 0.05$) over the average AOCC of all configurations. The standard deviation is denoted between brackets. \label{tab:de_configs_perf}}
\resizebox{\textwidth}{!}{\begin{tabular}{llllllll}
\toprule
& \multicolumn{3}{c}{$d=5$} & \multicolumn{3}{c}{$d=30$} \\
Function & \multicolumn{1}{c}{single-best} & \multicolumn{1}{c}{avg-best} & \multicolumn{1}{c}{all} & \multicolumn{1}{c}{single-best} & \multicolumn{1}{c}{avg-best} & \multicolumn{1}{c}{all} \\
\midrule
f1 Sphere & \textbf{0.97 (0.00)} & \textbf{0.93 (0.00)} & 0.66 (0.26) & 0.79 (0.01) & \textbf{0.79 (0.01)} & 0.43 (0.07) \\
f2 Ellipsoid & \textbf{0.94 (0.01)} & \textbf{0.88 (0.01)} & 0.48 (0.33) & \textbf{0.57 (0.02)} & \textbf{0.52 (0.04)} & 0.17 (0.07) \\
f3 Rastrigin & \textbf{0.89 (0.01)} & \textbf{0.66 (0.21)} & 0.29 (0.25) & \textbf{0.42 (0.01)} & \textbf{0.39 (0.01)} & 0.34 (0.02) \\
f4 BuecheRastrigin & \textbf{0.82 (0.03)} & \textbf{0.46 (0.27)} & 0.22 (0.20) & \textbf{0.41 (0.01)} & \textbf{0.38 (0.01)} & 0.33 (0.02) \\
f5 LinearSlope & \textbf{1.00 (0.00)} & 0.99 (0.01) & 0.94 (0.17) & \textbf{0.97 (0.00)} & \textbf{0.84 (0.01)} & 0.57 (0.20) \\
f6 AttractiveSector & 0.88 (0.05) & \textbf{0.75 (0.26)} & 0.30 (0.22) & 0.38 (0.01) & \textbf{0.38 (0.01)} & 0.27 (0.05) \\
f7 StepEllipsoid & \textbf{0.91 (0.01)} & \textbf{0.89 (0.02)} & 0.37 (0.25) & \textbf{0.42 (0.01)} & \textbf{0.40 (0.01)} & 0.35 (0.02) \\
f8 Rosenbrock & 0.81 (0.06) & \textbf{0.71 (0.23)} & 0.25 (0.21) & 0.38 (0.01) & \textbf{0.37 (0.02)} & 0.24 (0.06) \\
f9 RosenbrockRotated & 0.80 (0.03) & \textbf{0.70 (0.24)} & 0.23 (0.21) & 0.38 (0.01) & \textbf{0.38 (0.01)} & 0.24 (0.06) \\
f10 EllipsoidRotated & \textbf{0.78 (0.03)} & \textbf{0.68 (0.18)} & 0.08 (0.13) & \textbf{0.22 (0.01)} & \textbf{0.19 (0.01)} & 0.14 (0.02) \\
f11 Discus & \textbf{0.83 (0.01)} & \textbf{0.67 (0.20)} & 0.17 (0.16) & \textbf{0.39 (0.01)} & \textbf{0.35 (0.01)} & 0.34 (0.01) \\
f12 BentCigar & 0.55 (0.19) & \textbf{0.49 (0.16)} & 0.08 (0.11) & 0.34 (0.03) & \textbf{0.34 (0.03)} & 0.05 (0.06) \\
f13 SharpRidge & 0.71 (0.01) & \textbf{0.62 (0.18)} & 0.17 (0.14) & 0.42 (0.02) & \textbf{0.42 (0.02)} & 0.31 (0.02) \\
f14 DifferentPowers & \textbf{0.90 (0.01)} & \textbf{0.86 (0.04)} & 0.49 (0.18) & 0.60 (0.01) & \textbf{0.59 (0.01)} & 0.43 (0.04) \\
f15 RastriginRotated & \textbf{0.25 (0.19)} & \textbf{0.14 (0.02)} & 0.10 (0.04) & 0.36 (0.01) & \textbf{0.35 (0.00)} & 0.33 (0.01) \\
f16 Weierstrass & \textbf{0.47 (0.17)} & \textbf{0.27 (0.11)} & 0.17 (0.06) & \textbf{0.43 (0.01)} & \textbf{0.42 (0.00)} & 0.41 (0.00) \\
f17 Schaffers10 & \textbf{0.71 (0.06)} & \textbf{0.55 (0.12)} & 0.29 (0.11) & \textbf{0.50 (0.01)} & \textbf{0.47 (0.01)} & 0.44 (0.01) \\
f18 Schaffers1000 & \textbf{0.66 (0.08)} & \textbf{0.44 (0.11)} & 0.21 (0.08) & \textbf{0.46 (0.01)} & \textbf{0.43 (0.01)} & 0.41 (0.01) \\
f19 GriewankRosenbrock & 0.26 (0.15) & \textbf{0.21 (0.01)} & 0.19 (0.03) & \textbf{0.46 (0.01)} & \textbf{0.45 (0.00)} & 0.43 (0.01) \\
f20 Schwefel & \textbf{0.68 (0.23)} & \textbf{0.44 (0.24)} & 0.25 (0.14) & \textbf{0.47 (0.00)} & \textbf{0.47 (0.00)} & 0.30 (0.08) \\
f21 Gallagher101 & \textbf{0.73 (0.23)} & \textbf{0.47 (0.36)} & 0.29 (0.21) & \textbf{0.50 (0.07)} & \textbf{0.45 (0.02)} & 0.41 (0.03) \\
f22 Gallagher21 & 0.72 (0.23) & \textbf{0.50 (0.36)} & 0.26 (0.18) & 0.46 (0.03) & \textbf{0.45 (0.03)} & 0.40 (0.02) \\
f23 Katsuura & \textbf{0.24 (0.08)} & 0.19 (0.01) & 0.19 (0.01) & \textbf{0.48 (0.01)} & 0.46 (0.00) & 0.46 (0.00) \\
f24 LunacekBiRastrigin & \textbf{0.12 (0.02)} & \textbf{0.10 (0.01)} & 0.08 (0.02) & \textbf{0.35 (0.00)} & \textbf{0.35 (0.00)} & 0.33 (0.01) \\
\bottomrule
\end{tabular}}
\end{table*}

\begin{table*}
\centering
\caption{Performance summary of \textbf{modular DE}. The gain avg-best is the average improvement of the average best configuration over all configurations. The gain single-best is the average improvement of the single-best configurations (per function) over all configurations. \textbf{Boldface} in the table indicates a significant improvement (avg-best versus all and single-best versus avg-best).}\label{tab:de_perf_summary}
\begin{tabular}{lrr}
\toprule
Measure & $d=5$ & $d=30$ \\
\midrule
Average performance & 0.28 & 0.34 \\
Gain avg-best & \textbf{0.29} &  \textbf{0.10} \\
Gain single-best & \textbf{0.41}  & \textbf{0.13} \\
\bottomrule
\end{tabular}
\end{table*}

\include{hallOfFame/de-hall-of-fame}




\subsection{Performance comparison between ModCMA and ModDE}
While we have to stress that it is not the main aim of this work to compare ModCMA with ModDE, we can do a (limited) comparison in performance between the two modular frameworks using the data collected. In Table~\ref{tab:perf_comparison_d5} we listed the performance of the single-best, avg-best and ``default'' performance of the frameworks in $5$ dimensions and in Table~\ref{tab:perf_comparison_d30} for $30$ dimensions. The ``default'' performance denotes the average performance over all algorithm configurations. Results in boldface indicates a significant improvement (with p-value threshold of $0.05$)  between the two frameworks in the same category (so single-best vs single-best, avg-best vs avg-best etc.).
It can be observed that in general ModCMA seems to be better performing on most of the BBOB functions except for $f_2, f_3$ and $f_4$ (all from the separable functions group). However, take into account that not all ModDE and not all ModCMA parameter options are tested (as this is computationally infeasible). If we look at the stability or tune-ability of the different frameworks we can conclude that in general ModDE gains more performance on average by optimizing the hyperparameters. For ModCMA in 30 dimensions, the gain (in terms of AOCC) to optimize for a specific function is on average not significant over taking the average best configuration (See Table~\ref{tab:cma_perf_summary}).


\begin{table*}
\centering
\caption{Performance comparison of \textbf{modular CMA-ES} and \textbf{modular DE} for \textbf{d=5}. \textbf{Boldface} indicates a significant improvement either between single-best configurations, avg-best configurations or all configurations. The standard deviation is denoted between brackets. \label{tab:perf_comparison_d5}}
\resizebox{\textwidth}{!}{\begin{tabular}{lllllll}
\toprule
 & \multicolumn{2}{c}{single-best} & \multicolumn{2}{c}{avg-best} & \multicolumn{2}{c}{all} \\
Function & \multicolumn{1}{c}{modDE} & \multicolumn{1}{c}{modCMA} & \multicolumn{1}{c}{modDE} & \multicolumn{1}{c}{modCMA} & \multicolumn{1}{c}{modDE} & \multicolumn{1}{c}{modCMA} \\
\midrule
f1 Sphere & 0.97 (0.00) & \textbf{0.98 (0.00)} & 0.93 (0.00) & \textbf{0.97 (0.00)} & 0.66 (0.26) & \textbf{0.69 (0.30)} \\
f2 Ellipsoid & \textbf{0.94 (0.01)} & 0.91 (0.01) & 0.88 (0.01) & \textbf{0.90 (0.01)} & \textbf{0.48 (0.33)} & 0.39 (0.36) \\
f3 Rastrigin & \textbf{0.89 (0.01)} & 0.38 (0.29) & \textbf{0.66 (0.21)} & 0.30 (0.23) & \textbf{0.29 (0.25)} & 0.12 (0.08) \\
f4 BuecheRastrigin & \textbf{0.82 (0.03)} & 0.17 (0.16) & \textbf{0.46 (0.27)} & 0.13 (0.02) & \textbf{0.22 (0.20)} & 0.09 (0.04) \\
f5 LinearSlope & 1.00 (0.00) & 1.00 (0.00) & \textbf{0.99 (0.01)} & 0.99 (0.00) & 0.94 (0.17) & \textbf{0.95 (0.16)} \\
f6 AttractiveSector & 0.88 (0.05) & \textbf{0.95 (0.01)} & 0.75 (0.26) & \textbf{0.90 (0.01)} & 0.30 (0.22) & \textbf{0.48 (0.34)} \\
f7 StepEllipsoid & 0.91 (0.01) & \textbf{0.95 (0.00)} & 0.89 (0.02) & \textbf{0.94 (0.03)} & 0.37 (0.25) & \textbf{0.50 (0.31)} \\
f8 Rosenbrock & 0.81 (0.06) & \textbf{0.91 (0.02)} & 0.71 (0.23) & \textbf{0.85 (0.02)} & 0.25 (0.21) & \textbf{0.42 (0.34)} \\
f9 RosenbrockRotated & 0.80 (0.03) & \textbf{0.92 (0.01)} & 0.70 (0.24) & \textbf{0.86 (0.01)} & 0.23 (0.21) & \textbf{0.45 (0.33)} \\
f10 EllipsoidRotated & 0.78 (0.03) & \textbf{0.90 (0.01)} & 0.68 (0.18) & \textbf{0.89 (0.01)} & 0.08 (0.13) & \textbf{0.39 (0.36)} \\
f11 Discus & 0.83 (0.01) & \textbf{0.92 (0.01)} & 0.67 (0.20) & \textbf{0.91 (0.01)} & 0.17 (0.16) & \textbf{0.37 (0.32)} \\
f12 BentCigar & 0.55 (0.19) & \textbf{0.86 (0.07)} & 0.49 (0.16) & \textbf{0.73 (0.23)} & 0.08 (0.11) & \textbf{0.33 (0.33)} \\
f13 SharpRidge & 0.71 (0.01) & \textbf{0.90 (0.01)} & 0.62 (0.18) & \textbf{0.87 (0.01)} & 0.17 (0.14) & \textbf{0.37 (0.31)} \\
f14 DifferentPowers & 0.90 (0.01) & \textbf{0.95 (0.01)} & 0.86 (0.04) & \textbf{0.95 (0.00)} & 0.49 (0.18) & \textbf{0.62 (0.27)} \\
f15 RastriginRotated & 0.25 (0.19) & \textbf{0.48 (0.02)} & 0.14 (0.02) & \textbf{0.43 (0.19)} & 0.10 (0.04) & \textbf{0.13 (0.10)} \\
f16 Weierstrass & 0.47 (0.17) & \textbf{0.81 (0.16)} & 0.27 (0.11) & \textbf{0.69 (0.19)} & 0.17 (0.06) & \textbf{0.26 (0.19)} \\
f17 Schaffers10 & 0.71 (0.06) & \textbf{0.90 (0.02)} & 0.55 (0.12) & \textbf{0.87 (0.05)} & 0.29 (0.11) & \textbf{0.41 (0.21)} \\
f18 Schaffers1000 & 0.66 (0.08) & \textbf{0.82 (0.12)} & 0.44 (0.11) & \textbf{0.68 (0.16)} & 0.21 (0.08) & \textbf{0.30 (0.16)} \\
f19 GriewankRosenbrock & 0.26 (0.15) & 0.40 (0.23) & 0.21 (0.01) & \textbf{0.31 (0.02)} & 0.19 (0.03) & \textbf{0.24 (0.06)} \\
f20 Schwefel & \textbf{0.68 (0.23)} & 0.36 (0.11) & \textbf{0.44 (0.24)} & 0.21 (0.07) & \textbf{0.25 (0.14)} & 0.19 (0.05) \\
f21 Gallagher101 & 0.73 (0.23) & 0.84 (0.23) & 0.47 (0.36) & \textbf{0.53 (0.32)} & 0.29 (0.21) & \textbf{0.40 (0.31)} \\
f22 Gallagher21 & 0.72 (0.23) & 0.79 (0.26) & \textbf{0.50 (0.36)} & 0.46 (0.33) & 0.26 (0.18) & \textbf{0.36 (0.30)} \\
f23 Katsuura & 0.24 (0.08) & \textbf{0.66 (0.22)} & 0.19 (0.01) & \textbf{0.49 (0.20)} & 0.19 (0.01) & \textbf{0.21 (0.10)} \\
f24 LunacekBiRastrigin & 0.12 (0.02) & 0.14 (0.03) & 0.10 (0.01) & \textbf{0.13 (0.02)} & 0.08 (0.02) & \textbf{0.09 (0.02)} \\
\bottomrule
\end{tabular}}
\end{table*}

\begin{table*}
\centering
\caption{Performance comparison of \textbf{modular CMA-ES} and \textbf{modular DE} for \textbf{d=30}. \textbf{Boldface} indicates a significant improvement either between single-best configurations, avg-best configurations or all configurations. \label{tab:perf_comparison_d30}}
\resizebox{\textwidth}{!}{\begin{tabular}{lllllll}
\toprule
 & \multicolumn{2}{c}{single-best} & \multicolumn{2}{c}{avg-best} & \multicolumn{2}{c}{all} \\
Function & \multicolumn{1}{c}{modDE} & \multicolumn{1}{c}{modCMA} & \multicolumn{1}{c}{modDE} & \multicolumn{1}{c}{modCMA} & \multicolumn{1}{c}{modDE} & \multicolumn{1}{c}{modCMA} \\
\midrule
f1 Sphere & 0.79 (0.01) & \textbf{0.95 (0.00)} & 0.79 (0.01) & \textbf{0.90 (0.00)} & 0.43 (0.07) & \textbf{0.62 (0.21)} \\
f2 Ellipsoid & \textbf{0.57 (0.02)} & 0.29 (0.01) & \textbf{0.52 (0.04)} & 0.29 (0.01) & 0.17 (0.07) & \textbf{0.18 (0.06)} \\
f3 Rastrigin & \textbf{0.42 (0.01)} & 0.40 (0.00) & 0.39 (0.01) & \textbf{0.39 (0.01)} & 0.34 (0.02) & \textbf{0.35 (0.02)} \\
f4 BuecheRastrigin & \textbf{0.41 (0.01)} & 0.38 (0.01) & 0.38 (0.01) & \textbf{0.38 (0.01)} & 0.33 (0.02) & \textbf{0.34 (0.02)} \\
f5 LinearSlope & 0.97 (0.00) & \textbf{0.99 (0.00)} & 0.84 (0.01) & \textbf{0.98 (0.00)} & 0.57 (0.20) & \textbf{0.89 (0.17)} \\
f6 AttractiveSector & 0.38 (0.01) & \textbf{0.58 (0.01)} & 0.38 (0.01) & \textbf{0.58 (0.03)} & 0.27 (0.05) & \textbf{0.35 (0.11)} \\
f7 StepEllipsoid & 0.42 (0.01) & \textbf{0.45 (0.01)} & 0.40 (0.01) & \textbf{0.45 (0.01)} & 0.35 (0.02) & \textbf{0.38 (0.04)} \\
f8 Rosenbrock & 0.38 (0.01) & \textbf{0.41 (0.01)} & 0.37 (0.02) & \textbf{0.40 (0.01)} & 0.24 (0.06) & \textbf{0.32 (0.08)} \\
f9 RosenbrockRotated & 0.38 (0.01) & \textbf{0.41 (0.00)} & 0.38 (0.01) & \textbf{0.41 (0.00)} & 0.24 (0.06) & \textbf{0.34 (0.06)} \\
f10 EllipsoidRotated & 0.22 (0.01) & \textbf{0.29 (0.01)} & 0.19 (0.01) & \textbf{0.28 (0.01)} & 0.14 (0.02) & \textbf{0.19 (0.06)} \\
f11 Discus & 0.39 (0.01) & \textbf{0.41 (0.01)} & 0.35 (0.01) & \textbf{0.40 (0.01)} & 0.34 (0.01) & \textbf{0.35 (0.02)} \\
f12 BentCigar & 0.34 (0.03) & \textbf{0.51 (0.09)} & 0.34 (0.03) & \textbf{0.45 (0.06)} & 0.05 (0.06) & \textbf{0.21 (0.17)} \\
f13 SharpRidge & 0.42 (0.02) & \textbf{0.51 (0.05)} & 0.42 (0.02) & \textbf{0.49 (0.05)} & 0.31 (0.02) & \textbf{0.38 (0.07)} \\
f14 DifferentPowers & 0.60 (0.01) & \textbf{0.70 (0.01)} & 0.59 (0.01) & \textbf{0.69 (0.00)} & 0.43 (0.04) & \textbf{0.54 (0.10)} \\
f15 RastriginRotated & 0.36 (0.01) & \textbf{0.40 (0.01)} & 0.35 (0.00) & \textbf{0.39 (0.01)} & 0.33 (0.01) & \textbf{0.35 (0.02)} \\
f16 Weierstrass & 0.43 (0.01) & \textbf{0.50 (0.01)} & 0.42 (0.00) & \textbf{0.50 (0.01)} & 0.41 (0.00) & \textbf{0.43 (0.03)} \\
f17 Schaffers10 & 0.50 (0.01) & \textbf{0.64 (0.02)} & 0.47 (0.01) & \textbf{0.60 (0.03)} & 0.44 (0.01) & \textbf{0.48 (0.04)} \\
f18 Schaffers1000 & 0.46 (0.01) & \textbf{0.55 (0.01)} & 0.43 (0.01) & \textbf{0.52 (0.02)} & 0.41 (0.01) & \textbf{0.44 (0.04)} \\
f19 GriewankRosenbrock & 0.46 (0.01) & \textbf{0.55 (0.00)} & 0.45 (0.00) & \textbf{0.48 (0.01)} & 0.43 (0.01) & \textbf{0.46 (0.02)} \\
f20 Schwefel & 0.47 (0.00) & \textbf{0.48 (0.00)} & 0.47 (0.00) & \textbf{0.47 (0.00)} & 0.30 (0.08) & \textbf{0.40 (0.09)} \\
f21 Gallagher101 & 0.50 (0.07) & \textbf{0.66 (0.22)} & 0.45 (0.02) & \textbf{0.56 (0.20)} & 0.41 (0.03) & \textbf{0.45 (0.10)} \\
f22 Gallagher21 & 0.46 (0.03) & 0.48 (0.11) & \textbf{0.45 (0.03)} & 0.44 (0.01) & 0.40 (0.02) & \textbf{0.42 (0.03)} \\
f23 Katsuura & 0.48 (0.01) & \textbf{0.55 (0.01)} & 0.46 (0.00) & \textbf{0.49 (0.01)} & 0.46 (0.00) & \textbf{0.47 (0.03)} \\
f24 LunacekBiRastrigin & 0.35 (0.00) & \textbf{0.38 (0.00)} & 0.35 (0.00) & \textbf{0.38 (0.00)} & 0.33 (0.01) & \textbf{0.35 (0.01)} \\
\bottomrule
\end{tabular}}
\end{table*}

\subsection{Structural Bias Analysis} \label{SB}
Part of the \iohx{} framework is also the work on Structural Bias detection \cite{biastoolbox,deepbias}. The single-best and avg-best configurations are tested for structural bias and the results of the combined statistical and deep-learning SB detection methods are given in Table \ref{tab:fame-cma} for Modular CMA-ES and Table \ref{tab:fame-de} for Modular DE. 

To allow a visual in-depth analysis of the strength and type of structural bias, the toolbox provides for each test also a visualization of the distributions of locations of the final optima of a special test function $f_0$~\cite{Kononova2015} (which should be uniform when no structural bias is present). 
Each of the reported SB cases has been verified and if needed corrected by visual inspection of these distributions. This correction is particularly applied when the detection method is predicting the gaps/cluster class, since we know that centre bias is often misclassified as gaps/clusters (in theory centre bias is a special version of cluster bias). All visualizations are available in our Github repository~\cite{code_xplainer_niki_van_stein_2024_10568760}.

It can be observed from the two tables that many single-best solutions exhibit structurally biased behaviour. For Modular DE, this is often bias near the boundaries of the search domain, while for Modular CMA-ES, it is almost always biased around the centre of the domain except for a few configurations 30 dimensions. It is important to take this structural bias effect into account, especially as we know that the optima of the BBOB benchmark are not uniformly distributed in the whole search domain (but in $[-4,4]^d$, and in the case of $f_5$ -- at the boundaries)~\cite{long2023bbob}, which might explain why CMA-ES is often outperforming on this benchmark.

\section{Additional Insights} \label{AASD}

In the proposed \iohx{} pipeline, a lot of data is being collected about the behaviour and performance of algorithm configurations. Apart from the main insights we can derive using XAI methods on this data, there are additional benefits of storing and re-using this data. In this section we want to show a few ideas which can easily be done using the data collected and may prove beneficial to selecting, configuring and understanding these algorithm configurations.

The first ``by-product'' of our \iohx{} toolbox, is that we can easily train machine learning models to predict a complete algorithm configuration  (including module and hyper-parameter settings) based on landscape features. Basically performing automated algorithm configuration (and selection), similar to existing works such as \cite{kerschke2019automated} and others \cite{long2022learning,improvingnevergrad}. Note however, that in this work we combine the selection and configuration of the algorithm as one machine learning task, using multi-output mixed classification and regression models. In our case, we limit the ML models to shallow decision trees and Random Forests for the sake of simplicity.
To showcase this, we collected classical exploratory landscape analysis (ELA) features using a design of experiments of $1024$ samples on each BBOB instance used in our experiments.
For each BBOB instance, we then took the single-best performing algorithm configuration (per modular framework), as the output of our machine learning pipeline. Using multi-output models, such as Random Forest and simple (and interpretable) decision trees, we can then train these models to predict good performing algorithm configurations based on the ELA features.


\begin{figure}[t]
\centering
	\includegraphics[width=0.48\textwidth,trim=0mm 0mm 00mm 0mm,clip]{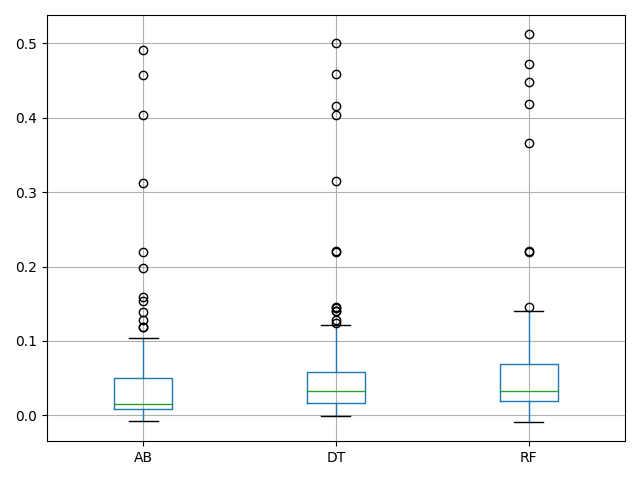}
    \includegraphics[width=0.48\textwidth,trim=0mm 0mm 00mm 0mm,clip]{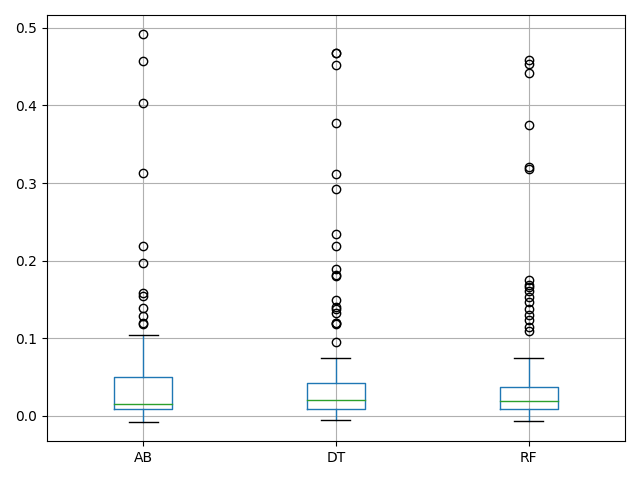}
\caption{Performance loss (AOCC) of \textbf{modular CMA-ES} on all BBOB functions for \textbf{d=30} for the predicted configurations using a Random Forest (RF) model, an interpretable (shallow) Decision Tree (DT) model and the baselines Average Best (AB) configuration against the performance of the single best run per instance. 
Using a leave-one-function out approach on the left and a leave-one-instance out approach on the right.\label{fig:classifiers_fid_cma}}
\end{figure}

\begin{figure}[!t]
\centering
	\includegraphics[width=0.48\textwidth,trim=0mm 0mm 00mm 0mm,clip]{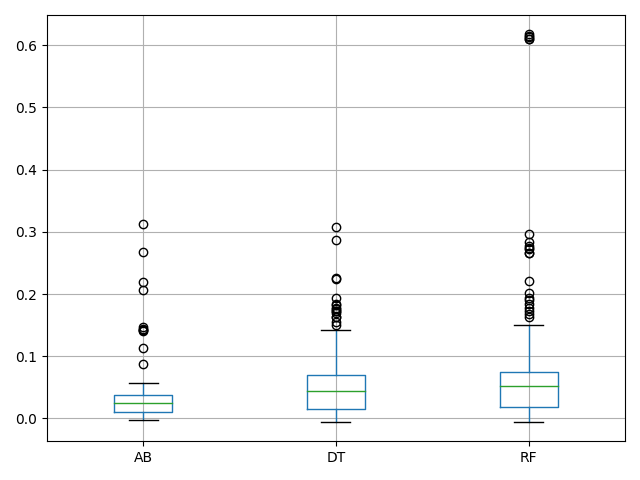}
    \includegraphics[width=0.48\textwidth,trim=0mm 0mm 00mm 0mm,clip]{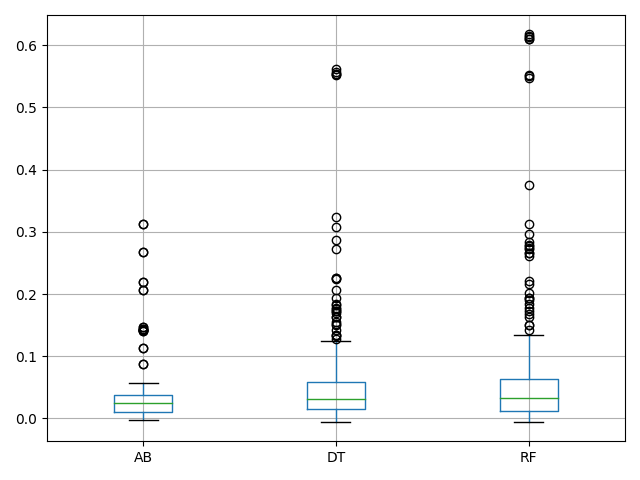}
\caption{Performance loss (AOCC) of \textbf{modular DE} on all BBOB functions for \textbf{d=30} for the predicted configurations using a Random Forest (RF) model, an interpretable (shallow) Decision Tree (DT) model and the baselines Average Best (AB) configuration against the performance of the single best run per instance. 
Using a leave-one-function out approach on the left and a leave-one-instance out approach on the right\label{fig:classifiers_fid_de}}
\end{figure}

In Figure ~\ref{fig:classifiers_fid_cma} we can observe the results of this Automated Algorithm Configuration (AAC) task on ModCMA. We see here the loss in AOCC compared to the best performing configuration per instance (for all BBOB functions and instances). We used two experimental setups, one that uses a leave-one-function out approach, where all minus one BBOB functions are used as training data and one function id is used as test, and a leave-one-instance out approach, where we only leave one instance of a BBOB function out as test set. The assumption is of course that leaving one instance out is an easier task than leaving a function out. We can observe that the Decision Tree and Random Forest model are quite close to the average best configuration, in the leave-one-instance-out experiment the results of the ML models are slightly more stable than just using the average best configuration. In general, we can see that the method works but that it still requires either better ML models or better ELA features to really outperform just taking an overall good configuration of the framework.
In Figure ~\ref{fig:classifiers_fid_de} we can observe similar results on the ModDE framework. On this framework, it seems even harder to beat the avg-best configuration. 



\begin{figure*}[!b]
\centering
	\includegraphics[width=1.\textwidth]{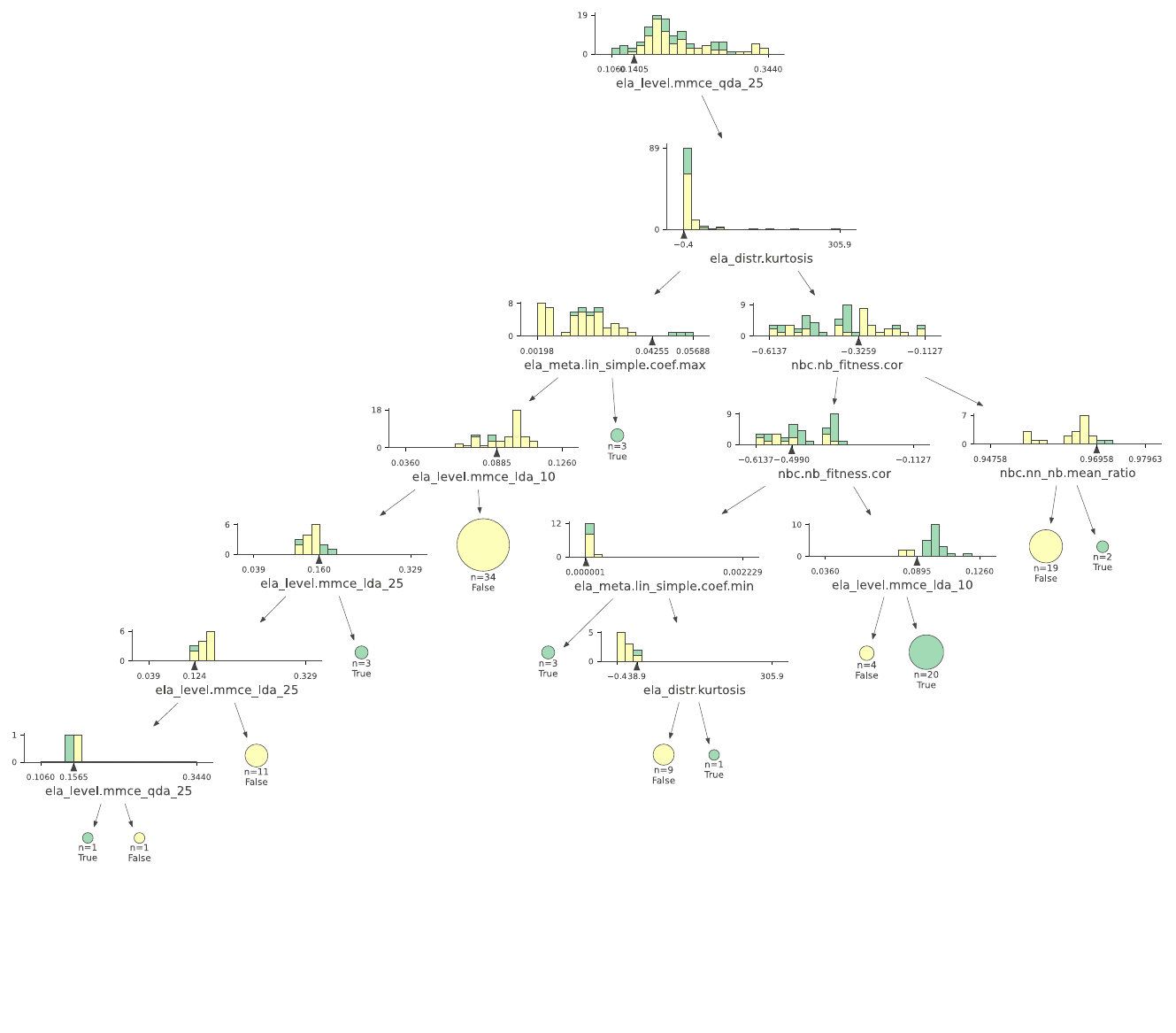}
\caption{Decision tree with max depth of 7 for elitist in \textbf{modular CMA-ES} with \textbf{d=30}. Each node of the tree represent a split where the distribution of the feature used for the split is visualized. Colors in the distribution denote the different labels (in this case yellow is `false' and green is `true' for the elitist module). The black arrow below the distribution denotes the splitting threshold.\label{fig:cmatree_elitist_d30}}
\end{figure*}

\begin{figure*}[t]
\centering
	\includegraphics[width=1.\textwidth]{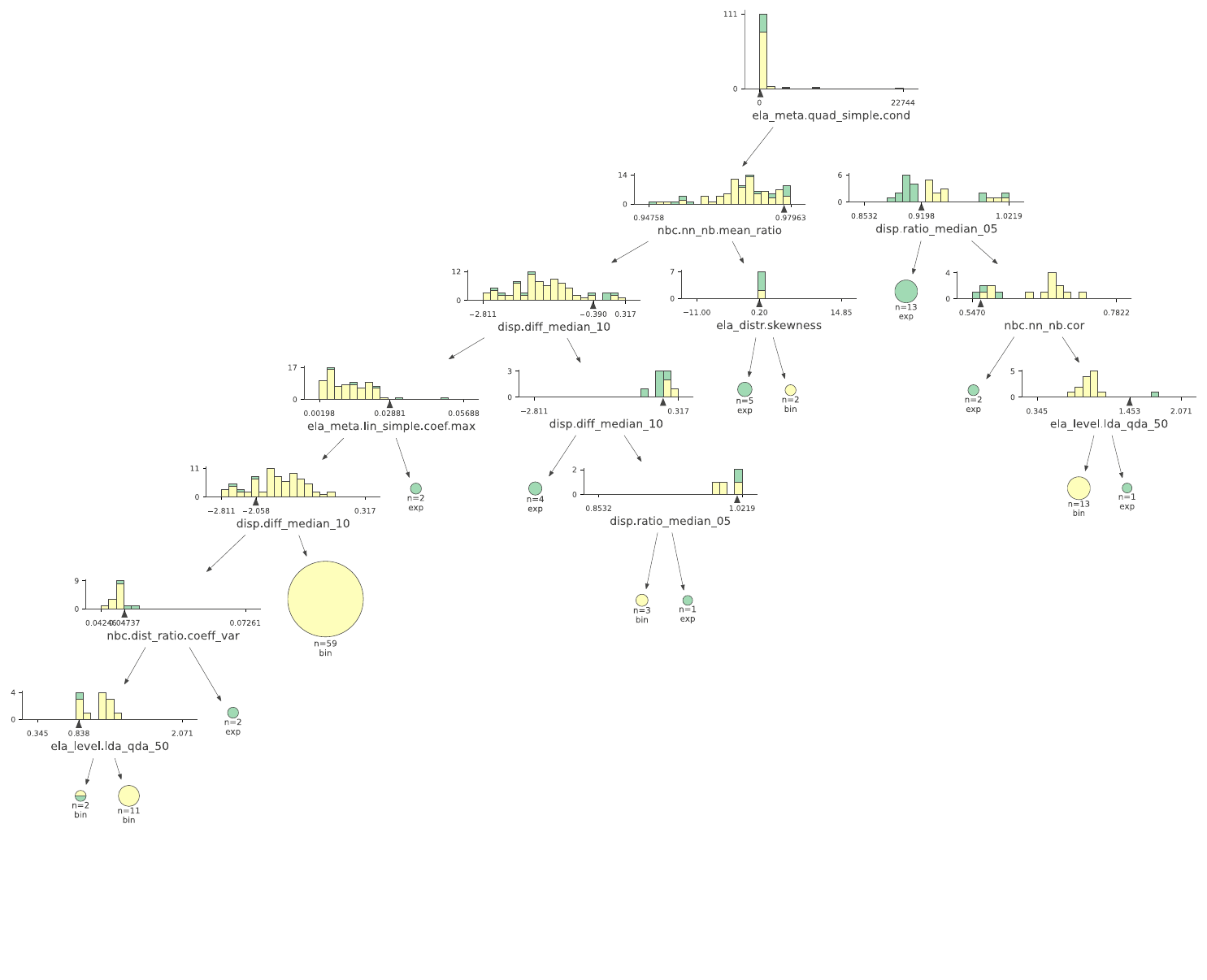}
\caption{Decision tree with max depth of 7 for crossover in \textbf{modular DE} with \textbf{d=30}. Each node of the tree represent a split where the distribution of the feature used for the split is visualized. Colors in the distribution denote the different labels (in this case yellow is `bin' and green is `exp' crossover). The black arrow below the distribution denotes the splitting threshold.\label{fig:detree_crossover_d30}}
\end{figure*}

As a second additional use of the collected data, we can build interpretable models such as shallow decision trees, to predict when to use a specific algorithm module. This types of ``wizards" can provide a lot of additional insights on what kind of landscapes and what kind of landscape features are important to consider when selecting certain modules and settings. An example for the module \emph{elitist} for ModCMA can be seen in Figure \ref{fig:cmatree_elitist_d30}. In this example the mean mis-classification error (mmce) from the ELA level feature set seems to be most important for either seting elitist to true or false. Another example for the module \emph{crossover} for the ModDE framework can be seen in Figure \ref{fig:detree_crossover_d30}.
All additional trees are available in our public Github repository \cite{code_xplainer_niki_van_stein_2024_10568760}.

In general, the take-away message is to keep behaviour and performance data of your algorithms as much as possible as it may serve additional research in the future, especially when better landscape analysis or algorithm configuration models become available.

\section{Conclusion}\label{sect:conclusion}

In this work a novel framework for explainable benchmarking for iterative optimization heuristics is proposed and implemented as software tool called \emph{\iohx{}}. With explainable benchmarking, we mean to shift the benchmarking paradigm from very much performance focused (usually with a limited setup, biased towards the proposed algorithm (component)), towards a benchmarking practise to gain insights into the behaviour and influence of different modules and hyper-parameter settings. These insights can either be generated from existing algorithms, modular frameworks or newly developed operators or algorithms.
The focus in explainable benchmarking lies in gaining understanding what search tactics and hyperparameters contribute on what kind of black box problems and function landscapes.
Using the \iohx{} toolbox, practisioners and researchers in the field can now easily perform such a detailed and broad analysis of components with relative ease. The ability of \iohx{} to provide detailed, explainable insights into algorithm performance marks a notable step forward in understanding and optimizing these complex systems. We showcase our proposed framework by two extensive use-cases, namely on the modular CMA-ES and the modular DE frameworks. Performing a detailed, explainable comparison of a staggering $52.128$ algorithm configurations over $24$ different benchmark functions ($5$ instances, $5$ random seeds) in both $5$ and $30$ dimensions.
We analysed the collected data from these experiments with the help of explainable AI techniques and showed how this analysis can lead to new insights for both algorithm frameworks. Note, that the conclusions drawn for these frameworks are of course limited to the experimental setup, such as the hyper-parameter choices included and the upper limit of the chosen $10.000$ evaluation budget. Increasing the budget (or using different commonly used budgets) and adding additional hyper-parameter choices would be a set for future work to investigate. 


In addition to the explanations and extensive analysis of these frameworks we also show that the data collected by the \iohx{} toolbox can be used for additional tasks such as Automated Algorithm Configuration and explaining algorithm behaviour in relation with exploratory landscape analysis.

Future research could include the exploration of the application of this framework to a broader range of optimization problems, enhancing its capabilities and integrating it with other machine learning techniques. This will not only deepen our understanding of heuristic optimization but also pave the way for the development of more sophisticated and efficient algorithms.
Additional future work includes the investigation on how our findings for Modular CMA and DE generalize over different benchmarks or dimensions and how different XAI methods would affect the outcome of our analysis.

\section*{Reproducibility statement}
We provide an open-source documented implementation of our package at \cite{code_xplainer_niki_van_stein_2024_10568760},
including install and how-to-use guide.
AOCC data collected from the runs are available as well, including detailed notebooks and scripts to repeat the experiments.
All experiments are performed carbon neutral ($CO_2$-free) by using solar power. 

\bibliographystyle{acm}
\bibliography{biblio.bib}

\end{document}

%% file: tables/modcma_modules.tex
\begin{table*}[]
    \centering
    \caption{Modules used for \textbf{modCMA} and their options. Only combinations where $\mu\leq\lambda$ are used. Values of default populations sizes are computed according to $4+\lfloor3\log(d)\rfloor$ for dimensionalities $d\in\{5,30\}$ and marked as $^\star$, while resulting offspring sizes  are computed as $\frac{\lambda}{2}$ and marked with $^\dagger$. Values of recombination weights for the $\lambda$-decay are computed as $w_i = \frac{1}{2^i} + \frac{1}{\lambda2^\lambda}$. Colour coding serves to distinguish options within each module in Figures~\ref{fig:shapxplaind5}, \ref{fig:shapxplaind30} and Table~\ref{tab:fame-cma}. \label{tab:modcma_modules}}
    \begin{tabular}{llp{55mm}}

       \textbf{Module Name} & \textbf{Shorthand} & \textbf{Domain} \\ \hline \hline
        Covariance adaptation & \texttt{covariance} &  \{\textcolor[HTML]{440154}{false}, \textcolor[HTML]{eed502}{true}\} \\
        Active update & \texttt{active} &  \{\textcolor[HTML]{440154}{false}, \textcolor[HTML]{eed502}{true}\} \\
        Base sampler & \texttt{base\_sampler} &  \{\textcolor[HTML]{440154}{Gaussian}, \textcolor[HTML]{21918c}{Halton}, \textcolor[HTML]{eed502}{Sobol}\} \\
        Elitism & \texttt{elitist} &  \{\textcolor[HTML]{440154}{false}, \textcolor[HTML]{eed502}{true}\} \\
        Mirrored sampling & \texttt{mirrored} &   \{\textcolor[HTML]{440154}{off}, \textcolor[HTML]{21918c}{mirrored}, \textcolor[HTML]{eed502}{mirrored pairwise}\} \\
        Recombination weights & \texttt{weights\_option} &  \{\textcolor[HTML]{440154}{default}, \textcolor[HTML]{21918c}{equal}, \textcolor[HTML]{eed502}{$\lambda$-decay}\}\\
        Step size adaptation & \texttt{step\_size\_adaptation} &  \{\textcolor[HTML]{440154}{CSA}, \textcolor[HTML]{eed502}{PSR}\} \\
        Local restarts & \texttt{local\_restart} & \{\textcolor[HTML]{440154}{none}, \textcolor[HTML]{21918c}{IPOP}, \textcolor[HTML]{eed502}{BIPOP}\} \\ 
        Population size & $\lambda$ & \{\textcolor[HTML]{440154}{5}, \textcolor[HTML]{472a7a}{8$^\star$}, \textcolor[HTML]{414487}{10}, \textcolor[HTML]{2e6f8e}{14$^\star$}, \textcolor[HTML]{22a884}{20}, \textcolor[HTML]{eed502}{200}\}\\
        Offspring size & $\mu$ & \{\textcolor[HTML]{440154}{2$^\dagger$}, \textcolor[HTML]{481b6d}{4$^\dagger$}, \textcolor[HTML]{482677}{5$^\dagger$}, \textcolor[HTML]{433d84}{7$^\dagger$}, \textcolor[HTML]{365c8d}{10$^\dagger$}, \textcolor[HTML]{2db27d}{20}, \textcolor[HTML]{eed502}{100$^\dagger$}\}\\
        \hline
    \end{tabular}
\end{table*}

%% file: tables/modde_modules.tex
\begin{table*}[]
    \centering
    \caption{Modules used for \textbf{modDE} and their options. Values of default populations sizes are computed according to $4+\lfloor3\log(d)\rfloor$ for dimensionalities $d\in\{5,30\}$ and marked as $^\star$.  Colour coding serves to distinguish options within each module in Figures~\ref{fig:de_shapxplaind5}, \ref{fig:de_shapxplaind30} and Table~\ref{tab:fame-de}.
    \label{tab:modde_modules}}
    \begin{tabular}{llp{55mm}}
       \textbf{Module Name} & \textbf{Shorthand} & \textbf{Domain} \\ \hline \hline
        Base vector & \texttt{base} &  \{\textcolor[HTML]{440154}{best}, \textcolor[HTML]{21918c}{rand}, \textcolor[HTML]{eed502}{target}\} \\
        Reference vector & \texttt{ref} &  \{\textcolor[HTML]{440154}{none}, \textcolor[HTML]{31688e}{best}, \textcolor[HTML]{35b779}{pbest}, \textcolor[HTML]{eed502}{rand}\} \\
        Number of differences & \texttt{diffs} &  \{\textcolor[HTML]{440154}{1}, \textcolor[HTML]{eed502}{2}\} \\
        Use archive & \texttt{archive} &  \{\textcolor[HTML]{440154}{false}, \textcolor[HTML]{eed502}{true}\} \\
        Crossover method & \texttt{crossover} &   \{\textcolor[HTML]{440154}{bin}, \textcolor[HTML]{eed502}{exp}\} \\
        F and CR adaptation method & \texttt{adaptation\_method} &  \{\textcolor[HTML]{440154}{none}, \textcolor[HTML]{21918c}{jDE}, \textcolor[HTML]{eed502}{shade}\} \\
        Population size reduction & \texttt{lpsr} & \{\textcolor[HTML]{440154}{false}, \textcolor[HTML]{eed502}{true}\} \\ 
        Population size & $\lambda$ & \{ \textcolor[HTML]{440154}{8$^\star$}, \textcolor[HTML]{482173}{10}, \textcolor[HTML]{38588c}{14$^\star$}, \textcolor[HTML]{1e9b8a}{50}, \textcolor[HTML]{69cd5b}{60}, \textcolor[HTML]{eed502}{300}\}\\ 
        Scale factor & $F$ &  \{\textcolor[HTML]{440154}{0.25}, \textcolor[HTML]{3b528b}{0.5}, \textcolor[HTML]{21918c}{0.75}, \textcolor[HTML]{5ec962}{1.25}, \textcolor[HTML]{eed502}{1.75}\} \\
        Crossover rate & $CR$ & \{\textcolor[HTML]{440154}{0.05}, \textcolor[HTML]{3b528b}{0.25}, \textcolor[HTML]{21918c}{0.5}, \textcolor[HTML]{5ec962}{0.75}, \textcolor[HTML]{eed502}{1.0}\} \\ 
        \hline
    \end{tabular}
\end{table*}

%% file: hallOfFame/cma_es-hall-of-fame.tex
\begin{table}[ht]
\centering
\caption{Hall of fame of \textbf{modular CMA-ES}: best configurations per function including average best (denoted by All in fid), with resulting values of AOCC and structural BIAS indicator. Denoted in brackets after each module setting is the average effect in AOCC score of choosing the respective option versus all other choices (keeping the rest of the settings the same). Colouring of module options is consistent with Table~\ref{tab:modcma_modules} and Figures~\ref{fig:shapxplaind5}, ~\ref{fig:shapxplaind30}. Options marked in \textit{italic} are consistently selected on this function for both dimensionalities. \label{tab:fame-cma}}
\resizebox{\textwidth}{!}{
\begin{tabular}{ccrrrrrrrrrrcc}
\toprule
\multicolumn{1}{c}{d} & \multicolumn{1}{c}{fid} & \multicolumn{1}{c}{active} & \multicolumn{1}{c}{base\_sampler} & \multicolumn{1}{c}{covariance} & \multicolumn{1}{c}{elitist} & \multicolumn{1}{c}{$\lambda$} & \multicolumn{1}{c}{local\_restart} & \multicolumn{1}{c}{mirrored} & \multicolumn{1}{c}{$\mu$} & \multicolumn{1}{c}{ssa} & \multicolumn{1}{c}{weights} & \multicolumn{1}{c}{AOCC} & \multicolumn{1}{c}{BIAS} \\
\midrule
5 & \textbf{All} & \textit{\textcolor[HTML]{eed502}{true}} (0.04) & \textcolor[HTML]{440154}{Gaussian} (0.19) & \textcolor[HTML]{eed502}{true} (0.13) & \textit{\textcolor[HTML]{440154}{false}} (0.28) & \textcolor[HTML]{22a884}{20} (0.38) & \textit{\textcolor[HTML]{21918c}{IPOP}} (0.02) & \textit{\textcolor[HTML]{21918c}{mirrored}} (0.05) & \textcolor[HTML]{365c8d}{10} (0.25) & \textit{\textcolor[HTML]{440154}{CSA}} (0.08) & \textit{\textcolor[HTML]{440154}{default}} (0.09) & 0.666806 & \textcolor[HTML]{cc4778}{centre}\\
5 & 1 & \textit{\textcolor[HTML]{440154}{false}} (0.00) & \textcolor[HTML]{21918c}{Halton} (0.00) & \textcolor[HTML]{eed502}{true} (0.85) & \textcolor[HTML]{eed502}{true} (0.00) & \textit{\textcolor[HTML]{440154}{5}} (0.10) & \textcolor[HTML]{eed502}{BIPOP} (0.00) & \textit{\textcolor[HTML]{21918c}{mirrored}} (0.00) & \textit{\textcolor[HTML]{482677}{5}} (0.00) & \textcolor[HTML]{440154}{CSA} (0.08) & \textit{\textcolor[HTML]{eed502}{$\lambda$-decay}} (0.16) & 0.983122 & \textcolor[HTML]{cc4778}{centre}\\
5 & 2 & \textit{\textcolor[HTML]{eed502}{true}} (0.03) & \textit{\textcolor[HTML]{eed502}{Sobol}} (0.02) & \textit{\textcolor[HTML]{440154}{false}} (0.01) & \textcolor[HTML]{440154}{false} (0.91) & \textcolor[HTML]{440154}{8} (0.20) & \textit{\textcolor[HTML]{21918c}{IPOP}} (0.00) & \textcolor[HTML]{21918c}{mirrored} (0.04) & \textcolor[HTML]{482677}{5} (0.00) & \textcolor[HTML]{440154}{CSA} (0.03) & \textcolor[HTML]{eed502}{$\lambda$-decay} (0.46) & 0.907190 & \textcolor[HTML]{cc4778}{centre}\\
\bottomrule
\end{tabular}}
\end{table}

%% file: hallOfFame/de-hall-of-fame.tex

\begin{table}[ht]
\caption{Hall of fame of \textbf{modular DE}: best configurations per function including average best (denoted by All in fid), with resulting values of AOCC and structural BIAS indicator. Denoted in brackets after each module setting is the average effect in AOCC score of choosing the respective option versus all other choices (keeping the rest of the settings the same). Colouring of module options is consistent with Table~\ref{tab:modde_modules} and Figures~\ref{fig:de_shapxplaind5}, ~\ref{fig:de_shapxplaind30}. Options marked in \textit{italic} are consistently selected on this function for both dimensionalities. \label{tab:fame-de}}
\centering
\resizebox{\textwidth}{!}{
\begin{tabular}{ccrrrrrrrrrrcc}
\toprule
\multicolumn{1}{c}{d} & \multicolumn{1}{c}{fid} & \multicolumn{1}{c}{CR} & \multicolumn{1}{c}{F} & \multicolumn{1}{c}{adaptation} & \multicolumn{1}{c}{crossover} & \multicolumn{1}{c}{$\lambda$} & \multicolumn{1}{c}{lpsr} & \multicolumn{1}{c}{base} & \multicolumn{1}{c}{diffs} & \multicolumn{1}{c}{ref} & \multicolumn{1}{c}{archive} & \multicolumn{1}{c}{AOCC} & \multicolumn{1}{c}{BIAS} \\
\midrule
5 & \textbf{All} & \textcolor[HTML]{5ec962}{0.75} (0.09) & \textit{\textcolor[HTML]{3b528b}{0.5}} (0.09) & \textit{\textcolor[HTML]{eed502}{shade}} (0.06) & \textcolor[HTML]{eed502}{exp} (0.05) & \textcolor[HTML]{482173}{10} (0.18) & \textit{\textcolor[HTML]{440154}{false}} (0.02) & \textit{\textcolor[HTML]{eed502}{target}} (0.16) & \textcolor[HTML]{eed502}{2} (0.08) & \textit{\textcolor[HTML]{35b779}{pbest}} (0.05) & \textit{\textcolor[HTML]{eed502}{true}} (0.10) & 0.56690\textcolor[HTML]{440154}{8} & \textcolor[HTML]{f89540}{bounds} \\
5 & 1 & \textit{\textcolor[HTML]{3b528b}{0.25}} (0.25) & \textcolor[HTML]{440154}{0.25} (0.29) & \textcolor[HTML]{440154}{none} (0.12) & \textit{\textcolor[HTML]{440154}{bin}} (0.01) & \textcolor[HTML]{440154}{8} (0.06) & \textit{\textcolor[HTML]{440154}{false}} (0.19) & \textcolor[HTML]{440154}{best} (0.34) & \textit{\textcolor[HTML]{440154}{1}} (0.00) & \textcolor[HTML]{eed502}{rand} (0.02) & \textit{\textcolor[HTML]{eed502}{true}} (0.55) & 0.971233 & \textcolor[HTML]{7e03a8}{gaps/clusters} \\
5 & 2 & \textit{\textcolor[HTML]{3b528b}{0.25}} (0.37) & \textcolor[HTML]{440154}{0.25} (0.45) & \textit{\textcolor[HTML]{440154}{none}} (0.14) & \textit{\textcolor[HTML]{440154}{bin}} (0.01) & \textcolor[HTML]{440154}{8} (0.13) & \textit{\textcolor[HTML]{440154}{false}} (0.38) & \textcolor[HTML]{440154}{best} (0.31) & \textcolor[HTML]{eed502}{2} (0.00) & \textcolor[HTML]{eed502}{rand} (0.01) & \textcolor[HTML]{eed502}{true} (0.11) & 0.940875 & \textcolor[HTML]{f89540}{bounds} \\
\bottomrule
\end{tabular}}
\end{table}

%% file: acm_main.bbl
\begin{thebibliography}{10}

\bibitem{code_xplainer_niki_van_stein_2024_10568760}
{\sc Anonymous}.
\newblock nikivanstein/iohxplainer: v0.9.1 pre-final release.
\newblock \url{https://doi.org/10.5281/zenodo.10682980}, Feb. 2024.

\bibitem{auger2005restart}
{\sc Auger, A., and Hansen, N.}
\newblock A restart cma evolution strategy with increasing population size.
\newblock In {\em 2005 IEEE congress on evolutionary computation\/} (2005),
  vol.~2, IEEE, pp.~1769--1776.

\bibitem{bartzbeielstein2020benchmarking}
{\sc Bartz-Beielstein, T., Doerr, C., van~den Berg, D., Bossek, J.,
  Chandrasekaran, S., Eftimov, T., Fischbach, A., Kerschke, P., Cava, W.~L.,
  Lopez-Ibanez, M., Malan, K.~M., Moore, J.~H., Naujoks, B., Orzechowski, P.,
  Volz, V., Wagner, M., and Weise, T.}
\newblock Benchmarking in optimization: Best practice and open issues, 2020.

\bibitem{beyer2017simplify}
{\sc Beyer, H.-G., and Sendhoff, B.}
\newblock Simplify your covariance matrix adaptation evolution strategy.
\newblock {\em IEEE Transactions on Evolutionary Computation 21}, 5 (2017),
  746--759.

\bibitem{Boks2021}
{\sc Boks, R., Kononova, A.~V., and Wang, H.}
\newblock Quantifying the impact of boundary constraint handling methods on
  differential evolution.
\newblock In {\em Proceedings of the 2021 Genetic and Evolutionary Computation
  Conference Companion\/} (New York, NY, USA, July 2021), GECCO '21 Companion,
  Association for Computing Machinery, pp.~1199--–1207.

\bibitem{Brownlee2007}
{\sc Brownlee, J.}
\newblock A note on research methodology and benchmarking optimization
  algorithms.
\newblock Tech. Rep. 70125, Swinburne University of Technology, Vicotrial,
  Australia, 2007.

\bibitem{baeck2023evolutionary}
{\sc Bäck, T. H.~W., Kononova, A.~V., van Stein, B., Wang, H., Antonov, K.~A.,
  Kalkreuth, R.~T., de~Nobel, J., Vermetten, D., de~Winter, R., and Ye, F.}
\newblock {Evolutionary algorithms for parameter optimization—thirty years
  later}.
\newblock {\em Evolutionary Computation 31}, 2 (06 2023), 81--122.

\bibitem{psox}
{\sc Camacho{-}Villal{\'{o}}n, C.~L., Dorigo, M., and St{\"{u}}tzle, T.}
\newblock {PSO-X:} {A} component-based framework for the automatic design of
  particle swarm optimization algorithms.
\newblock {\em {IEEE} Trans. Evol. Comput. 26}, 3 (2022), 402--416.

\bibitem{caraffini2019study}
{\sc Caraffini, F., and Neri, F.}
\newblock A study on rotation invariance in differential evolution.
\newblock {\em Swarm and Evolutionary Computation 50\/} (2019), 100436.

\bibitem{modcma}
{\sc de~Nobel, J., Vermetten, D., Wang, H., Doerr, C., and B{\"{a}}ck, T.}
\newblock Tuning as a means of assessing the benefits of new ideas in interplay
  with existing algorithmic modules.
\newblock In {\em Proc. of Genetic and Evolutionary Computation Conference
  (GECCO'21, Companion material)\/} (2021), {ACM}, pp.~1375--1384.

\bibitem{deb2002fast}
{\sc Deb, K., Pratap, A., Agarwal, S., and Meyarivan, T.}
\newblock A fast and elitist multiobjective genetic algorithm: Nsga-ii.
\newblock {\em IEEE transactions on evolutionary computation 6}, 2 (2002),
  182--197.

\bibitem{drake2020recent}
{\sc Drake, J.~H., Kheiri, A., {\"O}zcan, E., and Burke, E.~K.}
\newblock Recent advances in selection hyper-heuristics.
\newblock {\em European Journal of Operational Research 285}, 2 (2020),
  405--428.

\bibitem{paradiseo22}
{\sc Dr{\'{e}}o, J., Liefooghe, A., V{\'{e}}rel, S., Schoenauer, M.,
  Guerv{\'{o}}s, J. J.~M., Quemy, A., Bouvier, B., and Gmys, J.}
\newblock Paradiseo: from a modular framework for evolutionary computation to
  the automated design of metaheuristics: 22 years of paradiseo.
\newblock In {\em {GECCO} '21: Genetic and Evolutionary Computation Conference,
  Companion Volume, Lille, France, July 10-14, 2021\/} (2021), K.~Krawiec, Ed.,
  {ACM}, pp.~1522--1530.

\bibitem{espinoza2023}
{\sc Espinoza, O., Rodr\'{\i}guez-V\'{a}zquez, K., Hern\'{a}ndez, C.~I., and
  Rodriguez-Romo, S.}
\newblock Comparison of three versions of whale optimization algorithm (woa) on
  the bbob test suite.
\newblock In {\em Proceedings of the Companion Conference on Genetic and
  Evolutionary Computation\/} (New York, NY, USA, 2023), GECCO '23 Companion,
  Association for Computing Machinery, p.~1595–1602.

\bibitem{evans2019s}
{\sc Evans, B.~P., Xue, B., and Zhang, M.}
\newblock What's inside the black-box? a genetic programming method for
  interpreting complex machine learning models.
\newblock In {\em Proceedings of the genetic and evolutionary computation
  conference\/} (2019), pp.~1012--1020.

\bibitem{ferreira2020applying}
{\sc Ferreira, L.~A., Guimar{\~a}es, F.~G., and Silva, R.}
\newblock Applying genetic programming to improve interpretability in machine
  learning models.
\newblock In {\em 2020 IEEE congress on evolutionary computation (CEC)\/}
  (2020), IEEE, pp.~1--8.

\bibitem{fontaine2023}
{\sc Fontaine, M., and Nikolaidis, S.}
\newblock Covariance matrix adaptation map-annealing.
\newblock In {\em Proceedings of the Genetic and Evolutionary Computation
  Conference\/} (New York, NY, USA, 2023), GECCO '23, Association for Computing
  Machinery, p.~456–465.

\bibitem{Fyvie2021}
{\sc Fyvie, M., McCall, J. A.~W., and Christie, L.~A.}
\newblock Towards explainable metaheuristics: Pca for trajectory mining in
  evolutionary algorithms.
\newblock In {\em Artificial Intelligence XXXVIII\/} (Cham, 2021), M.~Bramer
  and R.~Ellis, Eds., Springer International Publishing, pp.~89--102.

\bibitem{mengqi2022}
{\sc Gao, M., Feng, X., Yu, H., and Zheng, Z.}
\newblock Multi-granularity competition-cooperation optimization algorithm with
  adaptive parameter configuration.
\newblock {\em Applied Intelligence 52\/} (2022), 13132–13161.

\bibitem{glotic2015}
{\sc Glotić, A., and Zamuda, A.}
\newblock Short-term combined economic and emission hydrothermal optimization
  by surrogate differential evolution.
\newblock {\em Applied Energy 141\/} (2015), 42--56.

\bibitem{hansen2009benchmarking}
{\sc Hansen, N.}
\newblock Benchmarking a bi-population cma-es on the bbob-2009 function
  testbed.
\newblock In {\em Proceedings of the 11th annual conference companion on
  genetic and evolutionary computation conference: late breaking papers\/}
  (2009), pp.~2389--2396.

\bibitem{hansen2022anytime}
{\sc Hansen, N., Auger, A., Brockhoff, D., and Tu{\v{s}}ar, T.}
\newblock Anytime performance assessment in blackbox optimization benchmarking.
\newblock {\em IEEE Transactions on Evolutionary Computation 26}, 6 (2022),
  1293--1305.

\bibitem{hansen2021coco}
{\sc Hansen, N., Auger, A., Ros, R., Mersmann, O., Tu{\v{s}}ar, T., and
  Brockhoff, D.}
\newblock Coco: A platform for comparing continuous optimizers in a black-box
  setting.
\newblock {\em Optimization Methods and Software 36}, 1 (2021), 114--144.

\bibitem{bbobfunctions}
{\sc Hansen, N., Finck, S., Ros, R., and Auger, A.}
\newblock {Real-Parameter Black-Box Optimization Benchmarking 2009: Noiseless
  Functions Definitions}.
\newblock Tech. Rep. RR-6829, {INRIA}, 2009.

\bibitem{hansen2011impacts}
{\sc Hansen, N., Ros, R., Mauny, N., Schoenauer, M., and Auger, A.}
\newblock Impacts of invariance in search: When cma-es and pso face
  ill-conditioned and non-separable problems.
\newblock {\em Applied Soft Computing 11}, 8 (2011), 5755--5769.

\bibitem{hutter2014efficient}
{\sc Hutter, F., Hoos, H., and Leyton-Brown, K.}
\newblock An efficient approach for assessing hyperparameter importance.
\newblock In {\em International conference on machine learning\/} (2014), PMLR,
  pp.~754--762.

\bibitem{dejong1981using}
{\sc Jong, K. A.~D., and Spears, W.~M.}
\newblock Using genetic algorithms to solve np-complete problems.
\newblock In {\em Proceedings of the 3rd International Conference on Genetic
  Algorithms, George Mason University, Fairfax, Virginia, USA, June 1989\/}
  (1989), J.~D. Schaffer, Ed., Morgan Kaufmann, pp.~124--132.

\bibitem{kaveh2016}
{\sc Kaveh, A., and Bakhshpoori, T.}
\newblock Water evaporation optimization: A novel physically inspired
  optimization algorithm.
\newblock {\em Computers \& Structures 167\/} (2016), 69--85.

\bibitem{kerschke2019automated}
{\sc Kerschke, P., and Trautmann, H.}
\newblock Automated algorithm selection on continuous black-box problems by
  combining exploratory landscape analysis and machine learning.
\newblock {\em Evolutionary computation 27}, 1 (2019), 99--127.

\bibitem{kohira2018}
{\sc Kohira, T., Kemmotsu, H., Akira, O., and Tatsukawa, T.}
\newblock Proposal of benchmark problem based on real-world car structure
  design optimization.
\newblock In {\em Proceedings of the Genetic and Evolutionary Computation
  Conference Companion\/} (New York, NY, USA, 2018), GECCO '18, Association for
  Computing Machinery, p.~183–184.

\bibitem{Kononova2015}
{\sc Kononova, A.~V., Corne, D.~W., Wilde, P.~D., Shneer, V., and Caraffini,
  F.}
\newblock Structural bias in population-based algorithms.
\newblock {\em Information Sciences 298\/} (2015), 468--490.

\bibitem{kononova2023tiobr}
{\sc Kononova, A.~V., Vermetten, D., Caraffini, F., Mitran, M.-A., and Zaharie,
  D.}
\newblock {The importance of being constrained: dealing with infeasible
  solutions in differential evolution and beyond}.
\newblock {\em Evolutionary Computation\/} (11 2023), 1--46.

\bibitem{kostovska2022importance}
{\sc Kostovska, A., Vermetten, D., D{\v{z}}eroski, S., Doerr, C., Korosec, P.,
  and Eftimov, T.}
\newblock The importance of landscape features for performance prediction of
  modular cma-es variants.
\newblock In {\em Proceedings of the Genetic and Evolutionary Computation
  Conference\/} (2022), pp.~648--656.

\bibitem{KG_mod_pred}
{\sc Kostovska, A., Vermetten, D., Dzeroski, S., Panov, P., Eftimov, T., and
  Doerr, C.}
\newblock Using knowledge graphs for performance prediction of modular
  optimization algorithms.
\newblock In {\em Applications of Evolutionary Computation - 26th European
  Conference, EvoApplications 2023, Held as Part of EvoStar 2023, Brno, Czech
  Republic, April 12-14, 2023, Proceedings\/} (2023), J.~Correia, S.~L. Smith,
  and R.~Qaddoura, Eds., vol.~13989 of {\em Lecture Notes in Computer Science},
  Springer, pp.~253--268.

\bibitem{kostovska_inprogress}
{\sc Kostovska, A., Vermetten, D., Korosec, P., D{\v{z}}eroski, S., Doerr, C.,
  and Eftimov, T.}
\newblock Unveiling the role of modules: Assessing importance and classifying
  modules in modcma-es and modde via algorithmic behavior analysis.

\bibitem{lensen2020genetic}
{\sc Lensen, A., Xue, B., and Zhang, M.}
\newblock Genetic programming for evolving a front of interpretable models for
  data visualization.
\newblock {\em IEEE transactions on cybernetics 51}, 11 (2020), 5468--5482.

\bibitem{lindauer2022smac3}
{\sc Lindauer, M., Eggensperger, K., Feurer, M., Biedenkapp, A., Deng, D.,
  Benjamins, C., Ruhkopf, T., Sass, R., and Hutter, F.}
\newblock Smac3: A versatile bayesian optimization package for hyperparameter
  optimization.
\newblock {\em The Journal of Machine Learning Research 23}, 1 (2022),
  2475--2483.

\bibitem{configspace}
{\sc Lindauer, M., Eggensperger, K., Feurer, M., Biedenkapp, A., Marben, J.,
  Müller, P., and Hutter, F.}
\newblock Boah: A tool suite for multi-fidelity bayesian optimization \&
  analysis of hyperparameters.
\newblock {\em arXiv:1908.06756 {[cs.LG]}\/}.

\bibitem{long2022learning}
{\sc Long, F.~X., van Stein, B., Frenzel, M., Krause, P., Gitterle, M., and
  B{\"a}ck, T.}
\newblock Learning the characteristics of engineering optimization problems
  with applications in automotive crash.
\newblock In {\em Proceedings of the Genetic and Evolutionary Computation
  Conference\/} (2022), pp.~1227--1236.

\bibitem{long2023bbob}
{\sc Long, F.~X., Vermetten, D., van Stein, B., and Kononova, A.~V.}
\newblock Bbob instance analysis: Landscape properties and algorithm
  performance across problem instances.
\newblock In {\em International Conference on the Applications of Evolutionary
  Computation (Part of EvoStar)\/} (2023), Springer, pp.~380--395.

\bibitem{lopez2010automatic}
{\sc L{\'o}pez-Ib{\'a}{\~n}ez, M., and St{\"u}tzle, T.}
\newblock Automatic configuration of multi-objective aco algorithms.
\newblock In {\em International Conference on Swarm Intelligence\/} (2010),
  Springer, pp.~95--106.

\bibitem{lopez2012automatic}
{\sc L{\'o}pez-Ib{\'a}nez, M., and Stutzle, T.}
\newblock The automatic design of multiobjective ant colony optimization
  algorithms.
\newblock {\em IEEE Transactions on Evolutionary Computation 16}, 6 (2012),
  861--875.

\bibitem{lundberg2020local2global}
{\sc Lundberg, S.~M., Erion, G., Chen, H., DeGrave, A., Prutkin, J.~M., Nair,
  B., Katz, R., Himmelfarb, J., Bansal, N., and Lee, S.-I.}
\newblock From local explanations to global understanding with explainable ai
  for trees.
\newblock {\em Nature Machine Intelligence 2}, 1 (2020), 2522--5839.

\bibitem{NIPS2017_7062}
{\sc Lundberg, S.~M., and Lee, S.-I.}
\newblock A unified approach to interpreting model predictions.
\newblock In {\em Advances in Neural Information Processing Systems 30},
  I.~Guyon, U.~V. Luxburg, S.~Bengio, H.~Wallach, R.~Fergus, S.~Vishwanathan,
  and R.~Garnett, Eds. Curran Associates, Inc., 2017, pp.~4765--4774.

\bibitem{malan_survey_2021}
{\sc Malan, K.~M.}
\newblock A {Survey} of {Advances} in {Landscape} {Analysis} for
  {Optimisation}.
\newblock {\em Algorithms 14}, 2 (Jan. 2021), 40.

\bibitem{mersmann2011exploratory}
{\sc Mersmann, O., Bischl, B., Trautmann, H., Preuss, M., Weihs, C., and
  Rudolph, G.}
\newblock Exploratory landscape analysis.
\newblock In {\em Proceedings of the 13th annual conference on Genetic and
  evolutionary computation\/} (2011), pp.~829--836.

\bibitem{Mitran2023}
{\sc Mitran, M.-A., Kononova, A.~V., Caraffini, F., and Zaharie, D.}
\newblock Patterns of convergence and bound constraint violation in
  differential evolution on sbox-cost benchmarking suite.
\newblock In {\em Proceedings of the Companion Conference on Genetic and
  Evolutionary Computation\/} (2023), GECCO '23 Companion, Association for
  Computing Machinery, p.~2337–2345.

\bibitem{munoz2015alg}
{\sc Muñoz, M.~A., Sun, Y., Kirley, M., and Halgamuge, S.~K.}
\newblock Algorithm selection for black-box continuous optimization problems: A
  survey on methods and challenges.
\newblock {\em Information Sciences 317\/} (2015), 224--245.

\bibitem{OCHOA2021SearchTrajectoryNetworks}
{\sc Ochoa, G., Malan, K.~M., and Blum, C.}
\newblock Search trajectory networks: A tool for analysing and visualising the
  behaviour of metaheuristics.
\newblock {\em Applied Soft Computing 109\/} (2021), 107492.

\bibitem{pierezan2018}
{\sc Pierezan, J., and Dos Santos~Coelho, L.}
\newblock Coyote optimization algorithm: A new metaheuristic for global
  optimization problems.
\newblock In {\em 2018 IEEE Congress on Evolutionary Computation (CEC)\/}
  (2018).

\bibitem{prager2020per}
{\sc Prager, R.~P., Trautmann, H., Wang, H., B{\"a}ck, T.~H., and Kerschke, P.}
\newblock Per-instance configuration of the modularized cma-es by means of
  classifier chains and exploratory landscape analysis.
\newblock In {\em 2020 IEEE Symposium Series on Computational Intelligence
  (SSCI)\/} (2020), IEEE, pp.~996--1003.

\bibitem{nevergrad}
{\sc Rapin, J., and Teytaud, O.}
\newblock {Nevergrad - A gradient-free optimization platform}.
\newblock \url{https://GitHub.com/FacebookResearch/Nevergrad}, 2018.

\bibitem{ribeiro2016should}
{\sc Ribeiro, M.~T., Singh, S., and Guestrin, C.}
\newblock " why should i trust you?" explaining the predictions of any
  classifier.
\newblock In {\em Proceedings of the 22nd ACM SIGKDD international conference
  on knowledge discovery and data mining\/} (2016), pp.~1135--1144.

\bibitem{schede2022survey}
{\sc Schede, E., Brandt, J., Tornede, A., Wever, M., Bengs, V.,
  H{\"u}llermeier, E., and Tierney, K.}
\newblock A survey of methods for automated algorithm configuration.
\newblock {\em Journal of Artificial Intelligence Research 75\/} (2022),
  425--487.

\bibitem{sobol1993sensitivity}
{\sc Sobol', I.}
\newblock Sensitivity estimates for nonlinear mathematical models.
\newblock {\em Math. Model. Comput. Exp. 1\/} (1993), 407.

\bibitem{GSApaper}
{\sc Stein, B.~V., Raponi, E., Sadeghi, Z., Bouman, N., Van~Ham, R. C. H.~J.,
  and Bäck, T.}
\newblock A comparison of global sensitivity analysis methods for explainable
  ai with an application in genomic prediction.
\newblock {\em IEEE Access 10\/} (2022), 103364--103381.

\bibitem{Tanabe2020}
{\sc Tanabe, R., and Ishibuchi, H.}
\newblock An easy-to-use real-world multi-objective optimization problem suite.
\newblock {\em Applied Soft Computing 89\/} (2020), 106078.

\bibitem{thomaser2023optimizing}
{\sc Thomaser., A., Vogt., M., Bäck., T., and Kononova, A.}
\newblock Optimizing cma-es with cma-es.
\newblock In {\em Proceedings of the 15th International Joint Conference on
  Computational Intelligence - ECTA\/} (2023), INSTICC, SciTePress,
  pp.~214--221.

\bibitem{trajanov2021explainable}
{\sc Trajanov, R., Dimeski, S., Popovski, M., Koro{\v{s}}ec, P., and Eftimov,
  T.}
\newblock Explainable landscape-aware optimization performance prediction.
\newblock In {\em Symposium Series on Computational Intelligence\/} (New York,
  NY, USA, 2021), IEEE, pp.~01--08.

\bibitem{trajanov2022explainable}
{\sc Trajanov, R., Dimeski, S., Popovski, M., Korošec, P., and Eftimov, T.}
\newblock Explainable landscape analysis in automated algorithm performance
  prediction, 2022.

\bibitem{nevergrad_wizard}
{\sc Trajanov, R., Nikolikj, A., Cenikj, G., Teytaud, F., Videau, M., Teytaud,
  O., Eftimov, T., L{\'{o}}pez{-}Ib{\'{a}}{\~{n}}ez, M., and Doerr, C.}
\newblock Improving nevergrad's algorithm selection wizard ngopt through
  automated algorithm configuration.
\newblock In {\em Parallel Problem Solving from Nature - {PPSN} {XVII} - 17th
  International Conference, {PPSN} 2022, Dortmund, Germany, September 10-14,
  2022, Proceedings, Part {I}\/} (2022), G.~Rudolph, A.~V. Kononova, H.~E.
  Aguirre, P.~Kerschke, G.~Ochoa, and T.~Tusar, Eds., vol.~13398 of {\em
  Lecture Notes in Computer Science}, Springer, pp.~18--31.

\bibitem{improvingnevergrad}
{\sc Trajanov, R., Nikolikj, A., Cenikj, G., Teytaud, F., Videau, M., Teytaud,
  O., Eftimov, T., L{\'o}pez-Ib{\'a}{\~{n}}ez, M., and Doerr, C.}
\newblock Improving nevergrad's algorithm selection wizard ngopt through
  automated algorithm configuration.
\newblock In {\em Parallel Problem Solving from Nature -- PPSN XVII\/} (Cham,
  2022), G.~Rudolph, A.~V. Kononova, H.~Aguirre, P.~Kerschke, G.~Ochoa, and
  T.~Tu{\v{s}}ar, Eds., Springer International Publishing, pp.~18--31.

\bibitem{vanderblom2023}
{\sc van~der Blom, K., Deist, T.~M., Volz, V., Marchi, M., Nojima, Y., Naujoks,
  B., Oyama, A., and Tu{\v{s}}ar, T.}
\newblock {\em Identifying Properties of Real-World Optimisation Problems
  Through a Questionnaire}.
\newblock Springer International Publishing, Cham, 2023, pp.~59--80.

\bibitem{van2023doe2vec}
{\sc van Stein, B., Long, F.~X., Frenzel, M., Krause, P., Gitterle, M., and
  B{\"a}ck, T.}
\newblock Doe2vec: Deep-learning based features for exploratory landscape
  analysis.
\newblock {\em arXiv preprint arXiv:2304.01219\/} (2023).

\bibitem{GSAcode}
{\sc Van~Stein, B., and Raponi, E.}
\newblock Gsareport: Easy to use global sensitivity reporting.
\newblock {\em Journal of Open Source Software 7}, 78 (2022), 4721.

\bibitem{deepbias}
{\sc Van~Stein, B., Vermetten, D., Caraffini, F., and Kononova, A.~V.}
\newblock Deep bias: Detecting structural bias using explainable ai.
\newblock In {\em Proceedings of the Companion Conference on Genetic and
  Evolutionary Computation\/} (New York, NY, USA, 2023), GECCO '23 Companion,
  Association for Computing Machinery, p.~455–458.

\bibitem{modDE}
{\sc Vermetten, D., Caraffini, F., Kononova, A.~V., and B\"{a}ck, T.}
\newblock Modular differential evolution.
\newblock In {\em Proceedings of the Genetic and Evolutionary Computation
  Conference\/} (New York, NY, USA, 2023), GECCO '23, Association for Computing
  Machinery, p.~864–872.

\bibitem{vermetten2024largescale}
{\sc Vermetten, D., Doerr, C., Wang, H., Kononova, A.~V., and Bäck, T.}
\newblock Large-scale benchmarking of metaphor-based optimization heuristics,
  2024.

\bibitem{biastoolbox}
{\sc Vermetten, D., van Stein, B., Caraffini, F., Minku, L.~L., and Kononova,
  A.~V.}
\newblock Bias: A toolbox for benchmarking structural bias in the continuous
  domain.
\newblock {\em IEEE Transactions on Evolutionary Computation 26}, 6 (2022),
  1380--1393.

\bibitem{wang2022iohanalyzer}
{\sc Wang, H., Vermetten, D., Ye, F., Doerr, C., and B{\"a}ck, T.}
\newblock Iohanalyzer: Detailed performance analyses for iterative optimization
  heuristics.
\newblock {\em ACM Transactions on Evolutionary Learning and Optimization 2}, 1
  (2022), 1--29.

\bibitem{NFLT}
{\sc Wolpert, D., and Macready, W.}
\newblock No free lunch theorems for optimization.
\newblock {\em IEEE Transactions on Evolutionary Computation 1\/} (1997),
  67--82.

\bibitem{cec2017}
{\sc Wu, G., Mallipeddi, R., and Suganthan, P.~N.}
\newblock Problem definitions and evaluation criteria for the cec 2017
  competition on constrained real-parameter optimization.
\newblock {\em National University of Defense Technology, Changsha, Hunan, PR
  China and Kyungpook National University, Daegu, South Korea and Nanyang
  Technological University, Singapore, Technical Report\/} (2017).

\bibitem{zamuda2019success}
{\sc Zamuda, A., and Sosa, J. D.~H.}
\newblock Success history applied to expert system for underwater glider path
  planning using differential evolution.
\newblock {\em Expert Systems with Applications 119\/} (2019), 155--170.

\bibitem{Zhou2024}
{\sc Zhou, R., and Hu, T.}
\newblock {\em Evolutionary Approaches to Explainable Machine Learning}.
\newblock Springer Nature Singapore, Singapore, 2024, pp.~487--506.

\end{thebibliography}
